\pdfoutput=1
\documentclass[final]{cvpr}

\usepackage{times}
\usepackage{epsfig}
\usepackage{graphicx}
\usepackage{amsmath}
\usepackage{amssymb}

\usepackage{booktabs, multirow} 
\usepackage{soul}
\usepackage[table]{xcolor} 
\usepackage{changepage,threeparttable} 
\usepackage{adjustbox}
\usepackage{appendix}

\usepackage[pagebackref=true,breaklinks=true,colorlinks,bookmarks=false]{hyperref}

\def\cvprPaperID{7067} 
\def\confYear{CVPR 2021}

\usepackage{appendix}
\newcommand*\rot{\rotatebox{90}}

\begin{document}

\title{A Closer Look at Fourier Spectrum Discrepancies\\for CNN-generated Images Detection }
\author{Keshigeyan Chandrasegaran
\qquad
Ngoc-Trung Tran
\qquad
Ngai-Man Cheung\\
Singapore University of Technology and Design (SUTD)\\
{\tt\small \{ keshigeyan, ngoctrung\_tran, ngaiman\_cheung \} @sutd.edu.sg}
}
\maketitle

\begin{abstract}
   CNN-based generative modelling has evolved to produce synthetic images indistinguishable from real images in the RGB pixel space. 
   Recent works have observed that CNN-generated images share a systematic shortcoming in replicating high frequency Fourier spectrum decay attributes. 
   Furthermore, these works have successfully exploited this systematic shortcoming to detect CNN-generated images reporting up to 99\% accuracy across multiple state-of-the-art GAN models.
   
   In this work, we investigate the validity of assertions claiming that CNN-generated images are unable to achieve high frequency spectral decay consistency. 
   We meticulously construct a counterexample space of high frequency spectral decay consistent CNN-generated images emerging from our handcrafted experiments using DCGAN, LSGAN, WGAN-GP and StarGAN, where we empirically show that this frequency discrepancy can be avoided by a minor architecture change in the last upsampling operation. 
   We subsequently use images from this counterexample space to successfully bypass the recently proposed forensics detector which leverages on high frequency Fourier spectrum decay attributes for CNN-generated image detection.
   
   Through this study, we show that high frequency Fourier spectrum decay discrepancies are not inherent characteristics for existing CNN-based generative models---contrary to the belief of some existing work---, and such features are not robust to perform synthetic image detection. 
   Our results prompt re-thinking of using high frequency Fourier spectrum decay attributes for CNN-generated image detection. 
   Code and models are available at
   \url{https://keshik6.github.io/Fourier-Discrepancies-CNN-Detection/}
\end{abstract}

\section{Introduction}
With serious concerns over Deepfakes being widely used for malicious purposes \cite{hao_heaven_2020, harrison_2021,  doi:10.1080/23268743.2020.1757499, thomas_2020, baines_2020, simonite_2020, citron_chesney_2020, metz_2019},
detection of deepfake multimedia content has become an important research field. With substantial improvement of CNN-based generative modelling in the recent years~\cite{NEURIPS2020_8d30aa96, zhao2020diffaugment, Karras_2019_CVPR, Karras_2020_CVPR, karras2018progressive, NEURIPS2019_5f8e2fa1, Choi_2020_CVPR, brock2018large, CycleGAN2017, Park_2019_CVPR, 8099502, arjovsky2017wasserstein, brock2019large, NIPS2014_5ca3e9b1, Tran2018DistGANAI, NEURIPS2019_d04cb95b},
it is becoming more and more difficult to assess the ``fakeness'' of such synthetic content in the RGB pixel space.

\subsection{Fourier spectrum discrepancies in CNN-generated images}
\label{ssec:intro:discrepancy}
\begin{figure}
\centering
\begin{tabular}{ccc}
      \includegraphics[width=0.9\linewidth]{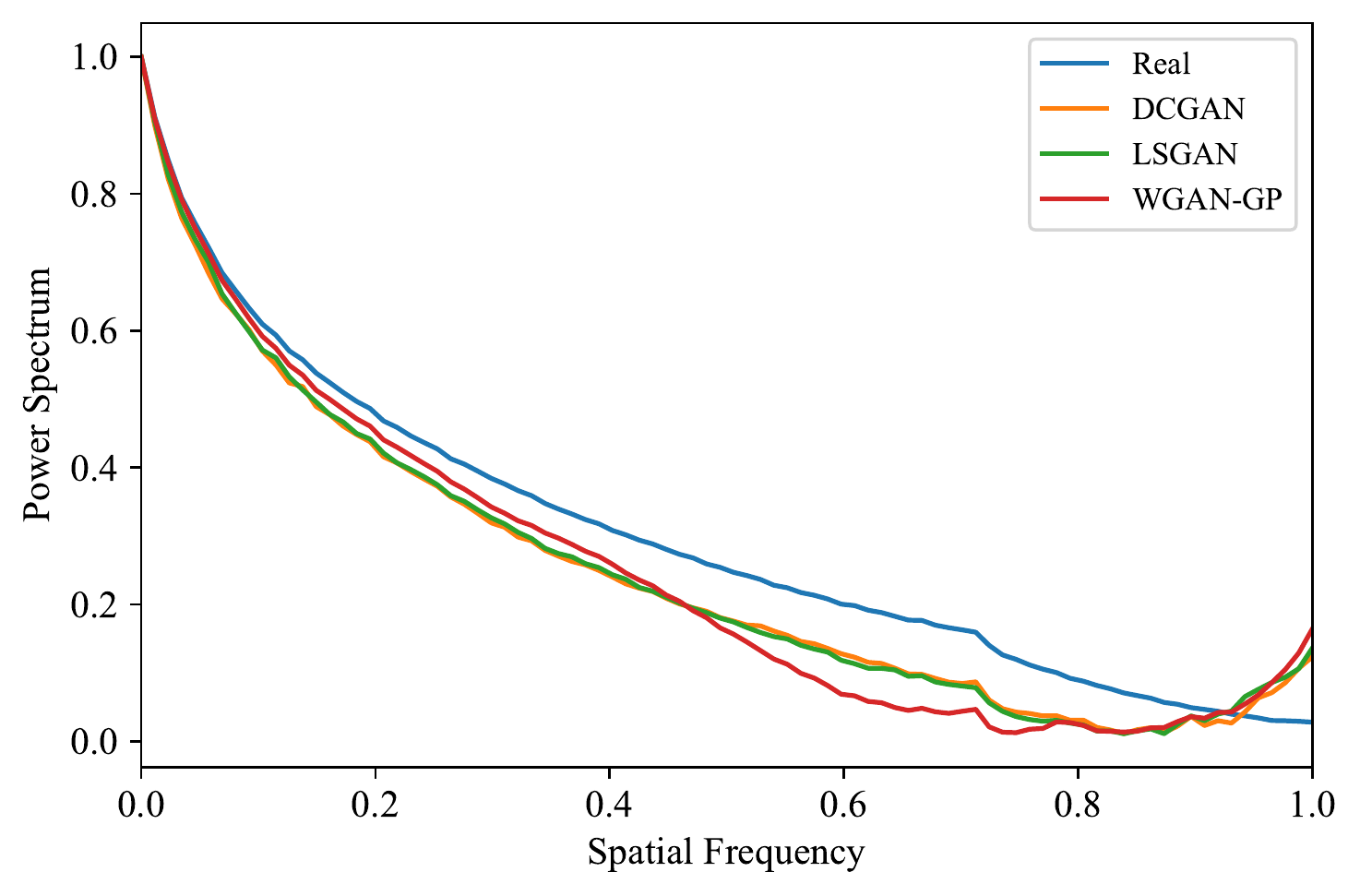}\\
      \includegraphics[width=0.9\linewidth]{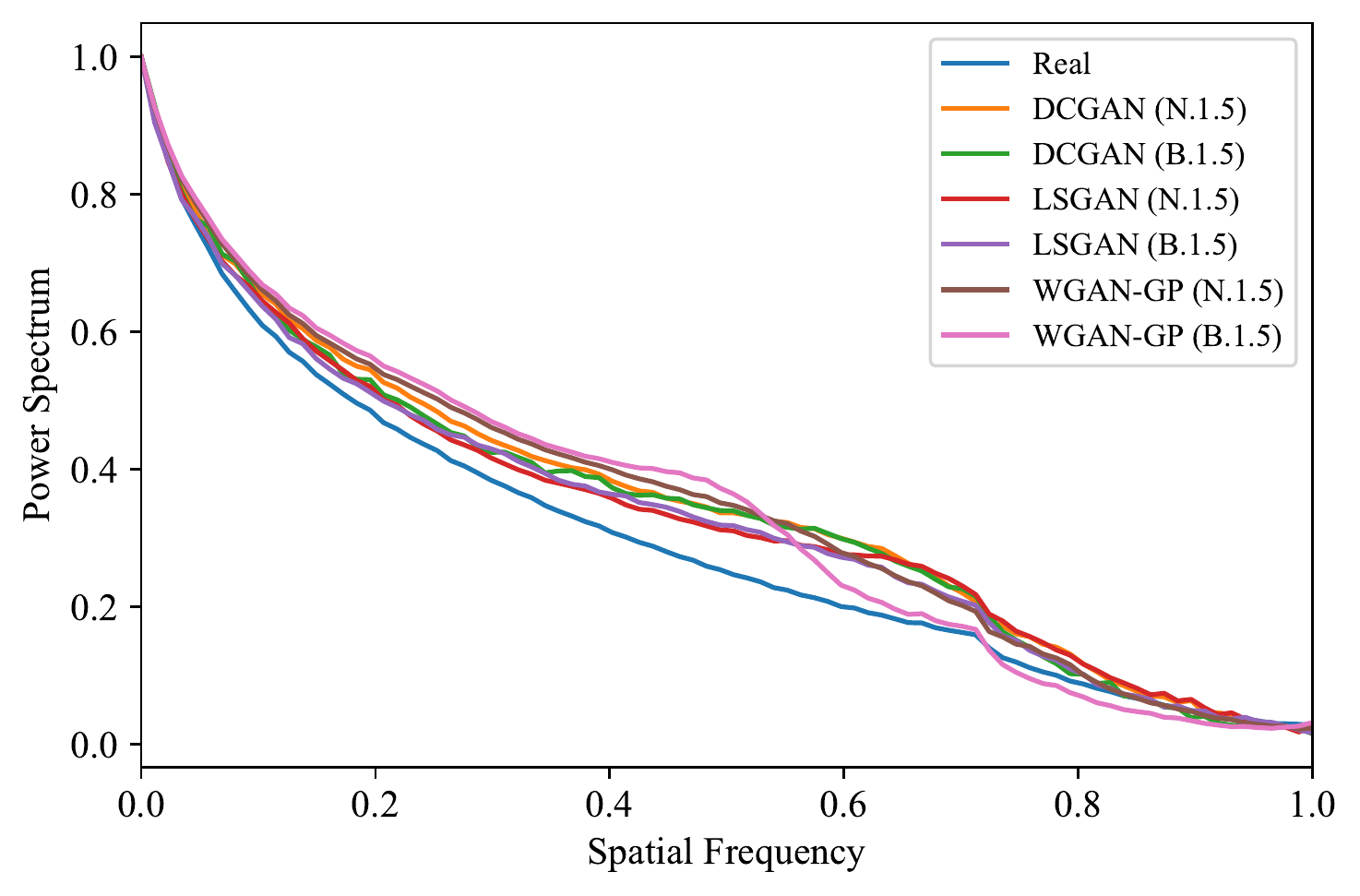}
\end{tabular}
\caption{The curves show the average azimuthal integration over the power spectrum. (See section \ref{sec:fft}). Top row shows the evaluation on DCGAN \cite{radford2016unsupervised}, LSGAN \cite{lsgan}, WGAN-GP \cite{NIPS2017_892c3b1c}.
Note the discrepancies at the highest frequencies, the same as reported in recent works. Note also that these models use transpose convolutions. The bottom row shows the evaluation after replacing the last feature map scaling operation with nearest and bilinear interpolation methods. Refer to table \ref{table:1} for experiment codes. All evaluation are done using CelebA \cite{liu2015faceattributes} (128x128). We observe that spectral consistent GANs are obtained when using nearest and bilinear interpolation methods for the last feature map scaling operation.}
\label{fig:intro}
\vspace{-0.1cm}
\end{figure}

Recent research suggests that CNN-based generation methods are unable to reproduce high frequency distribution of real images.
Existing work tends to conclude that this incompetency is an intrinsic property of CNN-based generative models \cite{dzanic2020fourier, Durall_2020_CVPR, khayatkhoei2020spatial}. While Zhang \etal\cite{GAN_Artifacts} and Wang \etal\cite{Wang_2020_CVPR} report that CNN generated images have frequency artifacts, Dzanic \etal\cite{dzanic2020fourier} and  Durall \etal\cite{Durall_2020_CVPR}
suggest spectrum discrepancies in high frequency: 
{\em CNN-generated images at the highest frequencies do not decay as usually observed in real images} (Figure \ref{fig:intro}). In particular,
\begin{itemize}
\item
Dzanic \etal\cite{dzanic2020fourier} analyze high frequency of real and deep network generated images, and conclude that ``deep network generated images share an observable, systematic shortcoming in replicating the attributes of these high-frequency modes''. 
\item
Durall \etal\cite{Durall_2020_CVPR} observe that ``CNN based generative deep neural networks are failing to reproduce spectral distributions'', and ``this effect is independent of the underlying architecture''. 
\item
Dzanic \etal\cite{dzanic2020fourier} take a step further and propose to exploit this frequency discrepancies for detection of deep network generated images, claiming an accuracy of up to 99.2\% across multiple state-of-the-art GAN and VAE models.
\end{itemize}

Some works also propose different techniques to disguise
these high frequency discrepancies via post-processing the
deep network generated images
\cite{dzanic2020fourier, khayatkhoei2020spatial},
or modifying the
GAN training objective to avoid these discrepancies \cite{Durall_2020_CVPR}. 

It should be noted that 
the cause of this discrepancy\footnote{The terms spectral discrepancies and spectral
inconsistency are used invariably where we refer to high frequency spectral decay discrepancies. 
We also use the terms CNN-generated and synthetic invariably.} has not been agreed upon.
Zhang \etal\cite{GAN_Artifacts} and Durall \etal\cite{Durall_2020_CVPR} suggest that this discrepancy could be caused by transposed convolution. As transposed convolution is used throughout the generator architectures, it is difficult to replace them. Therefore, Durall \etal\cite{Durall_2020_CVPR} propose spectral regularization to counteract this throughout the GAN training.
Meanwhile, 
Dzanic \etal\cite{dzanic2020fourier}
attribute
this discrepancy to the linear dependencies in the spectrum of convolutional filters \cite{khayatkhoei2020spatial}, which hinder learning of high frequencies.

\subsection{CNN-generated Image Detectors}
Many works have addressed the possibilities of creating detectors to identify synthetic images apart from real images. Though synthetic image detectors (CNN-based) worked reasonably with RGB inputs \cite{Wang_2020_CVPR, deepfake_properties_eccv20}, with spectral discrepancies being  observed, several works have proposed to use the corresponding Fourier representation to train these detectors \cite{GAN_Artifacts, pmlr-v119-frank20a}. 
In particular, 
Frank \etal \cite{pmlr-v119-frank20a} showed that detectors using frequency domain yielded better results compared to using the RGB counterpart.

Though observations corresponding to distinguishable frequency footprints being left by CNN-generated images remained mostly qualitative, Dzanic \etal \cite{dzanic2020fourier} and Durall \etal \cite{Durall_2020_CVPR, durall2020unmasking} studied the spectral decay attributes and quantified this behaviour via averaging the power spectrum over frequencies radially to represent them as 1D information. Through this, they observed high frequency spectral decay inconsistencies in CNN-generated images.
Furthermore, 
Dzanic \etal \cite{dzanic2020fourier} proposed to use a simple KNN classifier using high frequency spectral attributes (3 features extracted per image) that surprisingly obtained very high accuracy in identifying synthetic images only with very small amount of training data. Similarly, Durall \etal \cite{durall2020unmasking} used the entire 1d-power spectrum (contains all frequency information) as features to perform detection using Logistic regression, SVM and K-means clustering algorithms. These detectors \cite{dzanic2020fourier, durall2020unmasking} rely on the belief that high frequency decay discrepancies are intrinsic in CNN-generated images.

\subsection{Our contributions}
\label{ssec:intro:contributions}
In this work, we take a closer look at the high frequency decay discrepancies in CNN-generated images. 
Analysis of CNN-generated images is a daunting task: a large number of different architectures, algorithms and objective functions have been proposed to train generators. 
Instead, our study focuses on the {\em last layer} of the generators.
Our justification is as follows. 
Based on Sampling Theorem \cite{jain:dip_book}, the 
frequency contents of the outputs of individual layers are limited by the
sampling 
resolution of the corresponding outputs.
As previous work has reported the  
discrepancies between real and CNN-generated images at the {\em highest frequencies}, 
we hypothesize that inner generator layers (which produce lower resolution outputs) are not directly responsible for the high frequency discrepancies. Therefore, we focus on the last upsampling layer of generative CNN models.

Due to our focus being localized at the last layer,
we are able to pinpoint the component that is related to this discrepancy 
across multiple GAN loss functions, architectures, datasets and resolutions.
Our candidate GANs are
similar to 
Durall \etal\cite{Durall_2020_CVPR}:
DCGAN \cite{radford2016unsupervised}, LSGAN \cite{lsgan}, WGAN-GP \cite{NIPS2017_892c3b1c} and we also extend to StarGAN \cite{choi2018stargan}.  
Our experiments suggest that the frequency discrepancies can be largely 
avoided
by simply
modifying the feature map scaling of the last layer. 
Importantly, using the {\em same} training algorithms, objective functions and network architectures (except using a different scaling in the last layer) 
as in standard GAN models, 
we are able to avoid the spectral discrepancies. Therefore, 
our work provides {\em counterexamples}
to argue that high frequency discrepancies are not intrinsic for CNN-generated images. Furthermore, we are able to successfully bypass the synthetic image detector proposed by Dzanic \etal\cite{dzanic2020fourier} with only such change in the last scaling step, showing that such approach may not be reliable for detection of deep network generated images. The {\em key takeaway} from our work is:
\begin{itemize}
\item High frequency spectral decay discrepancies are not intrinsic for CNN-generated images. Therefore, we urge re-thinking in using such features for CNN-generated image detection.
\end{itemize}

\section{Related Work}
Dzanic \etal\cite{dzanic2020fourier} show that CNN-generated images (GANs and VAEs) demonstrate different Fourier spectrum decay characteristics. Since the spectra of natural images tend to behave following the power law \cite{VANDERSCHAAF19962759}, Dzanic \etal\cite{dzanic2020fourier} show that the Fourier modes of deep network generated images at the highest frequencies did not decay as seen in real images, but instead stayed approximately constant. 
Furthermore,
they propose to exploit these discrepancies to detect CNN-generated synthetic images by fitting a decay function to the reduced spectra, and using the parameters of the fitted decay function to build a simple kNN classifier. 

Durall \etal\cite{Durall_2020_CVPR} show that popular GAN image generators fail to approximate the spectral distributions of real data, and they attribute this to the use of transpose convolutions for upsampling. They show that this effect is independent of the underlying architecture using 1-dimensional spectral characteristics of images generated from DCGAN \cite{radford2016unsupervised}, LSGAN \cite{lsgan}, WGAN-GP\cite {NIPS2017_892c3b1c} and DRAGAN \cite{kodali2017convergence}. 
Since  transpose convolutions are used in the entire generator models, the propose to counteract their effects by adding 
a spectral regularization term to the Generator, thereby penalizing the generator for spectral distorted samples. 

Khayatkhoei and Elgammal \cite{khayatkhoei2020spatial} suggest the presence of a systematic bias in GANs against learning high frequencies. They specifically show that for a given kernel size, as resolution increases, the correlation/ dependency of the kernel's spectrum increases thereby systematically preventing GANs to learn high frequencies without affecting the adjacent frequencies. To alleviate this shortcoming, they propose frequency shifted generators whose frequencies are shifted towards specific high frequencies.

\section{Background}
\label{sec:fft}
The 2D discrete Fourier transform $F(k_{x}, k_{y})$ of a $M \times N$ 2D image $f(x, y)$   can be written as,
\begin{equation}
F(k_x, k_y) = \frac{1}{MN}\sum_{u=0}^{M-1}\sum_{v=0}^{N-1}f(u, v)e^{-i2\pi(\frac{k_xu}{M} + \frac{k_yv}{N})}
\label{eq:1}
\end{equation}
for $k_x \in \{0, 1, 2, ..., M-1\}$, $k_y \in \{0, 1, 2, ..., N-1\}$. 
We follow the convention in~\cite{dzanic2020fourier, Durall_2020_CVPR} and compute the  azimuthally average of the magnitude of Fourier coefficients over radial frequencies to obtain the reduced spectrum and normalize it. The reduced spectra indicates the strength of the signal with respect to different spatial frequencies. Since our study focuses on high frequency Fourier attributes, we  pay attention to the last $25\%$ of the spatial frequencies (0.75 - 1.0 normalized spatial frequencies) similar to \cite{dzanic2020fourier}.

The Sampling Theorem (\cite{jain:dip_book}, chapter 4.2) states that: 
A bandlimited image $f(x,y)$ with bandwidths $\xi_{x0}, \xi_{y0}$ can be recovered without error from the sample values provided the
sampling rate is greater than the Nyquist rate: $2\xi_{x0}, 2\xi_{y0}$.
The image $f(x,y)$ with bandwidths $\xi_{x0}, \xi_{y0}$ means that
there is no frequency content outside a bounded region in the frequency plane defined by $\xi_{x0}, \xi_{y0}$. Details and mathematical proofs can be found in \cite{jain:dip_book} chapter 4.2.

\section{Last Upsampling Operation and Fourier Discrepancies}
Based on last section, the maximum frequency represented by a discrete 2D signal is constrained by the spatial resolution (sampling) of the signal. 
Previous work has consistently reported  discrepancies in the highest frequencies \cite{Durall_2020_CVPR, dzanic2020fourier}. Therefore, we hypothesize that inner generator layers that produce lower resolution outputs may not be directly responsible for the high frequency discrepancies. Therefore, we focus on the last upsampling step. In particular, we split the last step into 2 operations, 1) Feature map scaling and 2) Convolution. 

\subsection{Feature Map Scaling}
Feature map scaling is a non-parametric operation that scales the input in both dimensions by some factor (Usually 2 in most GAN architectures). In this work, we focus on 3 common feature map scaling techniques. \textbf{1) Zero-insert scaling} inserts zero between every row and column, scaling the input in both dimensions which is also used by transpose convolutions. \textbf{2) Nearest interpolation} scales the input by inserting nearest neighbour values. \textbf{3) Bilinear interpolation} scales the input by inserts new values by taking the weighted average of adjacent values.

\begin{figure}
    \centering
    \includegraphics[width=0.8\linewidth]{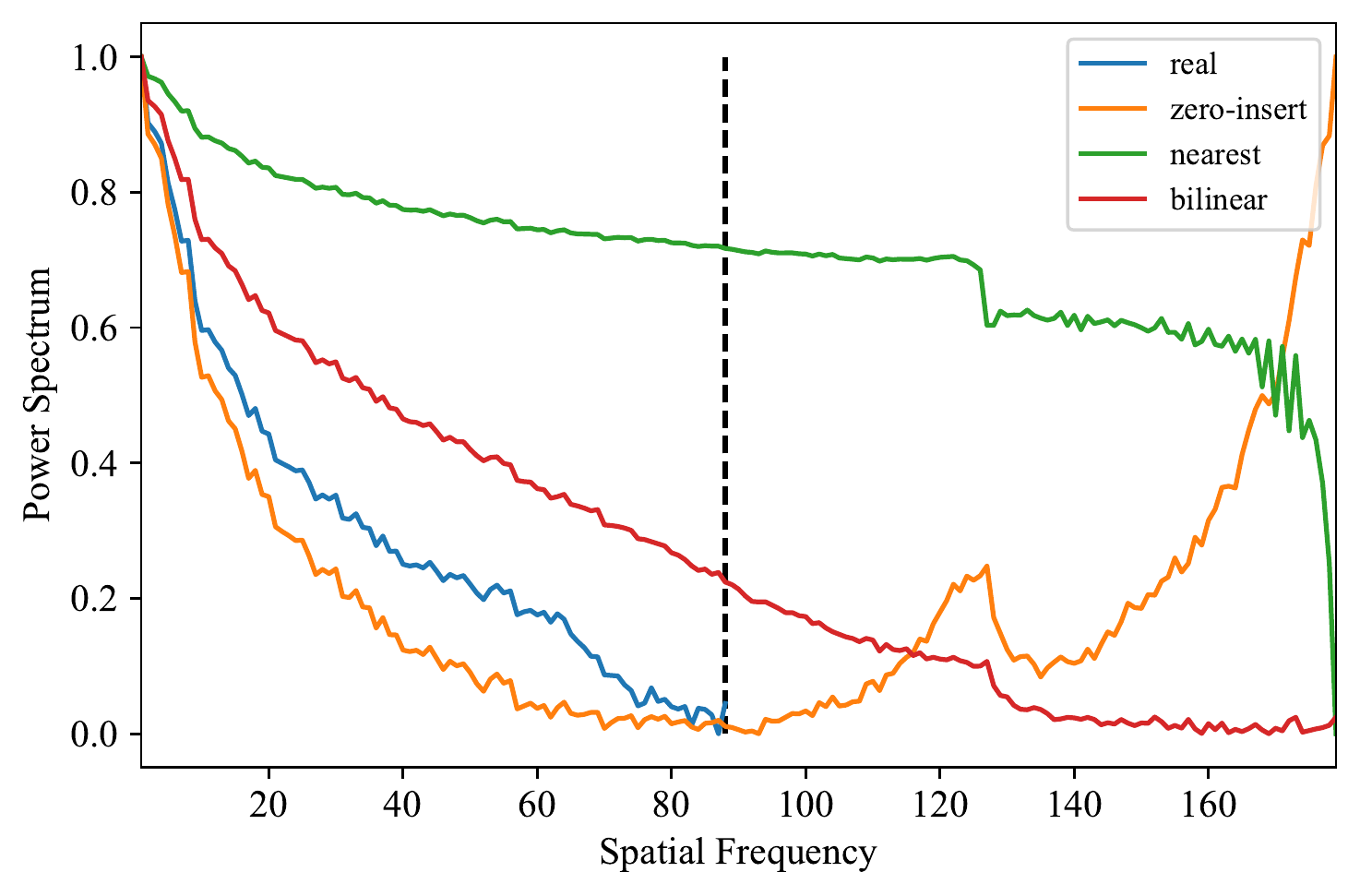}
    \caption{Example showing normalized spectral effects of upsampling an image. Vertical line at 88 shows the maximum radial frequency of reference image.}
    \label{fig:upsample_effects}
    \vspace{-0.2cm}
\end{figure}

From frequency perspective, zero-insertion introduces the largest amount of high frequency content as it replicates the low frequency spectrum for the high frequencies \cite{GAN_Artifacts}, followed by billinear and nearest interpolation. We focus more on the ``spectral trend'' than the frequency values, and show a schematic example of these upsampling effects on the normalized reduced spectra in Figure \ref{fig:upsample_effects} by upsampling a reference image of 128x128 from CelebA. \cite{liu2015faceattributes}

\subsection{Convolution}
The subsequent convolution operation learns kernels in order to satisfy the optimization objective. Convolutional kernels are capable of suppressing/ amplifying high frequencies. \eg A Gaussian kernel suppresses high frequencies and a bilateral kernel amplifies high frequencies. So when designing upsampling blocks in GANs, the general intuition is that irrespective of the feature map scaling method, the kernels will learn to manipulate the scaled feature maps to satisfy the objective function.

\section{Experiments}
Here we discuss the main experiments. Additional experiments and analysis can be found in Supplementary.

In order to investigate the effects of these 2 operations, we design a rigorous test bed that can address feature map scaling, kernel size and number of kernels independently. 
Our aim is to isolate the effects of these 2 operations and understand their roles (if any) in causing the high frequencies discrepancies.
We use celebA \cite{liu2015faceattributes} dataset at 128x128 resolution and use 3 GANs with identical architectures but different loss functions namely 1) DCGAN \cite{radford2016unsupervised}, 2) LSGAN \cite{lsgan} and 3) WGAN-GP \cite{NIPS2017_892c3b1c}. All baseline models consist of transpose convolutions with kernel size 4 identical to most out of the box GAN architectures including CycleGAN \cite{CycleGAN2017}, StarGAN \cite{choi2018stargan} and VQ-VAE \cite{NIPS2017_7a98af17}.

\textbf{Proposed Test Bed.}
Table~\ref{table:1} summarizes our test bed.
All baseline experiments using transpose convolutions of kernel size 4 are indicated by the experiment code Baseline and our handcrafted experiments are indicated using the 3 character code: An experiment code of X.Y.Z indicates X type of feature map scaling (Possible values are Z : Zero-insertion, N: Nearest interpolation, B: Bilinear interpolation), 
Y number of convolutional blocks and Z sized convolutional kernels for the last upsampling step. \eg A code of N.1.5 indicates nearest interpolation feature map scaling with a single convolutional block of 5x5 kernel for the last upsampling step. 
Note that we focus on up scaling  by a factor of 2 as used in most GAN models.
Our test bed summary is shown in table  \ref{table:1} and it contains experiments addressing the following factors:

{\bf Feature Map scaling.}
In order to investigate the effect of feature map scaling on high frequency Fourier attributes, we use experiments Z.1.5, N.1.5 and B.1.5. We explicitly conduct experiments using zero-insert feature map scaling as sanity check experiments to verify the effects of transpose convolutions. Do note that we had to use odd size kernel to maintain the spatial size after scaling, and hence we use kernels of size 5 for the last convolutional block.

{\bf Kernel Size.}
We use experiments x.1.3, x.1.5 and x.1.7 to investigate the kernel size effects on high frequency Fourier behaviour. Here x refers to 
Z, N or B for different types of scaling as discussed.

{\bf Number of kernels.}
Further, we use experiments x.1.5 and x.3.5 to study the effect of number of kernels on the high frequency Fourier behaviour. 

\begin{table}
\begin{center}
\begin{adjustbox}{width=\columnwidth,center}
\begin{tabular}{l|c}\toprule
\textbf{Code} &\textbf{Details} \\\midrule
Baseline &Transpose convolution (4x4 kernel) \\
\hline
N.1.5 &Nearest Upsampling + 1 conv block of 5x5 kernel \\
Z.1.5 &Zero insert Upsampling + 1 conv block of 5x5 kernel \\
B.1.5 &Bilinear Upsampling + 1 conv block of 5x5 kernel \\
\hline
N.1.3 &Nearest Upsampling + 1 conv block of 3x3 kernel \\
N.1.7 &Nearest Upsampling + 1 conv block of 7x7 kernel \\
\hline
Z.1.3 &Zero insert Upsampling + 1 conv block of 3x3 kernel \\
Z.1.7 &Zero insert Upsampling + 1 conv block of 7x7 kernel \\
\hline
B.1.3 &Bilinear Upsampling + 1 conv block of 3x3 kernel \\
B.1.7 &Bilinear Upsampling + 1 conv block of 7x7 kernel \\
\hline
N.3.5 &Nearest Upsampling + 3 conv blocks of 5x5 kernel \\
Z.3.5 &Zero insert upsampling + 3 conv blocks of 5x5 kernel \\
B.3.5 &Bilinear upsampling + 3 conv blocks of 5x5 kernel \\
\bottomrule
\end{tabular}
\end{adjustbox}
\end{center}
\caption{Test Bed to study the effect of feature map scaling and convolution in the {\em last} upsampling step of the generator, for different GANs: DCGAN, LSGAN, WGAN-GP, StarGAN.
``Baseline'' refers to public released code of the GAN model, which uses transpose convolution of $4\times4$ kernel. For other models, we replace the last transpose convolution in Baseline with the corresponding configurations shown.
We emphasize that we only modify the last step specified as above; the algorithms, learning objectives and architectures (except the last step) are kept identical as the public released code for different GAN models.
}

\label{table:1}
\vspace{-0.2cm}
\end{table}

\section{Metrics}
Spectral behaviour of synthetic images have been analysed only qualitatively, not quantitatively in previous works \cite{dzanic2020fourier, Durall_2020_CVPR, durall2020unmasking}. Defining a spectral consistency metric is non-trivial and to be consistent, we will qualitatively analyse the high frequency spectral distribution in our experiments. Synthetic images are spectral consistent if they demonstrate power spectrum decay behaviour similar to their training data, and if not, vice versa. We use 4000 real and GAN images to generate spectral distribution curves. To ensure that our setups were trained properly, we use FID scores to assess the quality of samples and make sure that they are consistent with the FID scores from baseline experiments.

\section{Results}
\subsection{Effect of Feature Map Scaling Methods}
\begin{figure*}
\begin{tabular}{ccc}
      \multicolumn{3}{c}{\includegraphics[width=0.9\linewidth]{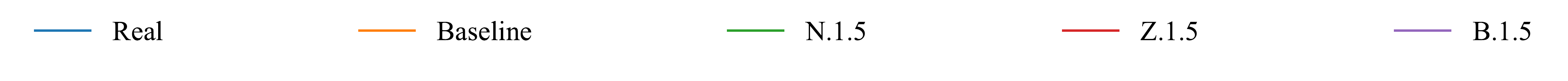}}\\
      \includegraphics[width=0.3\linewidth]{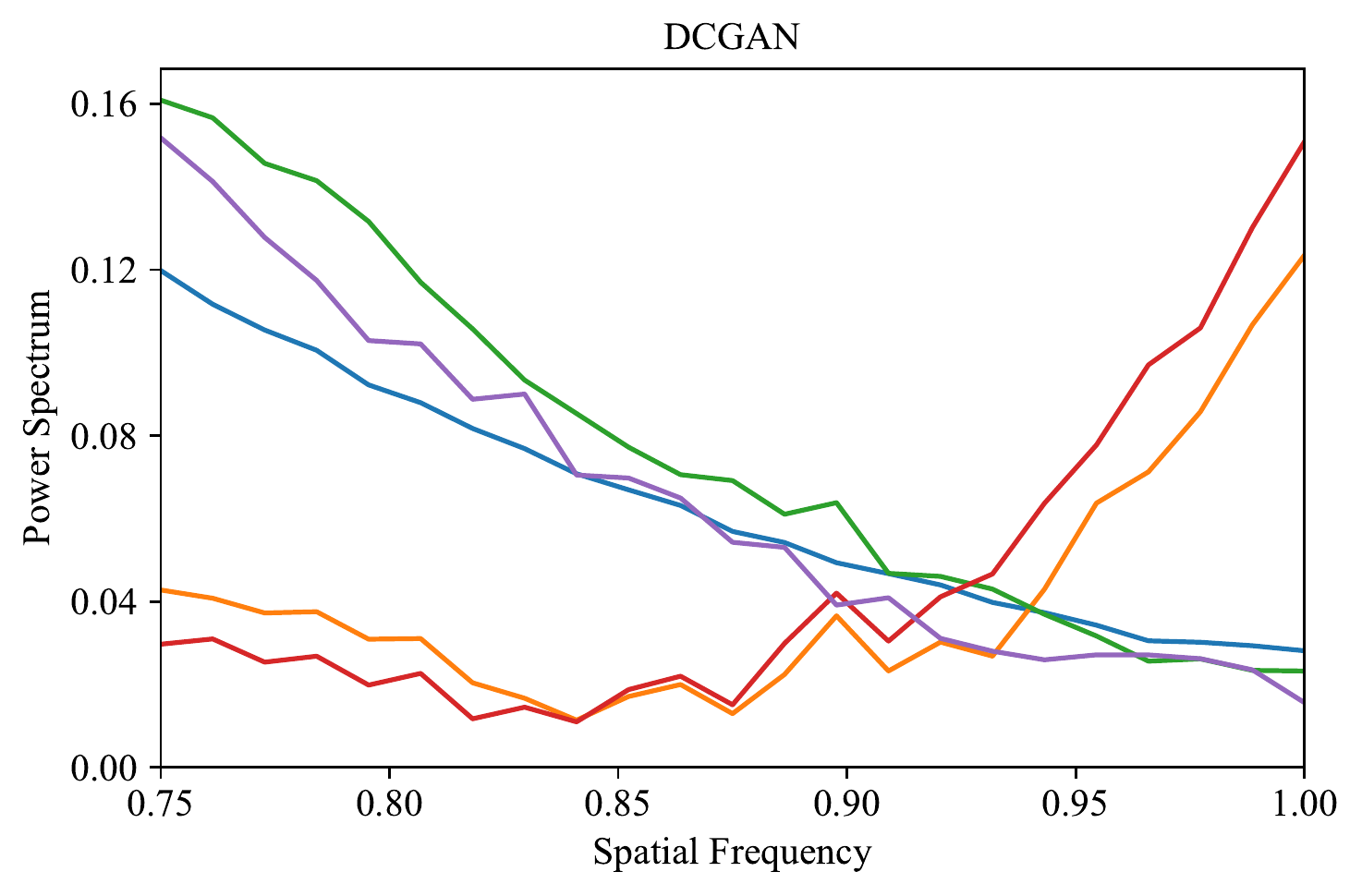} &   
      \includegraphics[width=0.3\linewidth]{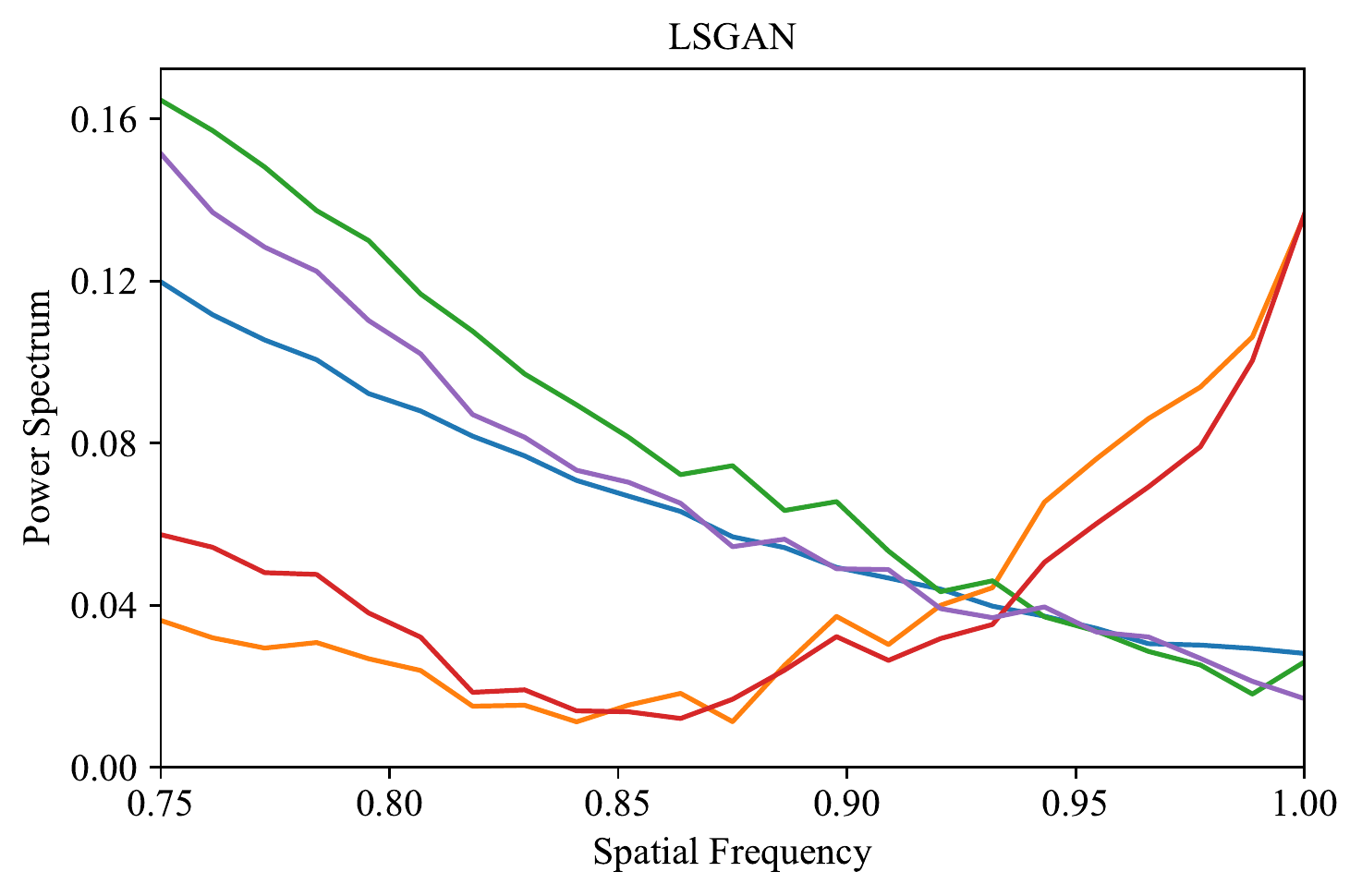} &
      \includegraphics[width=0.3\linewidth]{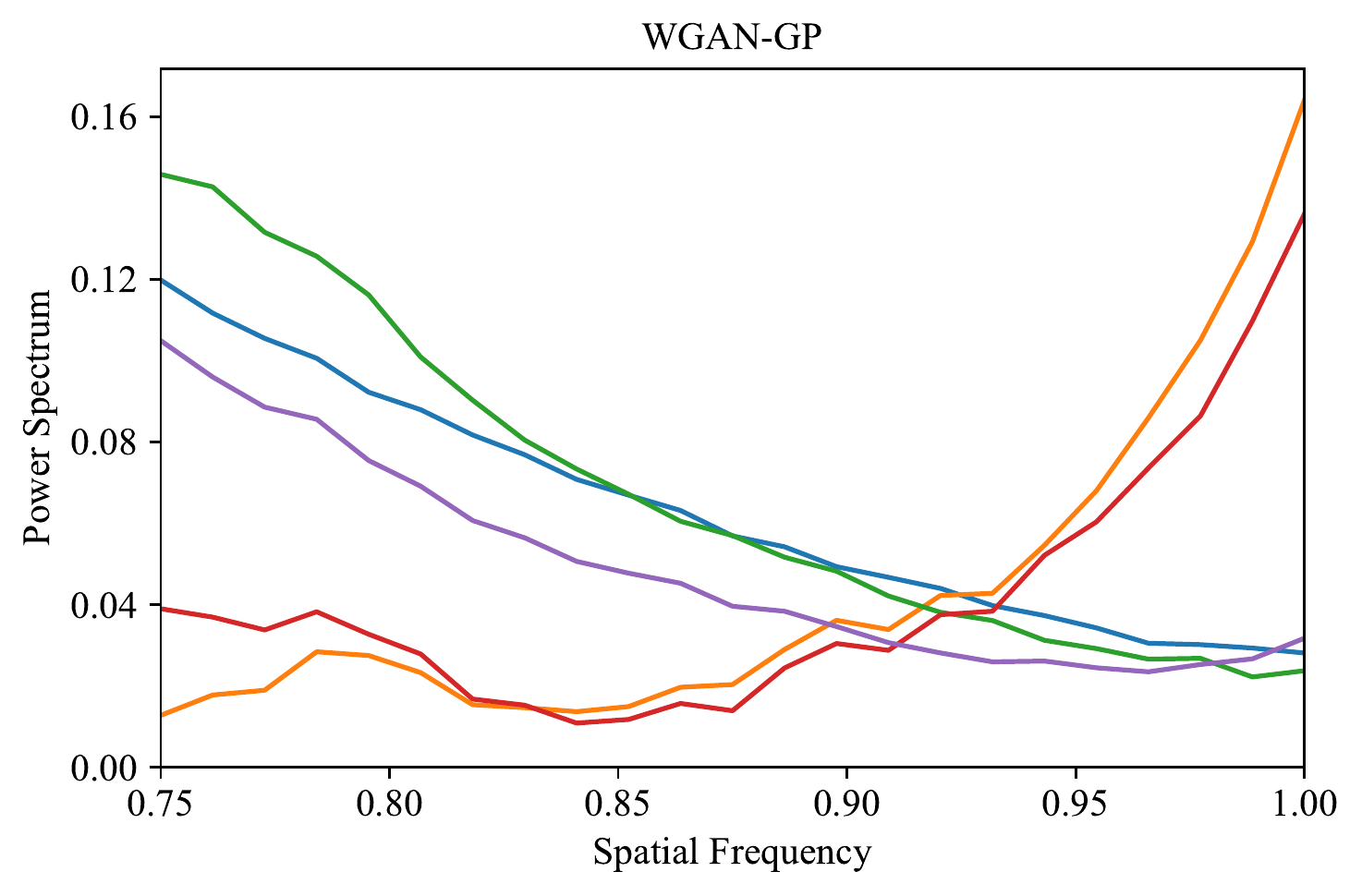}\\
\end{tabular}
\caption{Feature Map Scaling Results. We observe that experiments using nearest and bilinear interpolation methods 
in the last step
are able to produce spectral consistent GANs. Refer to table \ref{table:1} for experiment details.}
\label{fig:celeba1}
\vspace{-0.2cm}
\end{figure*}

Figure \ref{fig:celeba1} illustrates the resulting spectral distributions when using different feature map scaling methods. We observe that images from Z.1.5 and Baseline experiments are spectral inconsistent for all 3 GANs.
Nearest and bilinear interpolation methods are able to replicate the spectral distribution of real data reasonably across all 3 GAN models. 
Since the only change in our models is the feature map scaling method in the last layer while the algorithms, objective functions and majority of the network architectures are identical to public released code,
these results qualify to support our thesis that high frequency Fourier discrepancies are not inherent to GANs.

\subsection{Effect of Kernel Size}
\begin{figure*}
\begin{tabular}{ccc}
      \multicolumn{3}{c}{\includegraphics[width=0.9\linewidth]{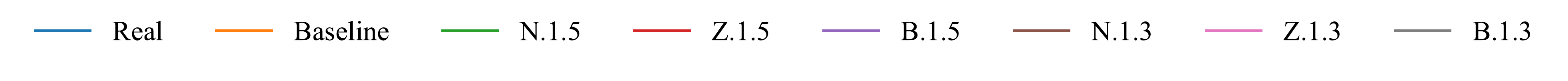}}\\
      \includegraphics[width=0.3\linewidth]{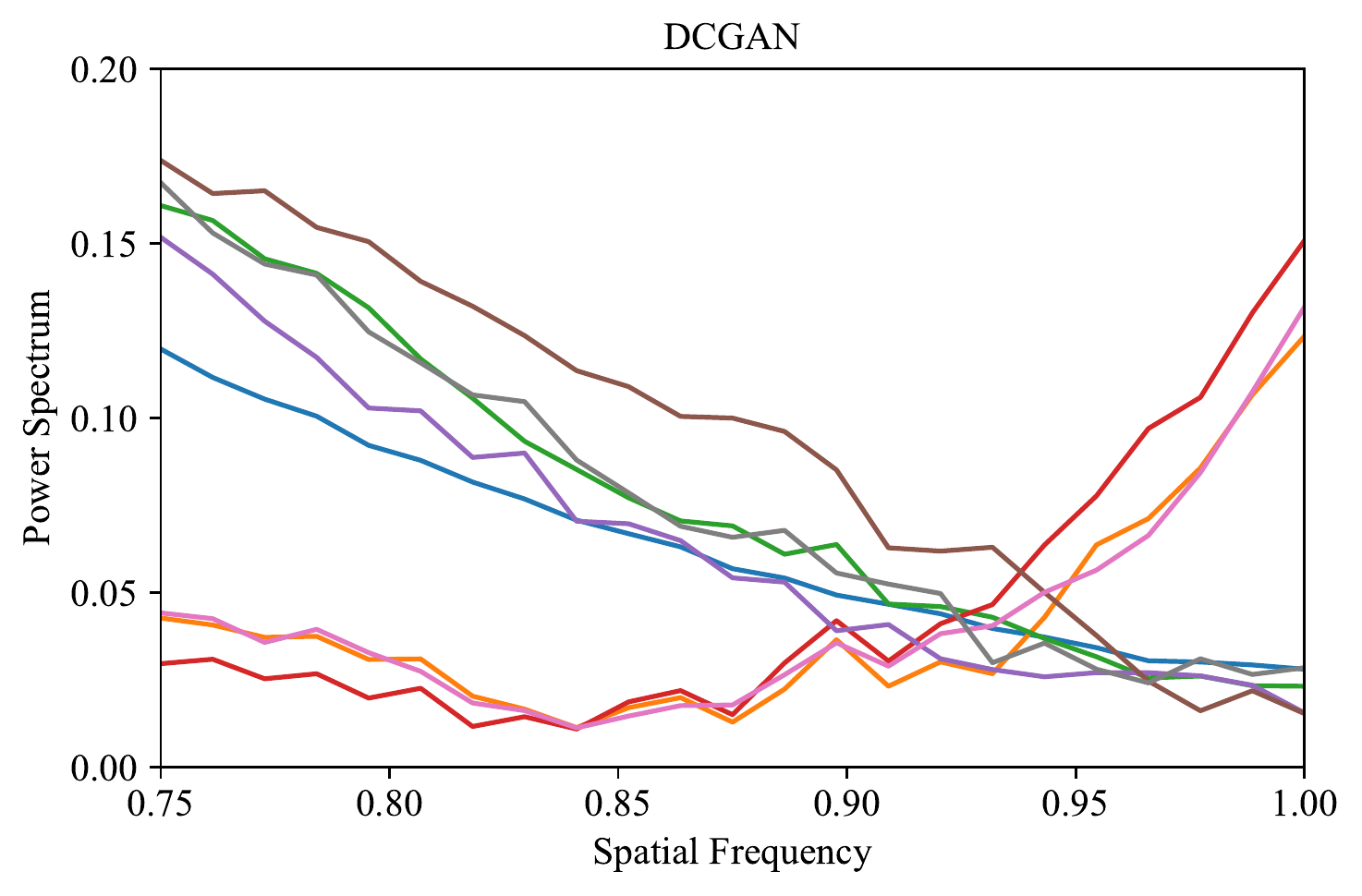} &   
      \includegraphics[width=0.3\linewidth]{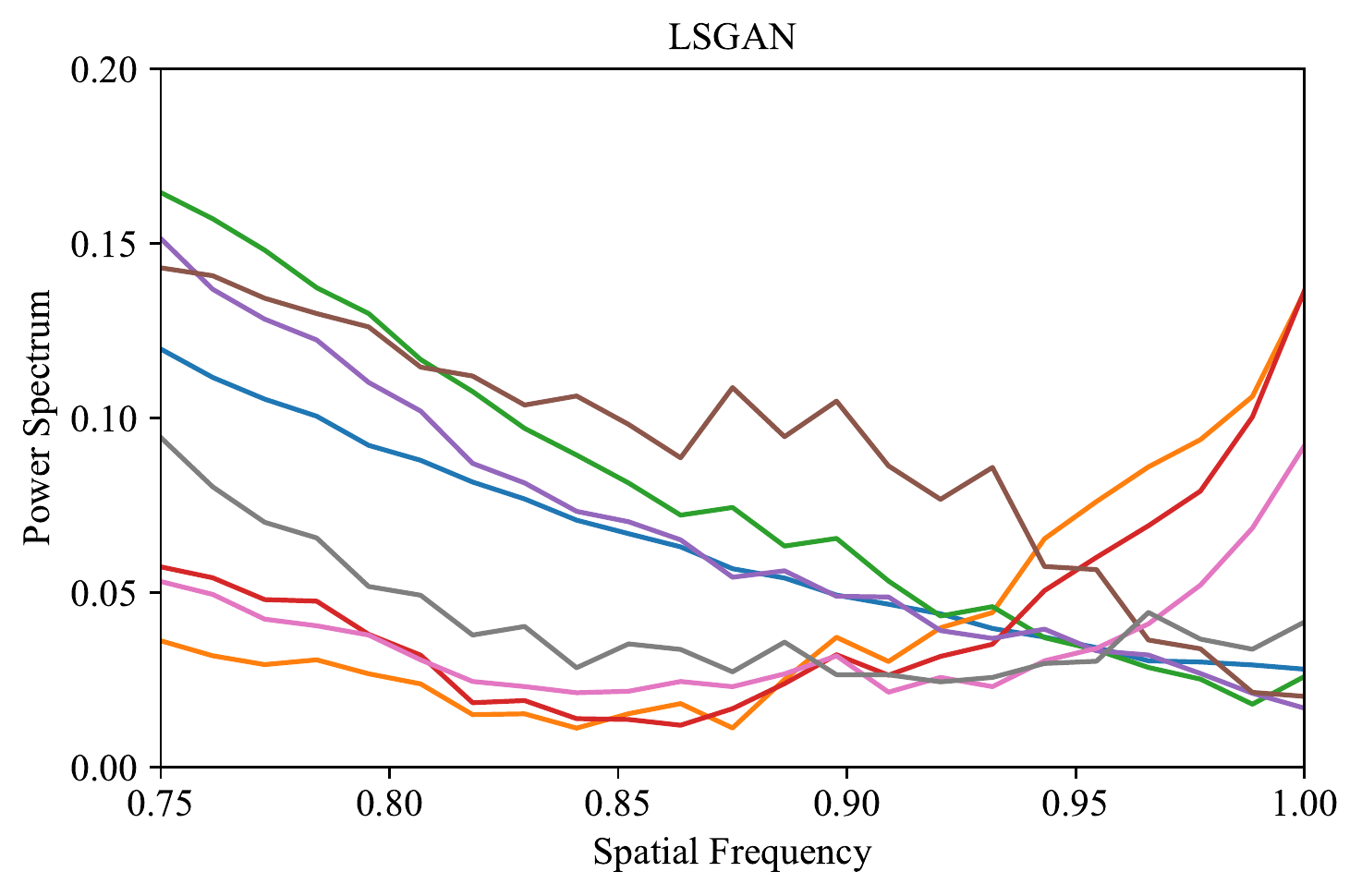} &
      \includegraphics[width=0.3\linewidth]{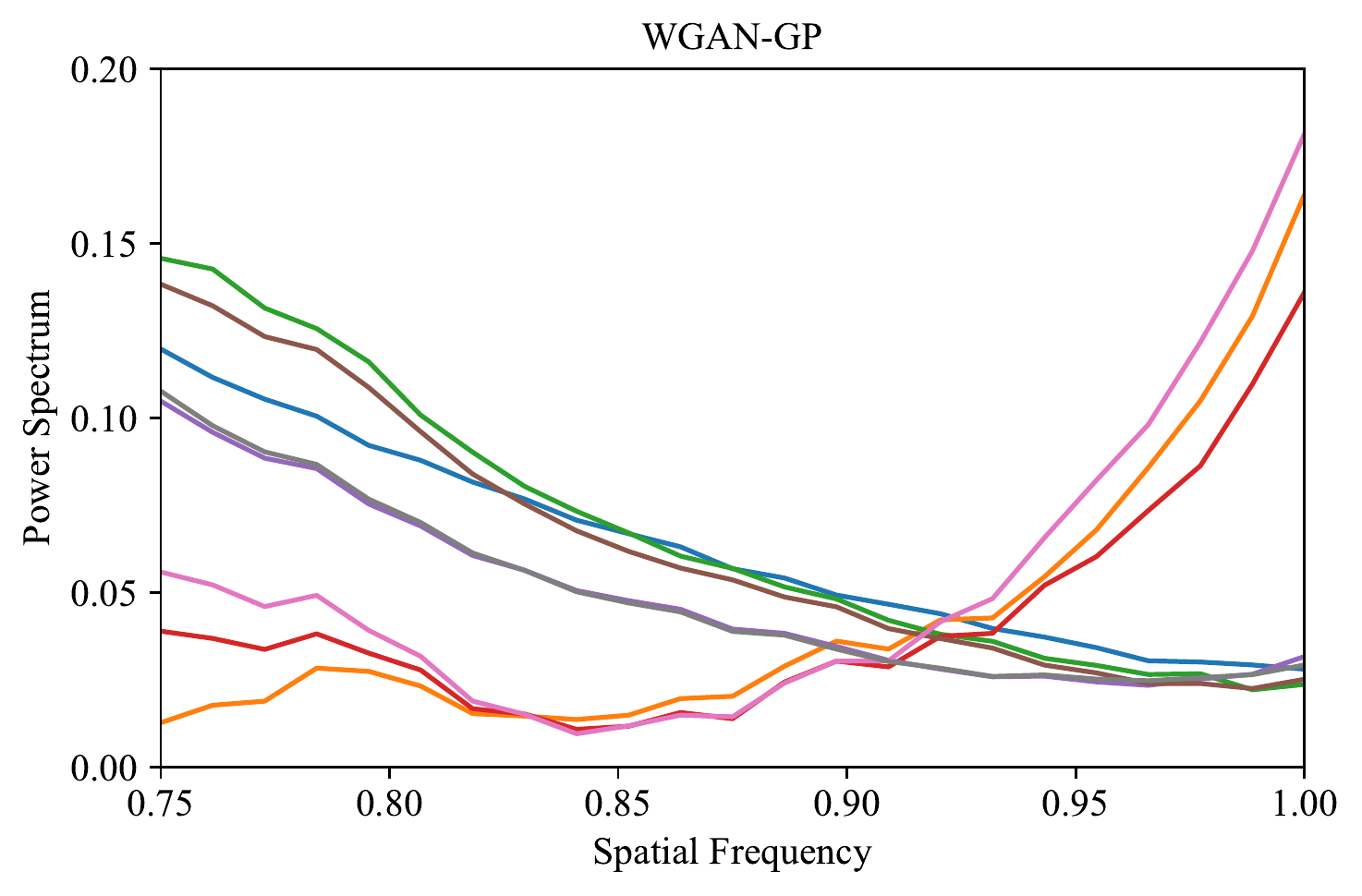}\\
\end{tabular}
\caption{Smaller Kernel (3x3) Results. We observe that smaller kernels do not substantially deteriorate spectral consistent GANs except for some turbulent behaviour observed in LSGAN for N.1.3 and B.1.3 experiments. Refer to table \ref{table:1} for experiment details.}
\label{fig:celeba2}
\vspace{-0.2cm}
\end{figure*}

\begin{figure*}
\begin{tabular}{ccc}
      \multicolumn{3}{c}{\includegraphics[width=0.9\linewidth]{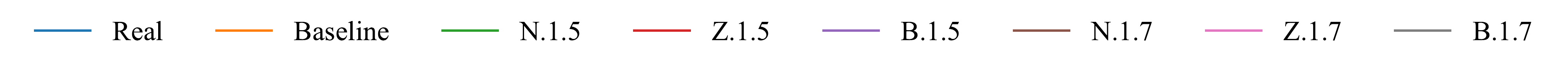}}\\
      \includegraphics[width=0.3\linewidth]{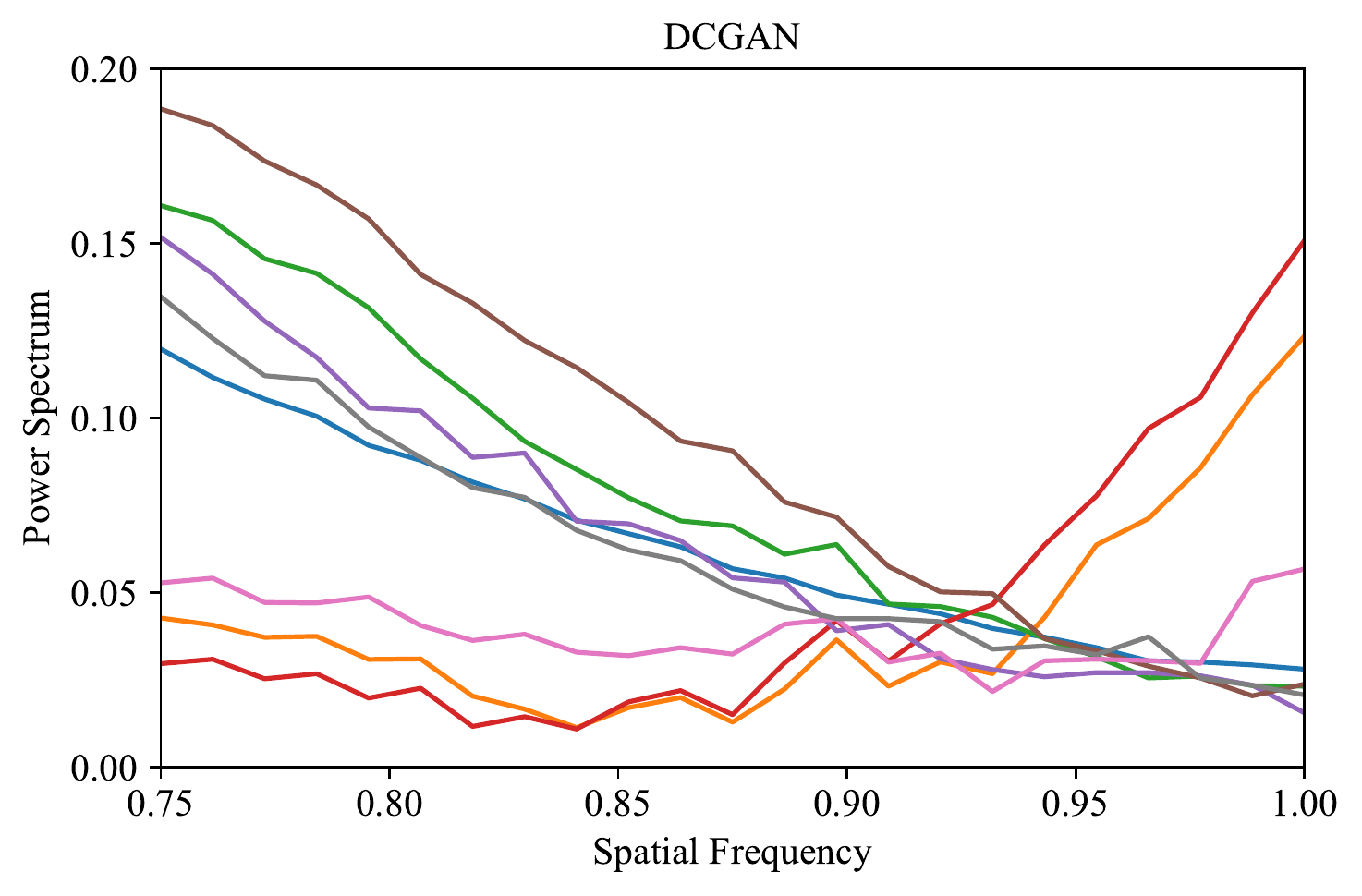} &   
      \includegraphics[width=0.3\linewidth]{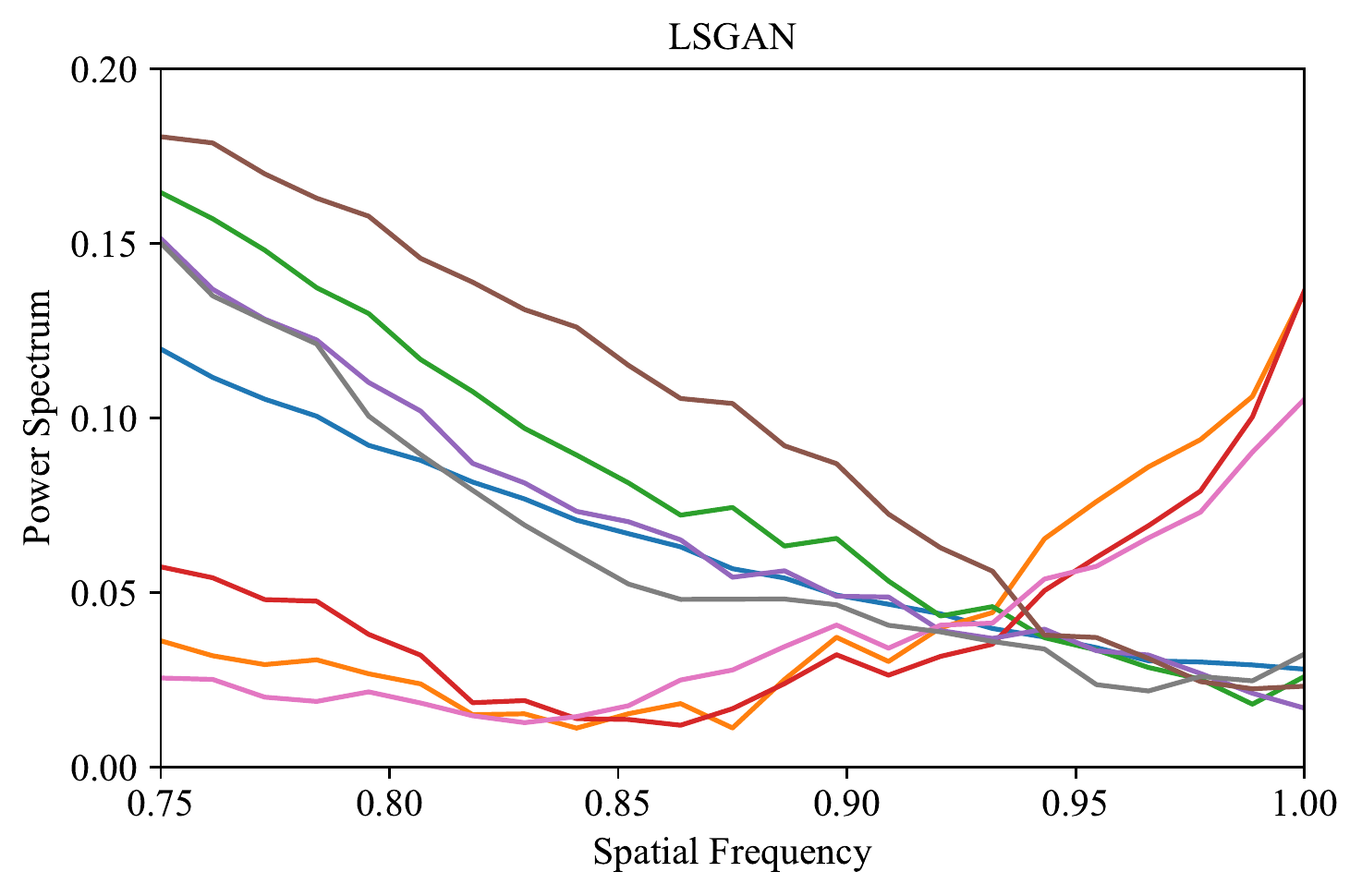} &
      \includegraphics[width=0.3\linewidth]{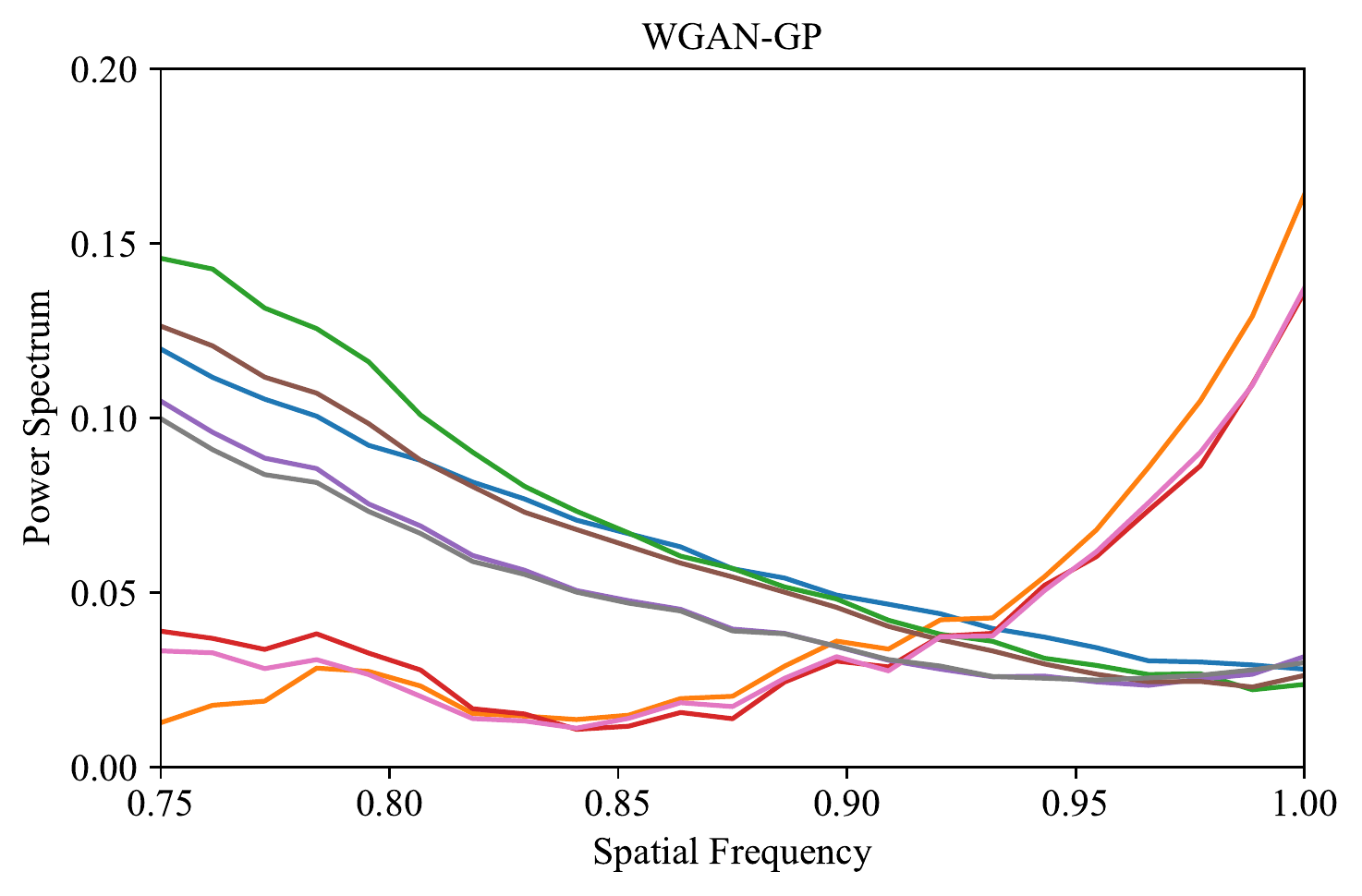}\\
\end{tabular}
\caption{Larger Kernel (7x7) Results. We observe that larger kernels do not substantially manipulate the discrepancies in Z.1.7 experiments. Refer to table \ref{table:1} for experiment details.}
\label{fig:celeba3}
\vspace{-0.2cm}
\end{figure*}

From Figure \ref{fig:celeba2}, we observe that smaller kernel sizes result in more turbulent spectral behaviour for N.1.3 and B.1.3 LSGAN experiments. Apart from this observation, experiments using nearest and bilinear interpolation methods are able to reproduce spectral distributions of real data reasonably well, and zero-insertion based methods always result in high frequency spectral distortions. Further, from Figure \ref{fig:celeba3}, we observe that when using larger kernels, all experiments using nearest and bilinear interpolation methods are able to replicate spectral behaviour of real data, and even with larger kernels zero-insertion based methods produce high frequency spectral distortions (Z.1.7). 

\subsection{Effect of Number of Kernels}
\begin{figure*}
\begin{tabular}{ccc}
      \multicolumn{3}{c}{\includegraphics[width=0.9\linewidth]{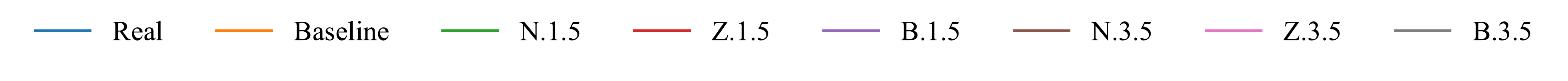}}\\
      \includegraphics[width=0.3\linewidth]{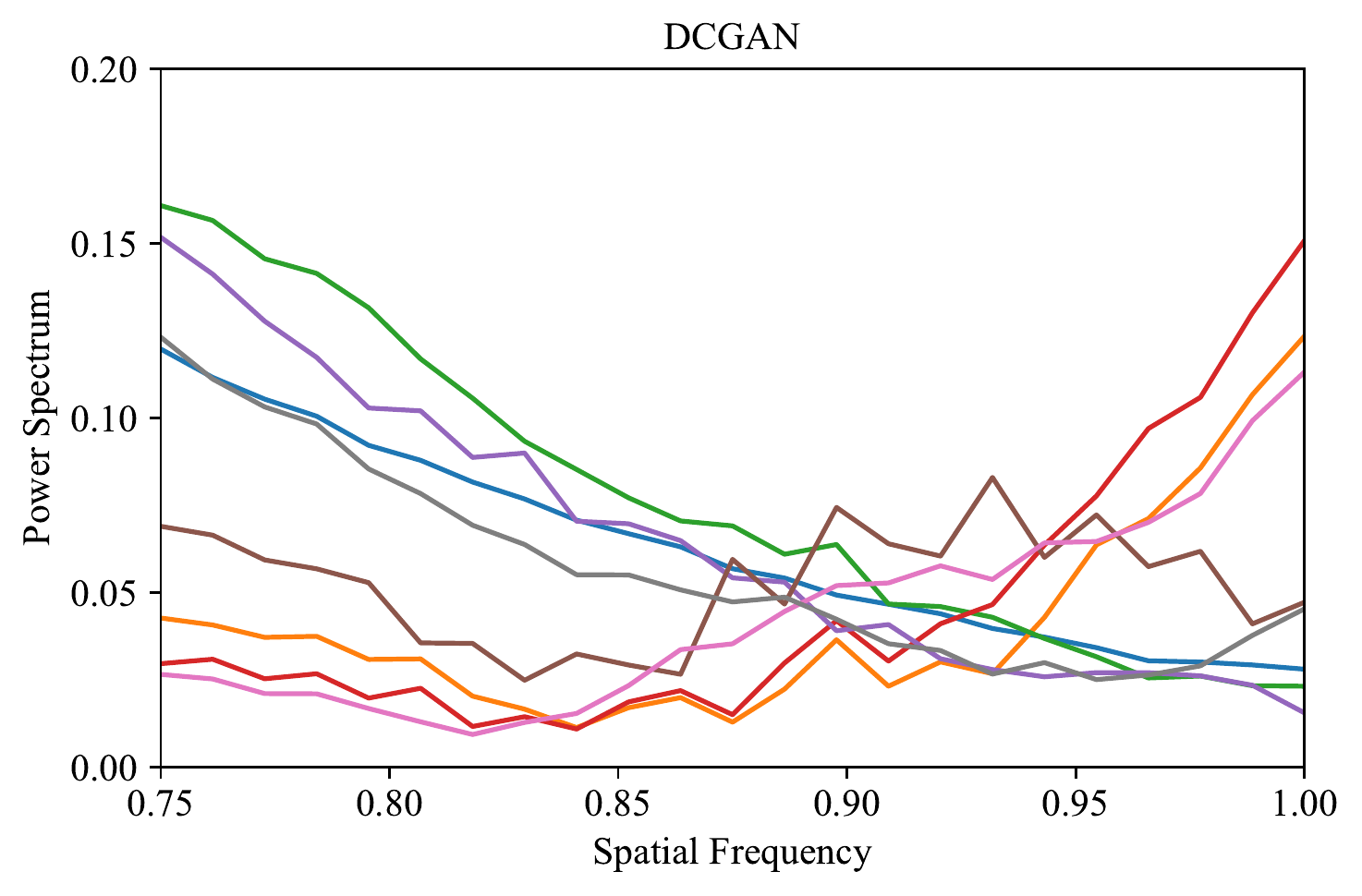} &   
      \includegraphics[width=0.3\linewidth]{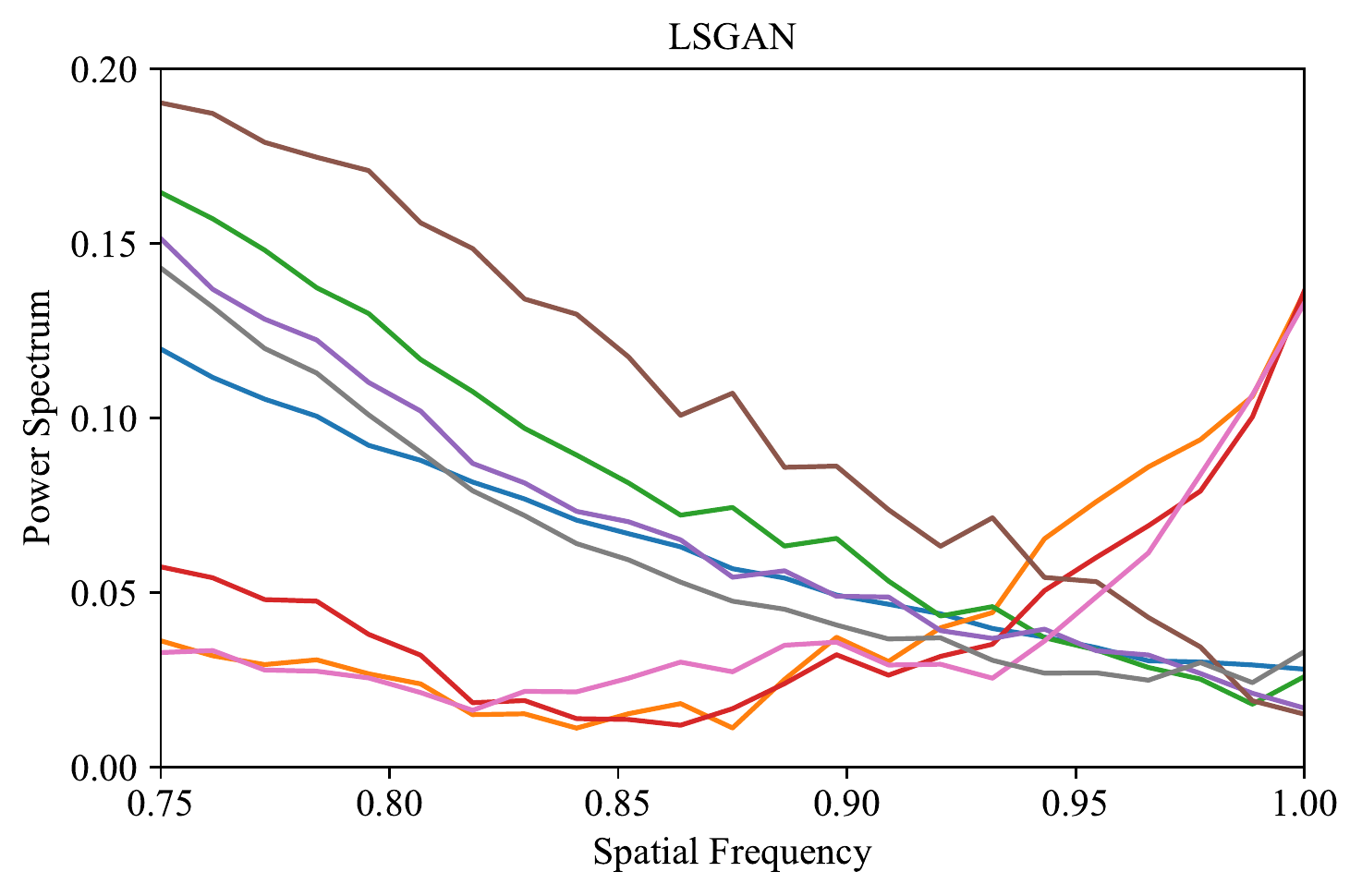} &
      \includegraphics[width=0.3\linewidth]{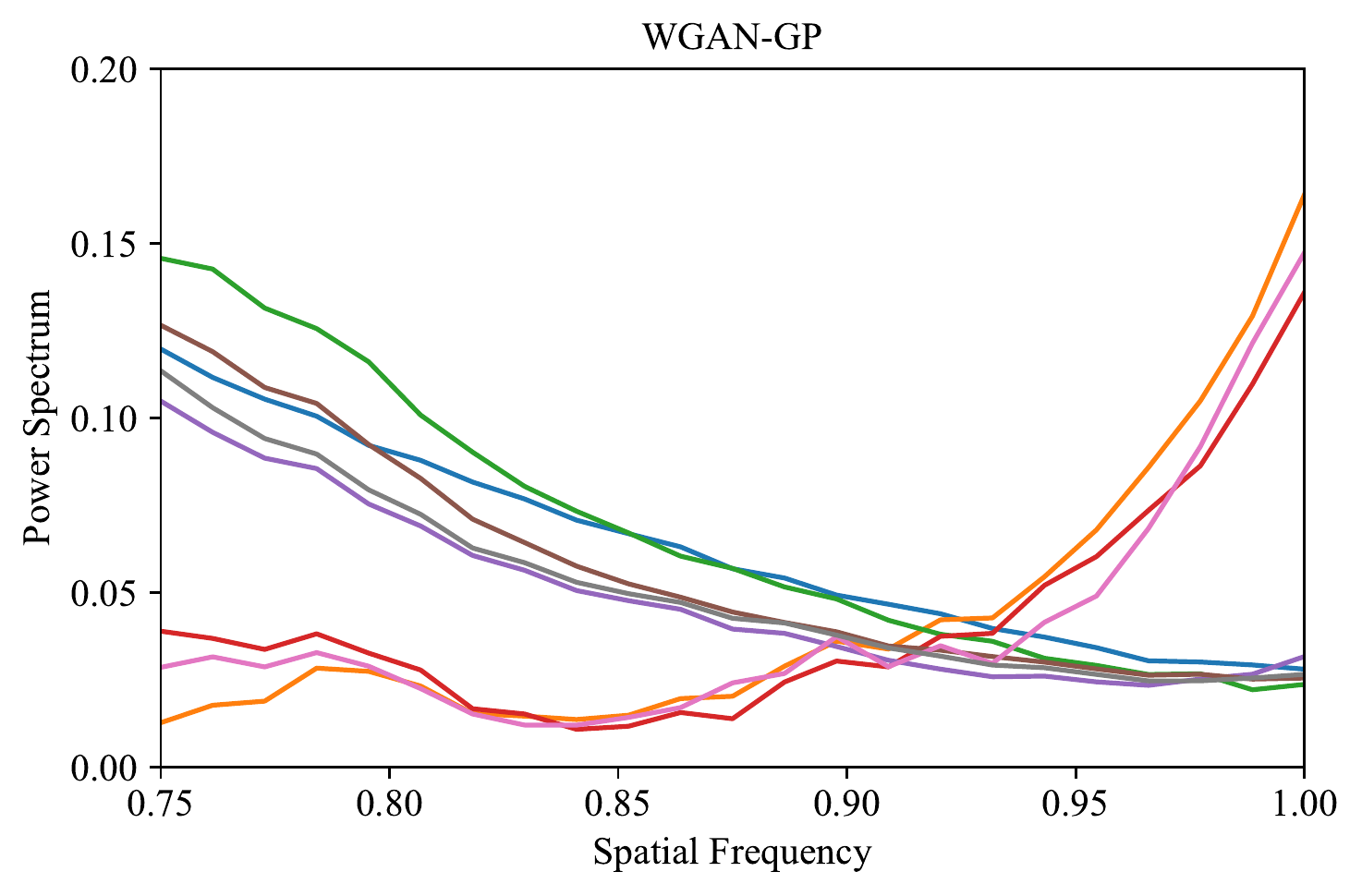}\\
\end{tabular}
\caption{Increased number of kernels (3 conv blocks) Results. We see that even with more number of kernels in the last upsampling step, Z.3.5 experiment is not able to produce spectral consistent GANs. Refer to table \ref{table:1} for experiment details.}
\label{fig:celeba4}
\vspace{-0.1cm}
\end{figure*}

Figure \ref{fig:celeba4} illustrates that increasing the number of kernels do not yield spectral consistency by itself. Apart from  N.3.5 DCGAN experiment, we observe that all other experiments using nearest and bilinear interpolation methods are able to approximate the spectral behaviour of real data. Also, GAN objective functions do not impose any spectral requirements. Thus the kernels do not have any direct incentive for imposing spectral consistency. 

\begin{figure*}[h]
\centering
\begin{tabular}{ccc}
    \includegraphics[width=0.3\linewidth]{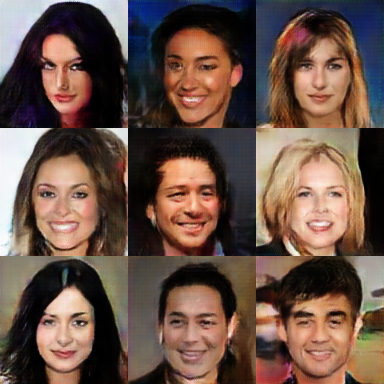} &   
      \includegraphics[width=0.3\linewidth]{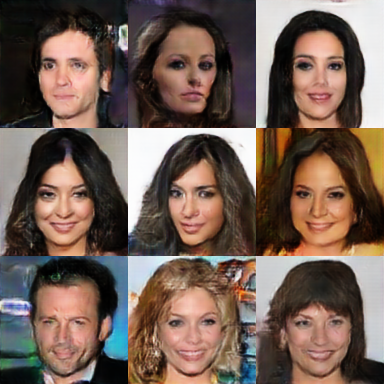} &
      \includegraphics[width=0.3\linewidth]{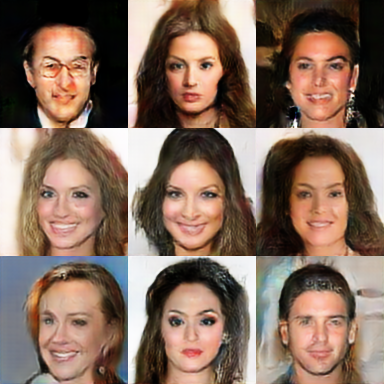}\\
\end{tabular}
\caption{WGAN-GP samples for Baseline (Left), N.1.5 (Middle) and B.1.5 (Right) for CelebA \cite{liu2015faceattributes}. We observe that the visual quality is comparable when replacing the {\em last} transpose convolutions with nearest and bilinear methods. More visual results in Supplementary \ref{sec_sup:image_samples}.}
\label{fig:gan_samples}
\end{figure*}

\textbf{Key observations.}
Throughout all 39 rigorous experiments, we observe that zero insertion based feature map scaling methods (including Baseline) are consistently showing high frequency spectral discrepancies, and most experiments (22/24) using bilinear and nearest interpolation methods are able to avoid these high frequency spectral discrepancies. All these results support our statement that high frequency spectral discrepancies are not inherent characteristics to GANs. The FID scores for all experiments were comparable with the baseline FID, and are included in the Supplementary \ref{sec_sup:fid}. We show image samples from WGAN-GP for Baseline, N.1.5 and B.1.5 setups in Figure \ref{fig:gan_samples}. Do note that for all experiments, we use the exact same discriminator architecture as the Baseline experiment.

\section{Fourier Synthetic Image Detector}
Dzanic \etal\cite{dzanic2020fourier} state that the spectral properties of real and deep network generated images are fundamentally different, and proposed a synthetic image detection method using a ``simple'' k-nearest neighbours (KNN) classifier to emphasize the extent of these spectral differences.  We follow the exact procedure as the original authors to train these classifiers and details can be found in Supplementary \ref{sec_sup:dzanic_classifier_details}.

\subsection{Experiment Setup}
With nearest and bilinear interpolation methods obtaining spectral consistent GANs for the previous experiments, we question whether the classifier proposed above would be robust enough to detect these samples as fake. To investigate this we follow the following steps:

\begin{enumerate}
    \item We train 3 KNN classifiers, one for DCGAN, LSGAN and WGAN-GP respectively. For GAN images, we use images generated from the Baseline experiment (using transpose convolutions) as training data.
    \item We test these classifiers using GAN images generated from the setups in our test bed to evaluate the robustness of the classifier.
    \item We also repeat the experiments using 50\% data for training the classifier (The original work used only 10\%) to observe any improvements in accuracy.
\end{enumerate}

\subsection{Detection Results}
The complete detection results are shown in table \ref{table:2}. We observe that all setups corresponding to N.x.x and B.x.x experiments are able to easily bypass the classifier. Even when 50\% training data is used, we are able to bypass the classifier with ease (Included in Supplementary \ref{sec_sup:additional_results}). The results clearly demonstrate that the proposed classifier relying on high frequency Fourier attributes to detect synthetic images, fails to detect images generated from identical GAN models with last feature map scaling replaced by nearest or bilinear interpolation methods. These results are consistent with the observed spectral distributions. By combining these detection results with the empirical finding that high frequency spectral discrepancies are not inherent characteristics of CNN-generated images, we suggest re-thinking of using such discrepancies to detect synthetic images.

\begin{table}
\begin{center}
\begin{adjustbox}{width=\columnwidth,center}
\begin{tabular}{l|ccc}\toprule
\textbf{Setup} &\textbf{DCGAN} &\textbf{LSGAN} &\textbf{WGAN-GP}\\
\hline
N.1.5 &\boldmath{$0.09\pm0.03$\%} &\boldmath{$0.34\pm0.08$\%} &\boldmath{$0.14\pm0.05$\%} \\
Z.1.5 &$84.82\pm3.72$\% &$88.16\pm3.98$\% &$99.75\pm0.14$\% \\
B.1.5 &\boldmath{$0\pm0$\%} &\boldmath{$0.1\pm0$\%} &\boldmath{$0.2\pm0.12$\%} \\
\hline
N.1.3 &\boldmath{$0\pm0$\%} &\boldmath{$0.06\pm0.05$\%} &\boldmath{$0.24\pm0.13$\%} \\
N.1.7 &\boldmath{$0\pm0$\%} &\boldmath{$0\pm0$\%} &\boldmath{$0.06\pm0.05$\%} \\
\hline
Z.1.3 &$98.73\pm0.56$\% &$73.09\pm3.5$\% &$97.94\pm0.87$\% \\
Z.1.7 &$97.23\pm1.1$\% &$95.66\pm1.93$\% &$99.94\pm0.07$\% \\
\hline
B.1.3 &\boldmath{$0\pm0$\%} &\boldmath{$0.19\pm0.1$\%} &\boldmath{$0.07\pm0.05$\%} \\
B.1.7 &\boldmath{$0\pm0$\%} &\boldmath{$0.1\pm0$\%} &\boldmath{$0.17\pm0.13$\%} \\
\hline
N.3.5 &\boldmath{$0.16\pm0.05$\%} &\boldmath{$0\pm0$\%} &\boldmath{$0\pm0$\%} \\
Z.3.5 &$77.67\pm6$\% &$67.66\pm11.9$\% &$99.9\pm0.19$\% \\
B.3.5 &\boldmath{$0.03\pm0.05$\%} &\boldmath{$0.48\pm0.04$\%} &\boldmath{$0.13\pm0.05$\%} \\
\bottomrule
\end{tabular}
\end{adjustbox}
\end{center}
\caption{
Detection results for the detectors proposed by
Dzanic \etal\cite{dzanic2020fourier}, using CelebA dataset.
We follow exactly the procedure in~\cite{dzanic2020fourier} to train the detector for each
GAN model (10\% data for training).
Then, the images generated by GAN models using different 
Setups are tested on the corresponding detectors. The table shows the successful detection rates, and we highlight the cases when the detection rates  are inferior (less than $10\%$).
The results consistently show that when a GAN model is trained with the
last feature scaling method based on nearest or bilinear, a detector trained using high frequency features
such as~\cite{dzanic2020fourier}
fails to detect GAN images (Consistent with spectral plot observations). All reported detection rates are averaged over 10 independent runs
}
\label{table:2}
\vspace{-0.3cm}
\end{table}

\section{Extended experiments}
With observations that high frequency Fourier spectrum discrepancies are not intrinsic characteristics of GANs, we extend our experiments to address different dataset, image resolution and GAN objective function to further find evidences to support our thesis statement. We select 3 setups from our test bed Z.1.5, N.1.5 and B.1.5 to conduct extended experiments since we have observed that kernel size/ number of kernels do not substantially manipulate high frequencies compared to feature map scaling methods. Similar to previous experiments, we analyze the resulting spectral distributions and evaluate the robustness of the synthetic detector proposed by Dzanic \etal \cite{dzanic2020fourier}.

\subsection{LSUN Bedrooms Dataset}
\begin{figure*}[ht]
\begin{tabular}{ccc}
      \multicolumn{3}{c}{\includegraphics[width=0.9\linewidth]{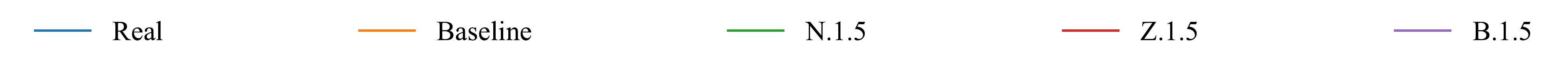}}\\
      \includegraphics[width=0.3\linewidth]{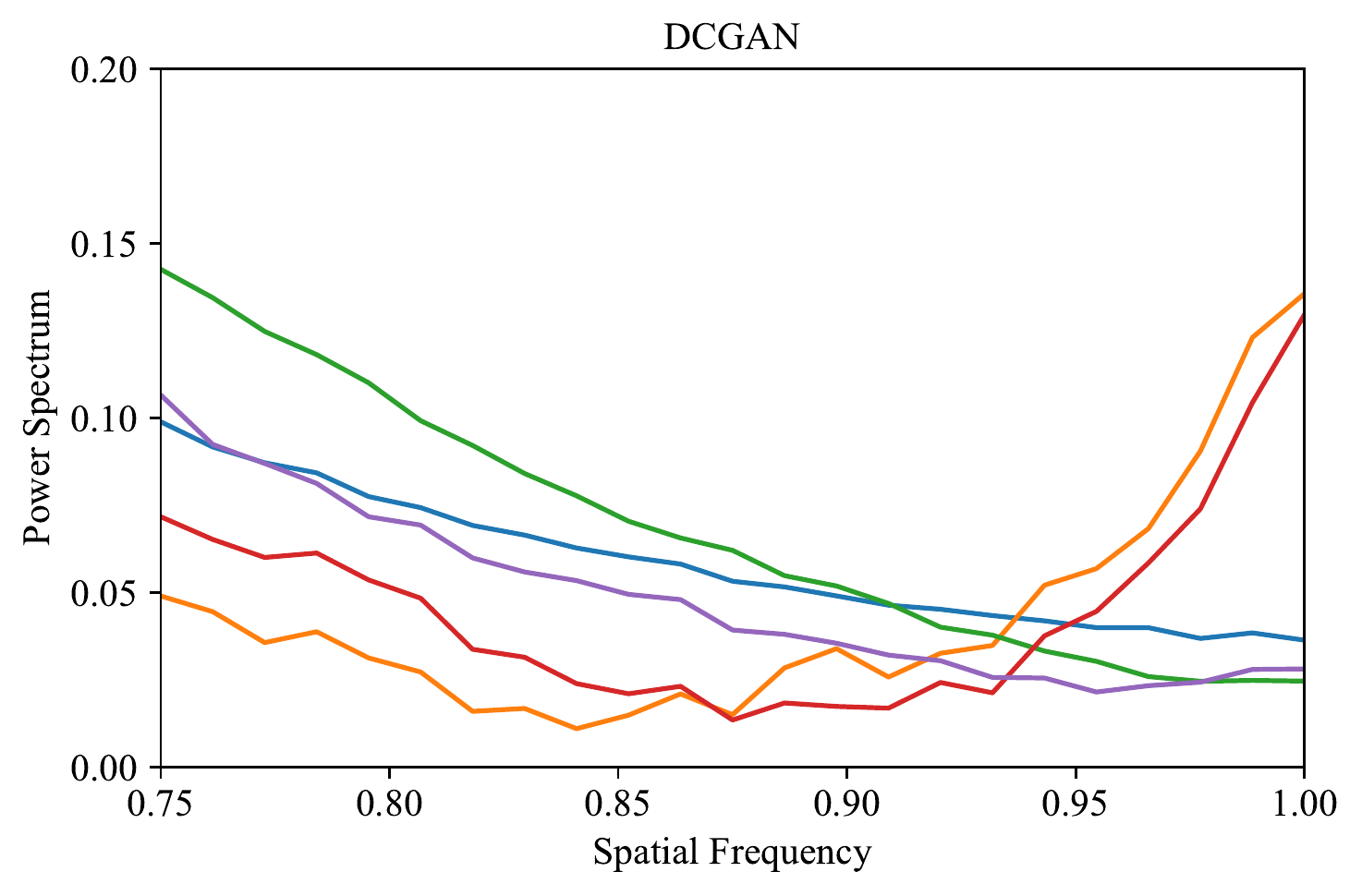} &   
      \includegraphics[width=0.3\linewidth]{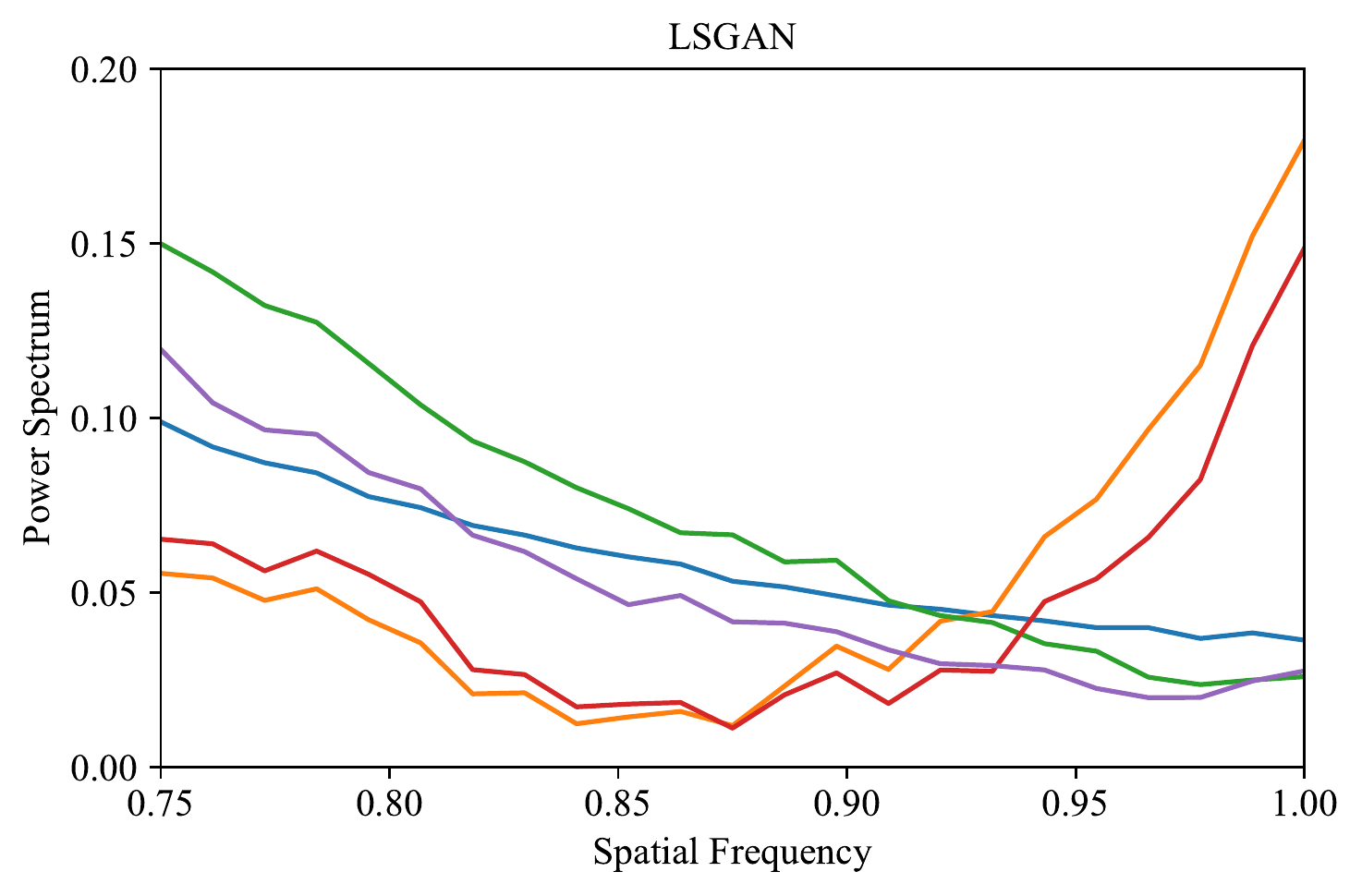} &
      \includegraphics[width=0.3\linewidth]{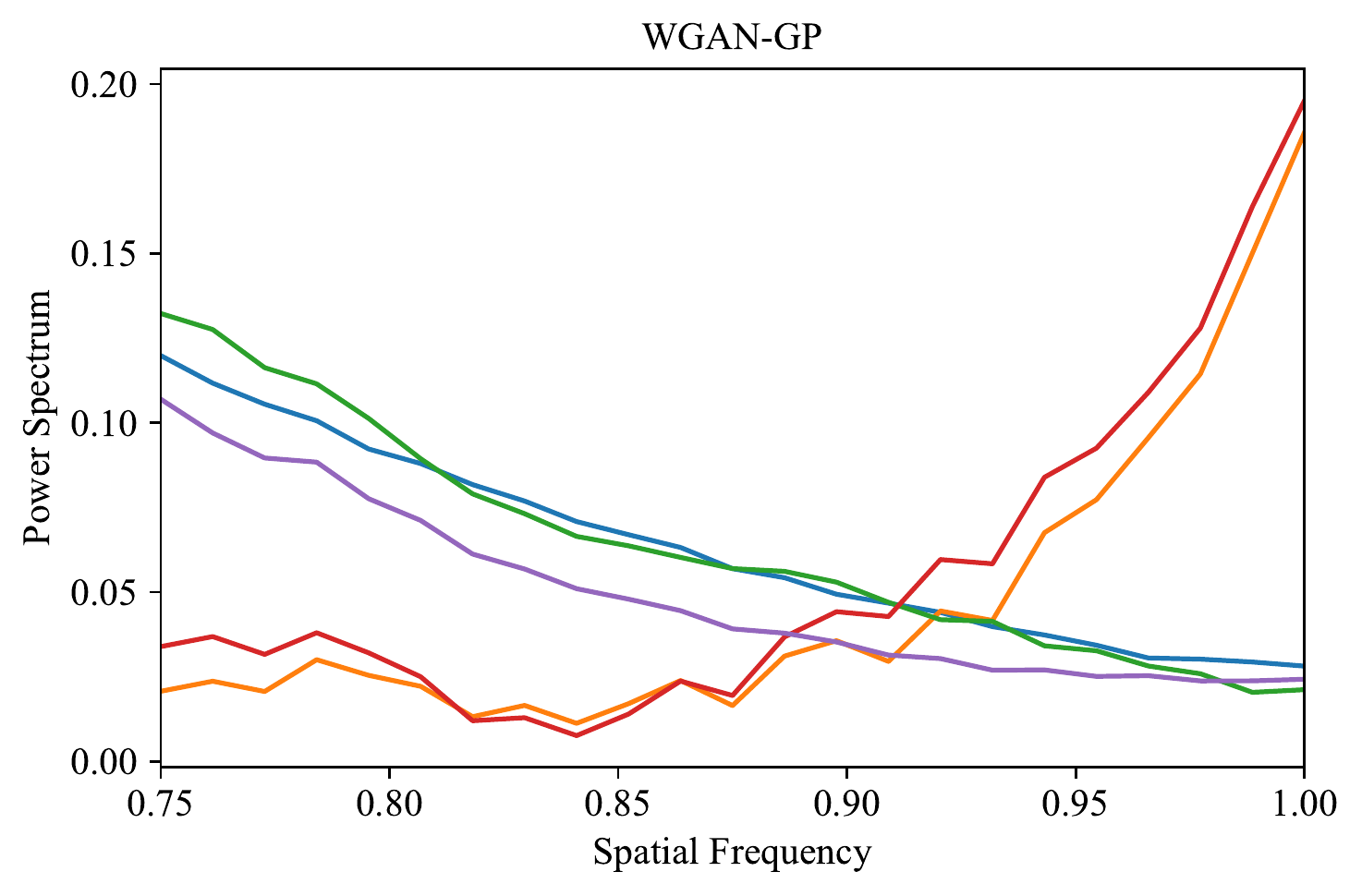}\\
\end{tabular}
\caption{LSUN results. We observe spectral plots identical to CelebA \cite{liu2015faceattributes} experiments. Refer to table \ref{table:1} for experiment details}
\label{fig:lsun}
\vspace{0.1cm}
\end{figure*}

In this experiment we use a subset of LSUN Bedrooms Dataset \cite{yu15lsun} (128x128) to train DCGAN \cite{radford2016unsupervised}, LSGAN \cite{lsgan} and WGAN-GP \cite {NIPS2017_892c3b1c} identical to previous setups. The spectral plots are shown in Figure \ref{fig:lsun}. We observe identical results to CelebA experiment (Figure \ref{fig:celeba1}). That is we observe N.1.5 and B.1.5 are producing spectral consistent GANs and this further supports our statement that high frequency spectral discrepancies are not inherent in GANs. We also evaluate the synthetic image detector and observe that N.1.5 and B.1.5 samples can easily bypass the detector. (See table \ref{table:3})

\begin{table}
\begin{center}
\begin{adjustbox}{width=\columnwidth,center}
\begin{tabular}{l|ccc}\toprule
\textbf{Setup} &\textbf{DCGAN} &\textbf{LSGAN} &\textbf{WGAN-GP} \\
\hline
N.1.5 &\boldmath{$1.5\pm2.09\%$} &\boldmath{$0.62\pm0.43\%$} &\boldmath{$0.03\pm0.05\%$} \\
Z.1.5 &$97.88\pm1.04$\% &$95.79\pm2.07$\% &$99.87\pm0.13$\% \\
B.1.5 &\boldmath{$3.54\pm6.15\%$} &\boldmath{$0.07\pm0.09\%$} &\boldmath{$0.14\pm0.13\%$} \\
\bottomrule
\end{tabular}
\end{adjustbox}
\end{center}
\caption{
Detection results for the forensics classifiers proposed by Dzanic \cite{dzanic2020fourier}, using LSUN dataset (128x128). The table shows the successful detection rates (10\% data for training).
}
\label{table:3}
\vspace{-0.2cm}
\end{table}

\begin{table}
\begin{center}
\begin{tabular}{l|c|c|c}\toprule
\textbf{Setup}& \textbf{N.1.5} &\textbf{Z.1.5} &\textbf{B.1.5} \\\midrule
\textbf{Accuracy}& $52.06\pm3.77$\% & $64.3\pm2.7$\% & \boldmath{$0\pm0\%$}\\
\bottomrule
\end{tabular}
\end{center}
\caption{
Detection results for the forensics classifiers proposed by Dzanic \cite{dzanic2020fourier}, using CelebA dataset (256x256). The table shows the successful detection rates (10\% data for training).
}
\label{table:4}
\vspace{-0.2cm}
\end{table}

\subsection{Image-to-Image Translation}
\begin{figure}
    \centering
    \includegraphics[width=0.90\linewidth]{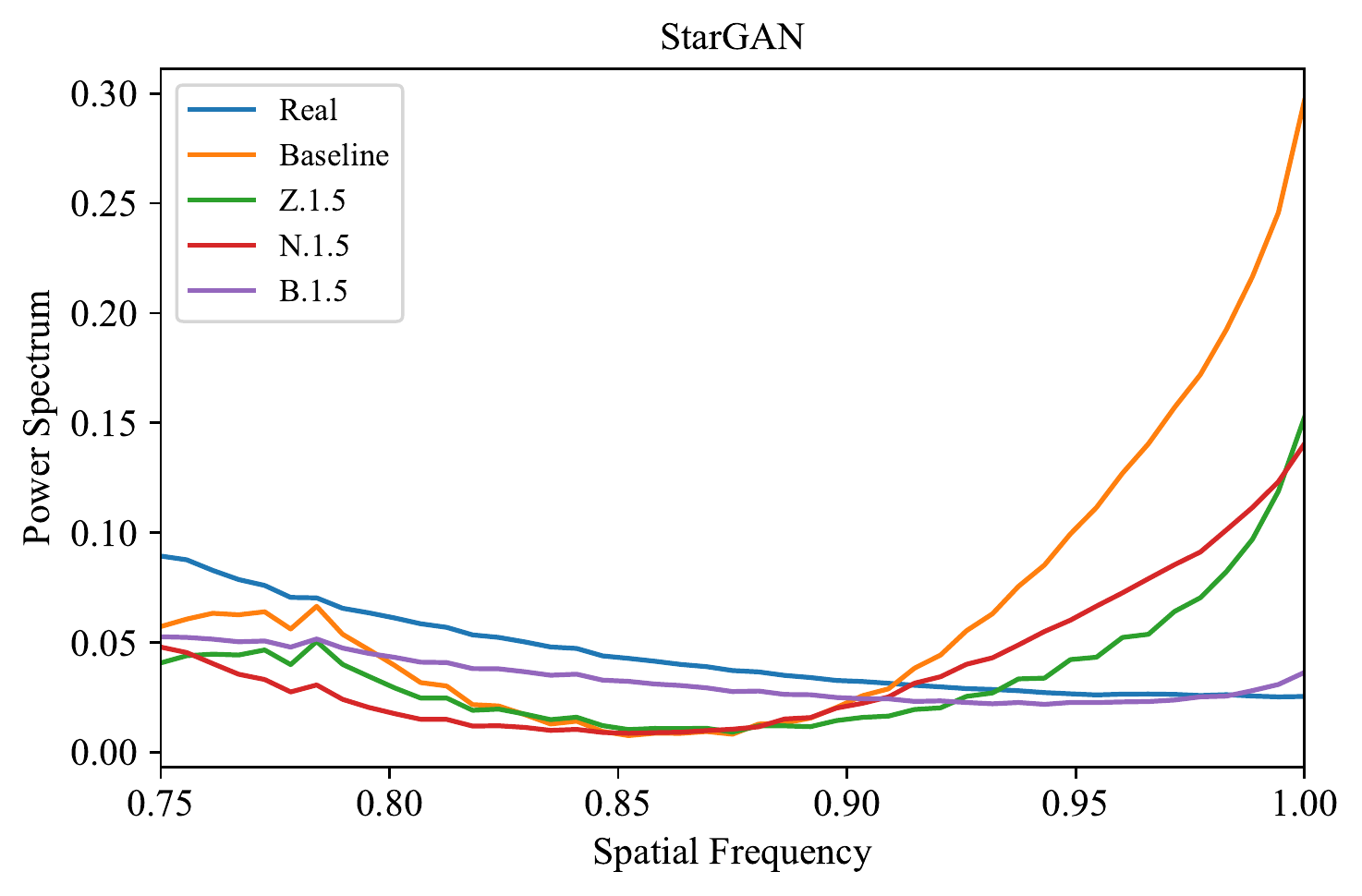}
    \caption{Spectral plots for StarGAN images. We observe that Bilinear feature map scaling produces spectral consistent GANs.}
    \label{fig:stargan}
    \vspace{-0.2cm}
\end{figure}

We extend our experiments to Image-to-Image translation domain using StarGAN \cite{choi2018stargan}. Here we use resized CelebA \cite{liu2015faceattributes} (256x256) used by the official StarGAN \cite{choi2018stargan} implementation to study whether spectral consistency can be achieved by modifying the last feature map scaling operation. We train StarGANs for Z.1.5, N.1.5 and B.1.5 setups. The spectral plots are shown in Figure  \ref{fig:stargan}. We observe that only B.1.5 is able to produce spectral consistent GANs and N.1.5 produces high frequency Fourier discrepancies. We would not ask ourselves why nearest interpolation method behaves differently than bilinear, but rather confirm that we are able to find bilinear interpolation results as more evidence to support our statement that high frequency spectral discrepancies are not inherent characteristics to GANs. We further evaluate the synthetic image detector and observe that B.1.5 samples can bypass the classifier. (See table \ref{table:4})

\section{Spectral Regularization}
The recent Spectral regularization (SR) by Durall \etal \cite{Durall_2020_CVPR} proposed to add a regularizer term to the generator loss to explicitly penalize the generator for spectral distortions. Using SR, they were able to obtain spectral consistency for DCGAN \cite{radford2016unsupervised}, LSGAN \cite{lsgan}, WGAN-GP \cite {NIPS2017_892c3b1c} and DRAGAN \cite{kodali2017convergence} using the celebA \cite{liu2015faceattributes} (128x128). This method encounters computational overhead due to calculation of reduced spectrum for images during training. We show that by modifying the last feature map scaling operation, we are able to achieve spectral consistent GANs without SR.
Importantly, we show that no change in objective functions is needed.
These results and more discussion on SR can be found in Supplementary \ref{sec_sup:sr}.

\section{Discussion}
In this study, we investigated the validity of contemporary beliefs that CNN-based generative models are unable to reproduce
high frequency decay attributes of real images.
We employ a systematic study to design counterexamples to challenge the existing beliefs. With maximum frequency bounded by the spatial resolution, and Fourier discrepancies reported at the highest frequencies, we hypothesized that the last upsampling operation is mostly related to this shortcoming. With carefully designed experiments spanning multiple GAN architectures, loss functions, datasets and resolutions, we observe that high frequency spectral decay discrepancies can be avoided by replacing zero insertion based scaling used by transpose convolutions with nearest or bilinear at the last step. 
\textit{Note that we do not claim that modifying the last feature map scaling method will always fix spectral decay discrepancies in every situation, but rather the goal of our study is to provide counterexamples to argue that high frequency spectral decay discrepancies are not inherent characteristics
of CNN-generated images.}
Further, we easily bypass the recently proposed synthetic image detector that exploits this discrepancy information to detect CNN-generated images indicating that such features are not robust for the purposes of synthetic image detection.

In Supplementary material, we provide more GAN models \cite{Chen2017PhotographicIS, imle} with no high frequency decay discrepancies.  
We also investigate whether such high frequency decay discrepancies are found in other types of computational image synthesis methods (synthesis using Unity game engine\footnote{\url{https://unity.com/}}) \cite{Geiger2012CVPR, vkitti}.
To conclude, through this work we hope to help image forensics research manoeuvre in more plausible directions to combat the fight against CNN-synthesized visual disinformation.

\bigbreak
\noindent
\small
\textbf{Acknowledgments: }
This project was supported by SUTD project PIE-SGP-AI-2018-01.
This research was also supported by the National
Research Foundation Singapore under its AI Singapore Programme [Award Number: AISG-100E2018-005].
This work was also supported by ST Electronics and the National Research Foundation (NRF), Prime
Minister’s Office, Singapore under Corporate Laboratory @ University Scheme (Programme Title:
STEE Infosec - SUTD Corporate Laboratory). 

\noindent
We also gratefully acknowledge the support of NVIDIA AI Technology Center (NVAITC) for our research.




\newpage
\renewcommand\appendixname{Supplementary}
\renewcommand\appendixpagename{Supplementary}

\appendix
\appendixpage
\addappheadtotoc
\noindent 
Supplementary materials include research reproducibility details, additional results and more experiments to support our thesis statement.



\section{Standard Deviation of Experiments}
\begin{figure*}
\centering
\begin{tabular}{ccc}
      \multicolumn{3}{c}{\includegraphics[width=0.9\linewidth]{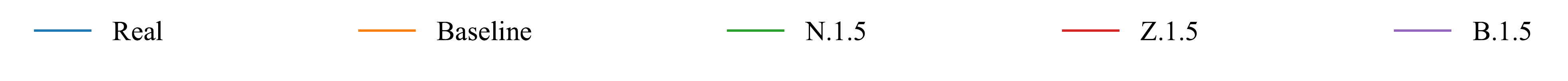}}\\
      \includegraphics[width=0.3\linewidth]{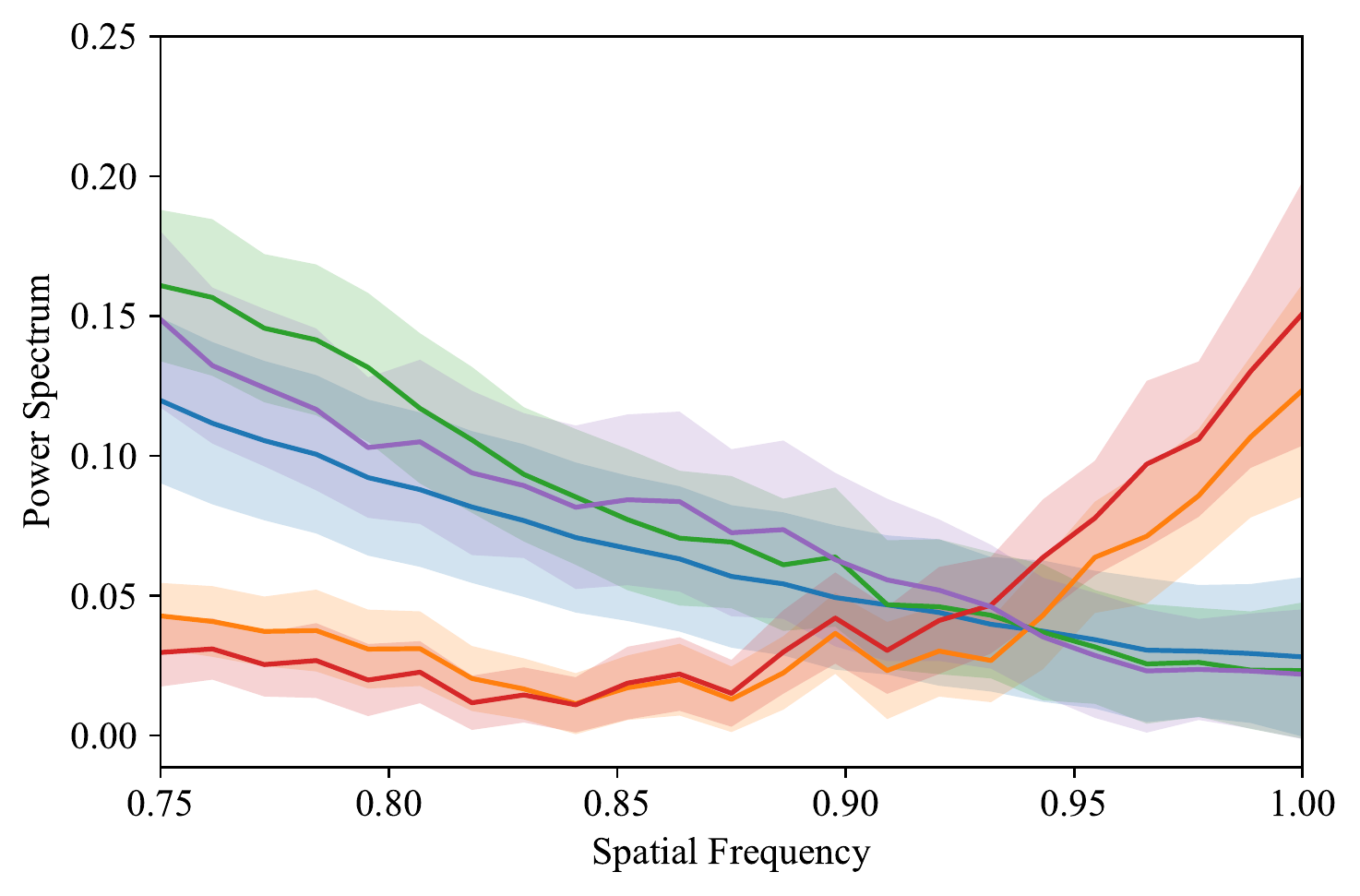} &   
      \includegraphics[width=0.3\linewidth]{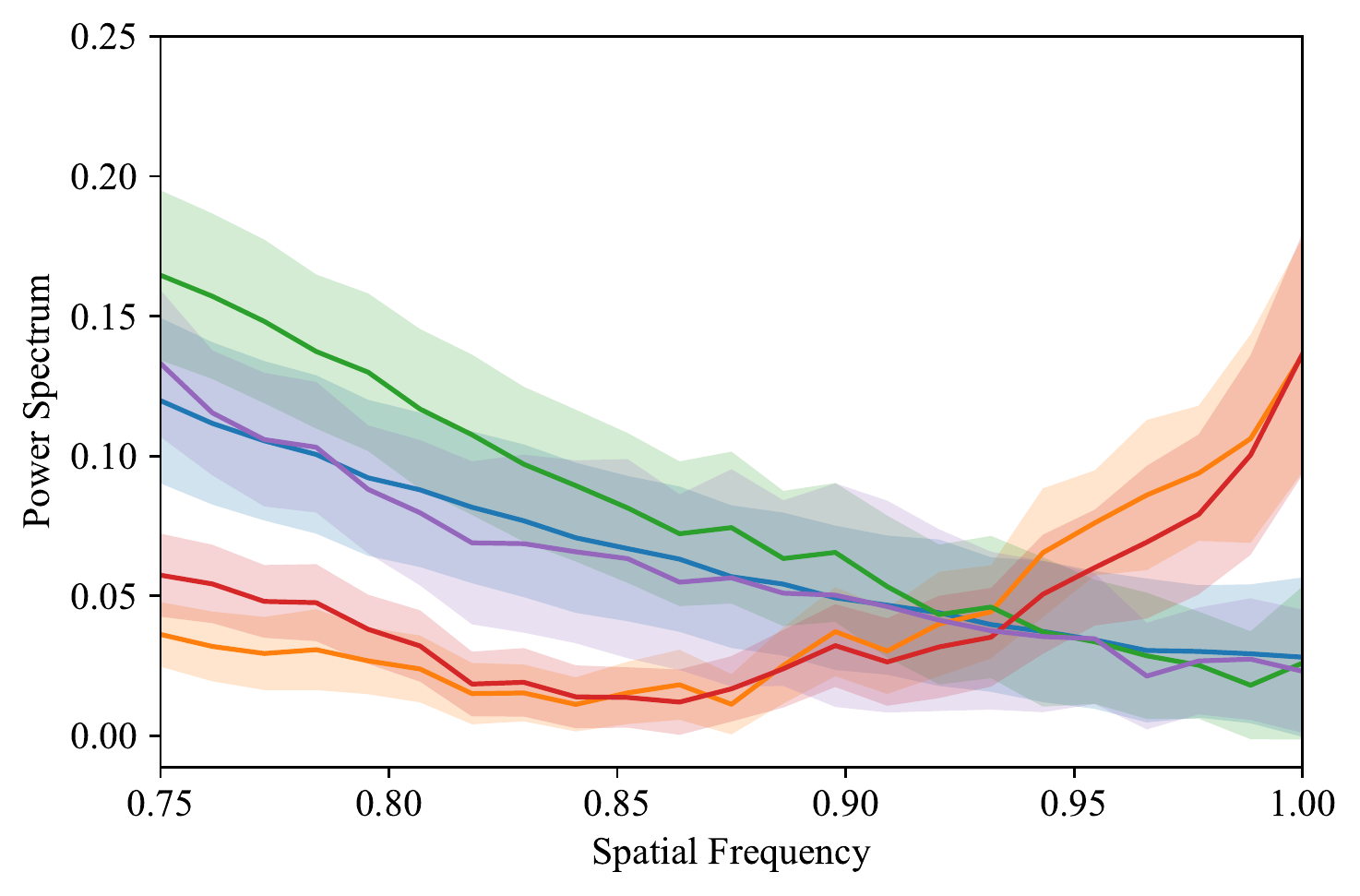} &
      \includegraphics[width=0.3\linewidth]{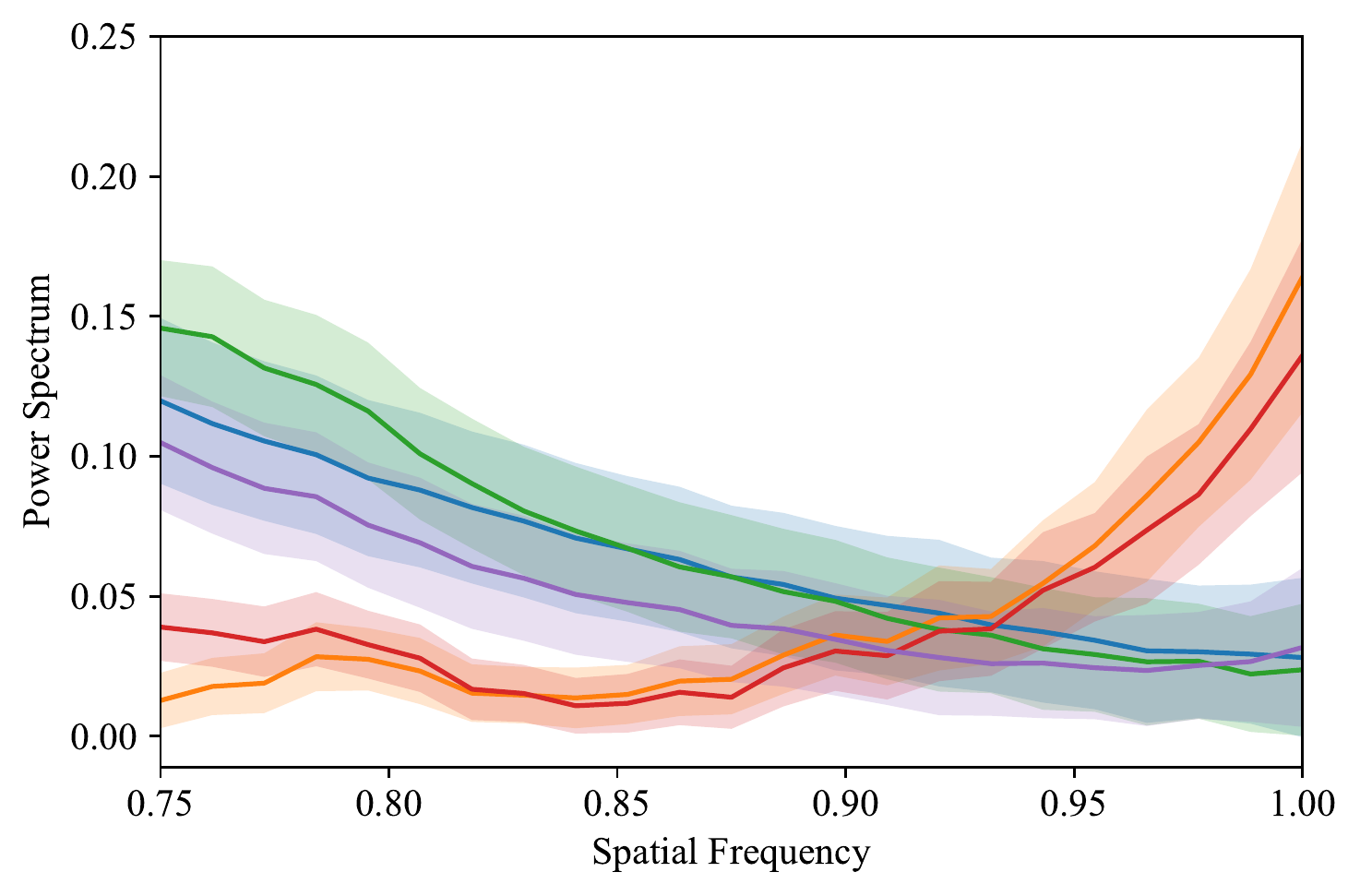}\\
\end{tabular}
\caption{This figure shows spectral plots from Figure 3 in the paper, with standard deviations indicated.
}
\label{fig_sup:sd_celeba}
\end{figure*}

\begin{figure*}
\centering
\begin{tabular}{ccc}
      \multicolumn{3}{c}{\includegraphics[width=0.9\linewidth]{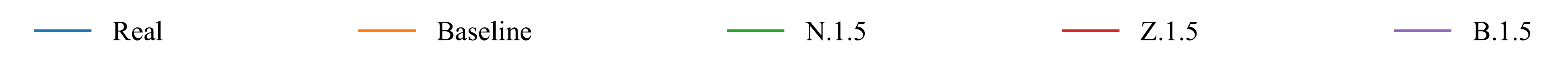}}\\
      \includegraphics[width=0.3\linewidth]{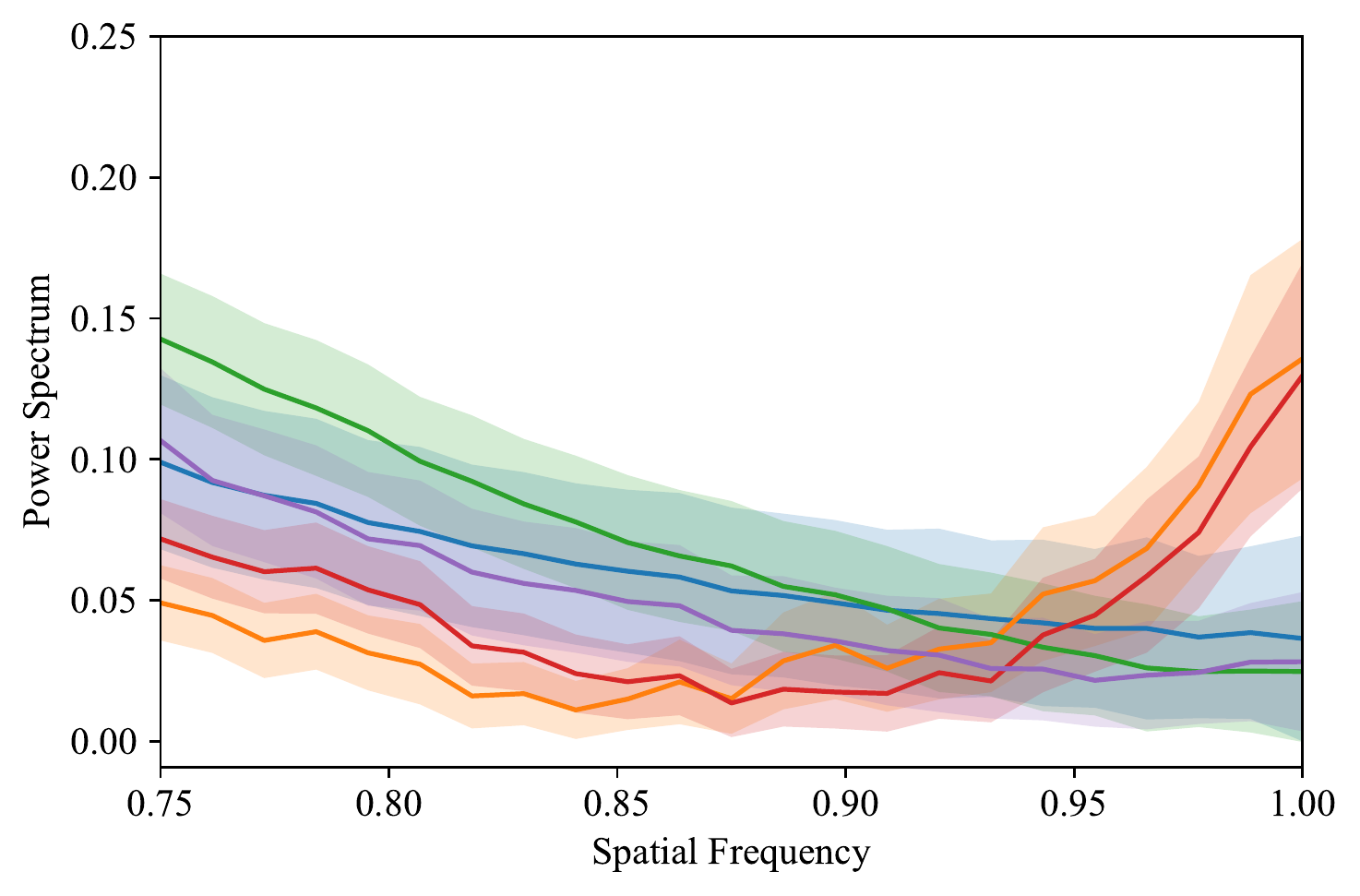} &   
      \includegraphics[width=0.3\linewidth]{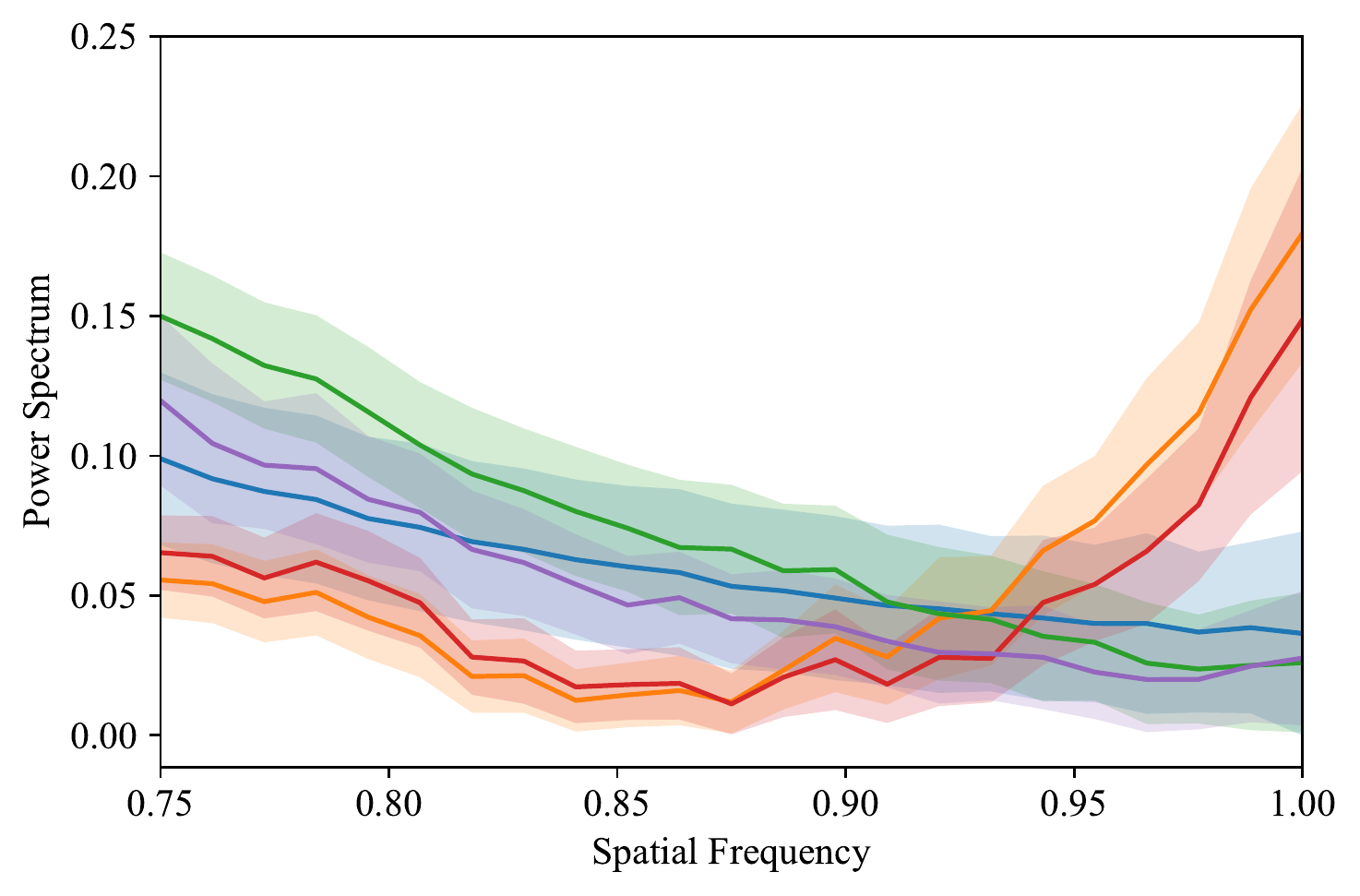} &
      \includegraphics[width=0.3\linewidth]{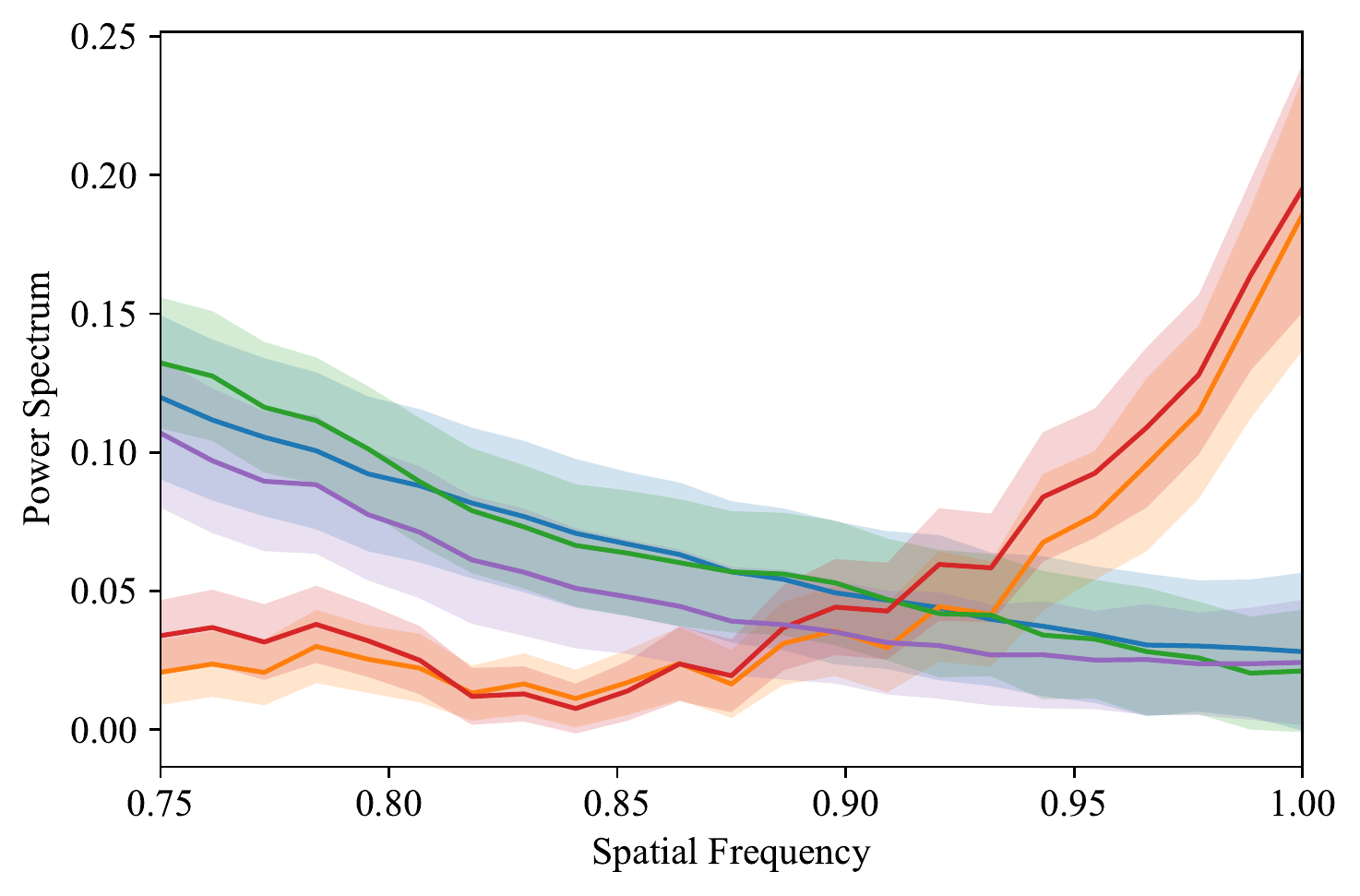}\\
\end{tabular}
\caption{This figure shows spectral plots from Figure 8 in the paper, with standard deviations indicated.
}
\label{fig_sup:sd_lsun}
\end{figure*}

\begin{figure*}
\centering
\begin{tabular}{ccc}
      \multicolumn{3}{c}{\includegraphics[width=0.9\linewidth]{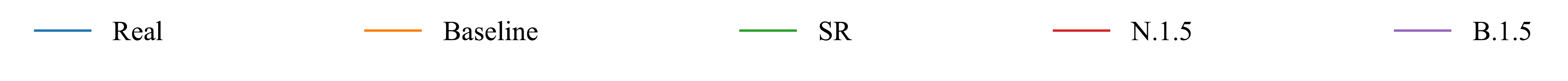}}\\
      \includegraphics[width=0.3\linewidth]{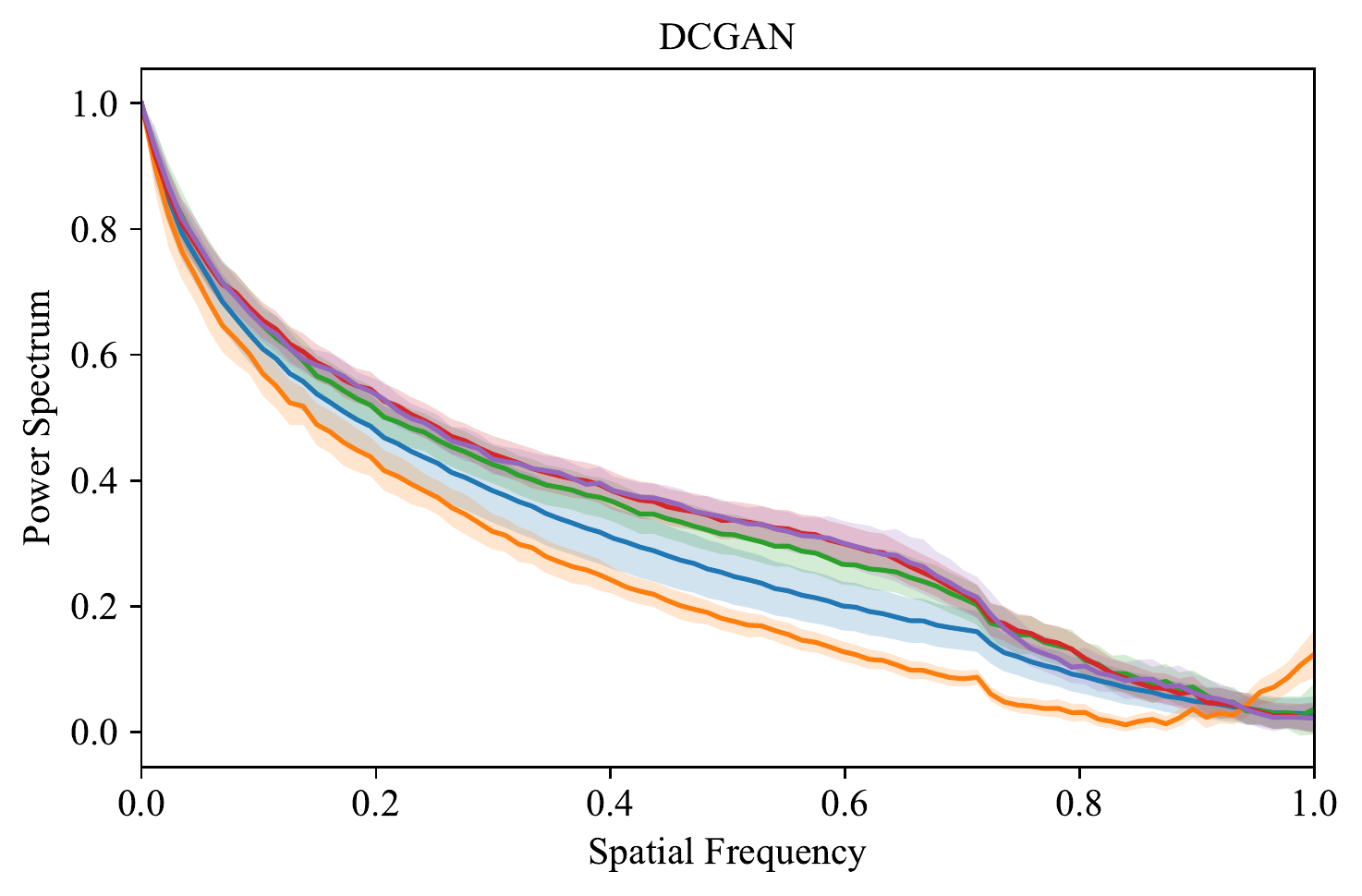} &   
      \includegraphics[width=0.3\linewidth]{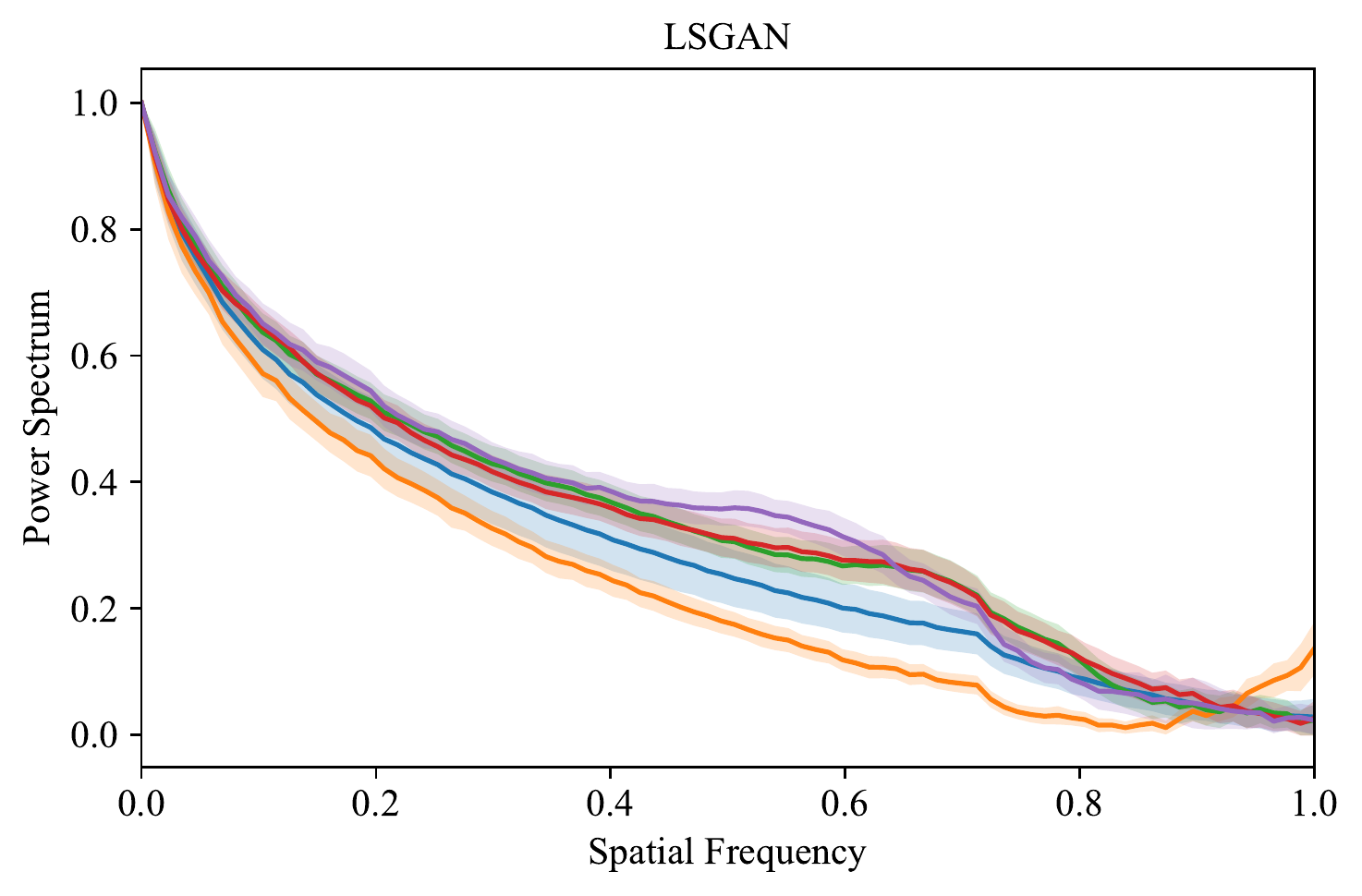} &
      \includegraphics[width=0.3\linewidth]{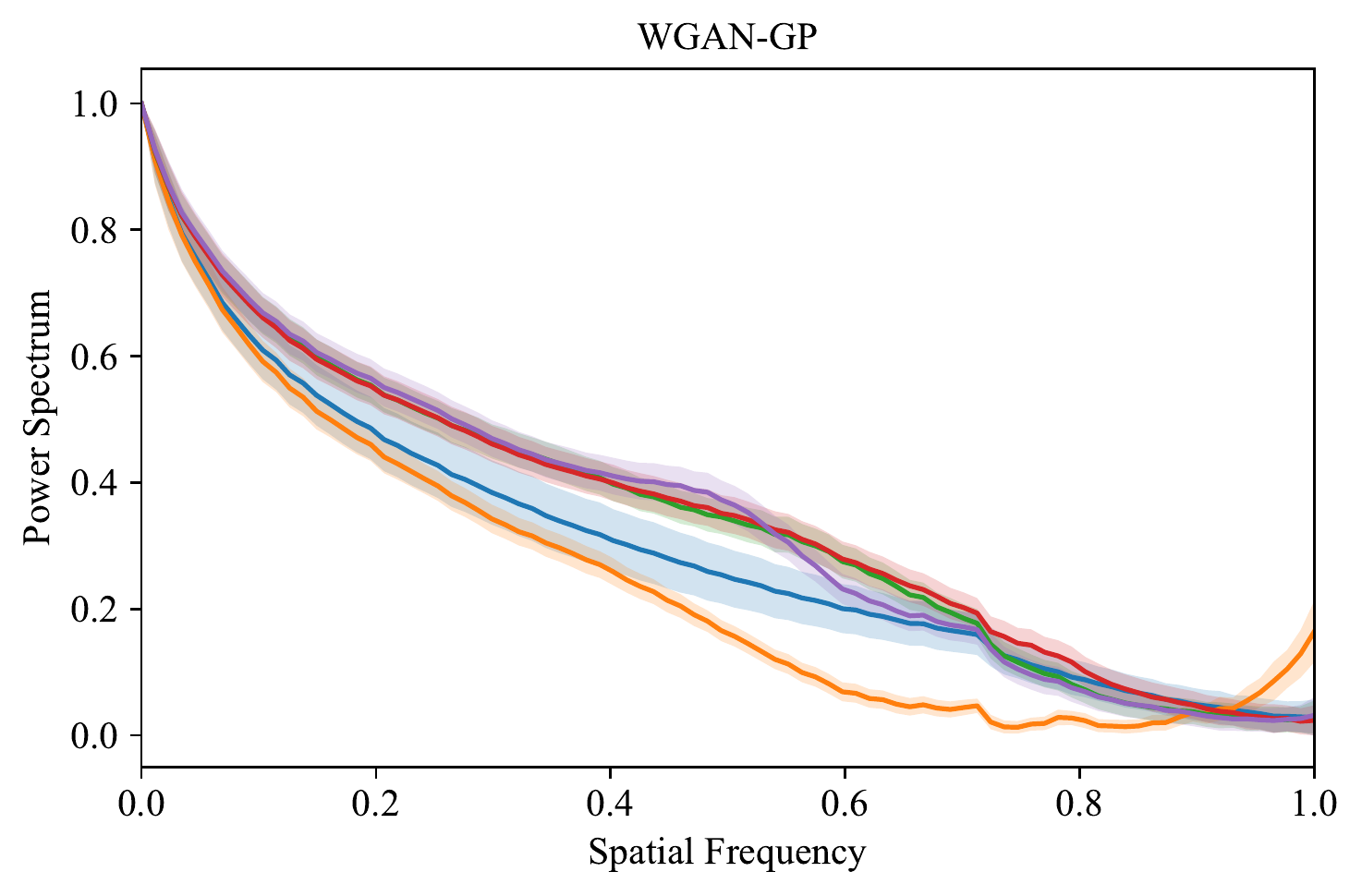}\\
\end{tabular}
\caption{This figure shows spectral plots from Figure 9 in the paper, with standard deviations indicated. ``SR'' refers to Spectral Regularization
}
\label{fig_sup:sd_sr}
\vspace{-0.2cm}
\end{figure*}

\begin{figure}
    \centering
    \includegraphics[width=0.95\linewidth]{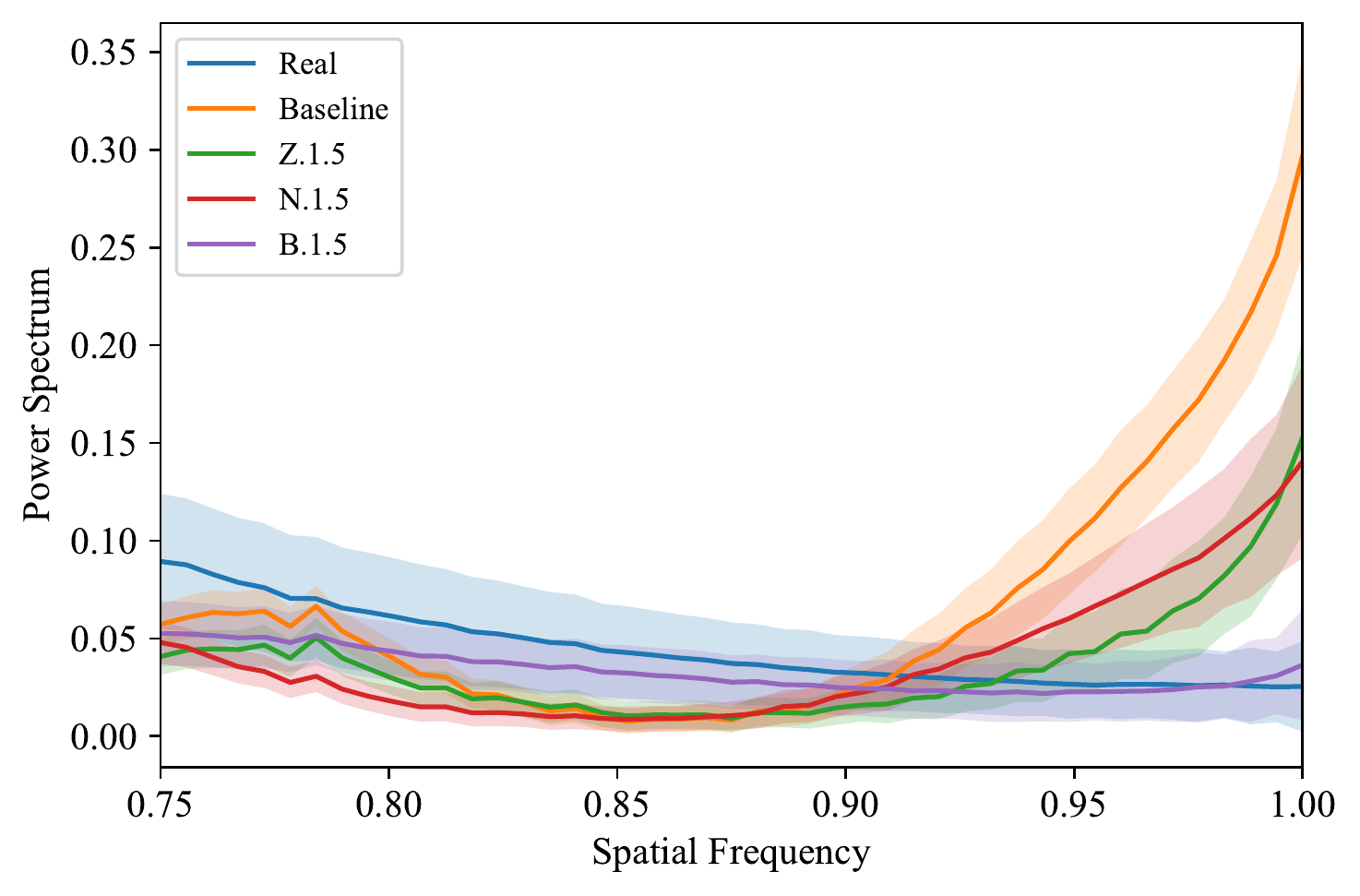}
    \caption{This figure shows spectral plots from Figure 10 (StarGAN) in the paper, with standard deviations indicated}
    \label{fig_sup:sd_stargan}
\end{figure}

Standard deviation of spectral distributions were not included in the paper to allow better readability of graphs. In this section, we include the standard deviation of experiments for Baseline, Z.1.5, N.1.5 and B.1.5 experiments. The spectral distributions with standard deviations for CelebA \cite{liu2015faceattributes}, LSUN \cite{yu15lsun} and StarGAN \cite{choi2018stargan} experiments are shown in Figures \ref{fig_sup:sd_celeba}, \ref{fig_sup:sd_lsun} and \ref{fig_sup:sd_stargan} respectively. We also show the Standard deviations of Spectral regularization experiments in Figure \ref{fig_sup:sd_sr}. In all cases, we observe that the standard deviations are within acceptable range.

\section{Spectral Regularization}
\label{sec_sup:sr}
\begin{figure*}
\centering
\begin{tabular}{ccc}
      \multicolumn{3}{c}{\includegraphics[width=0.9\linewidth]{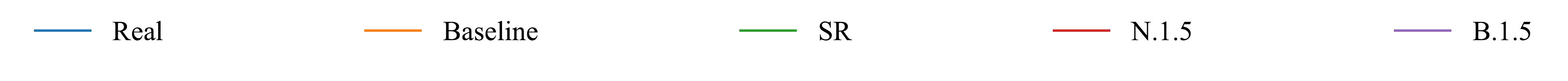}}\\
      \includegraphics[width=0.3\linewidth]{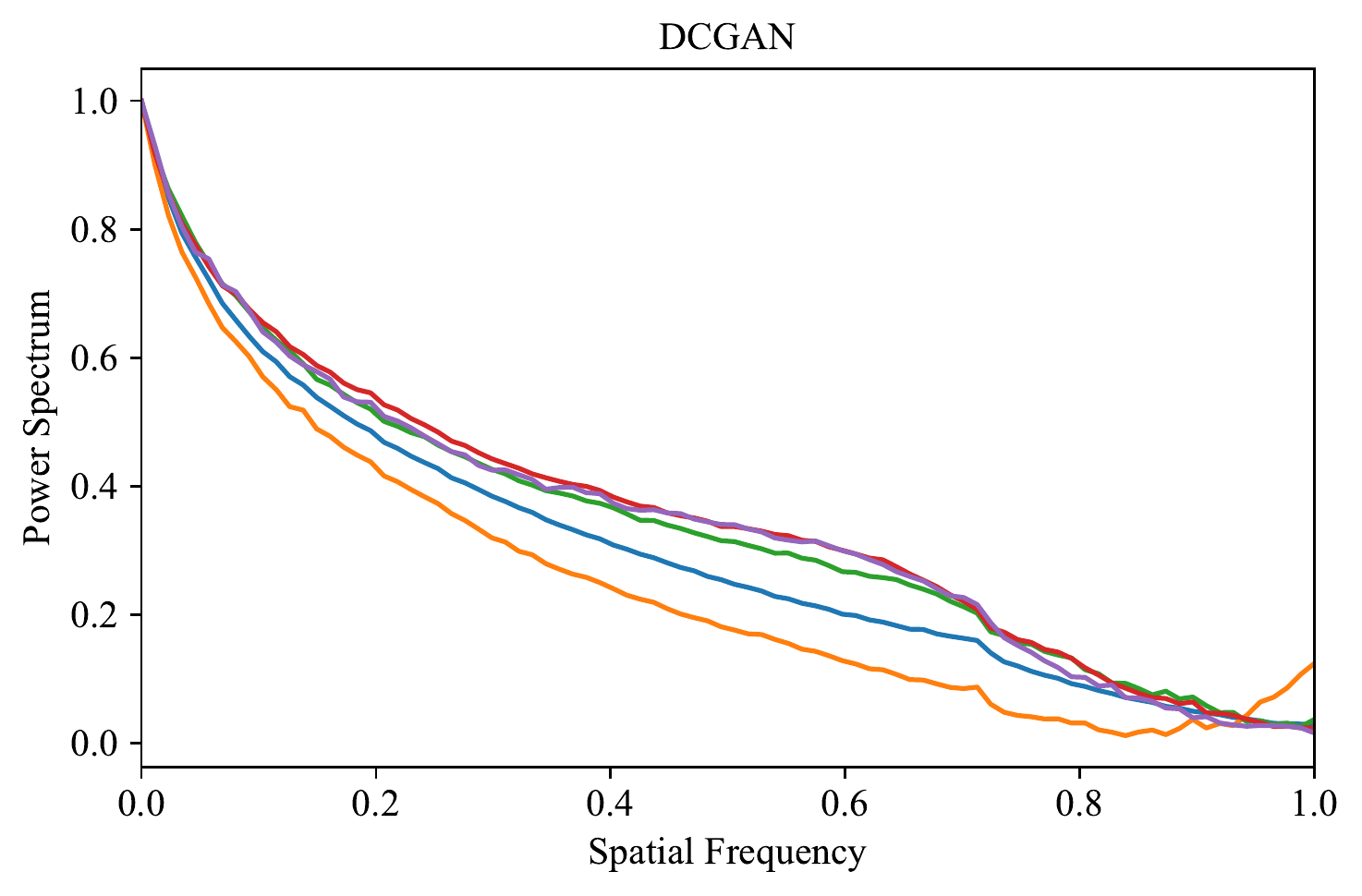} &   
      \includegraphics[width=0.3\linewidth]{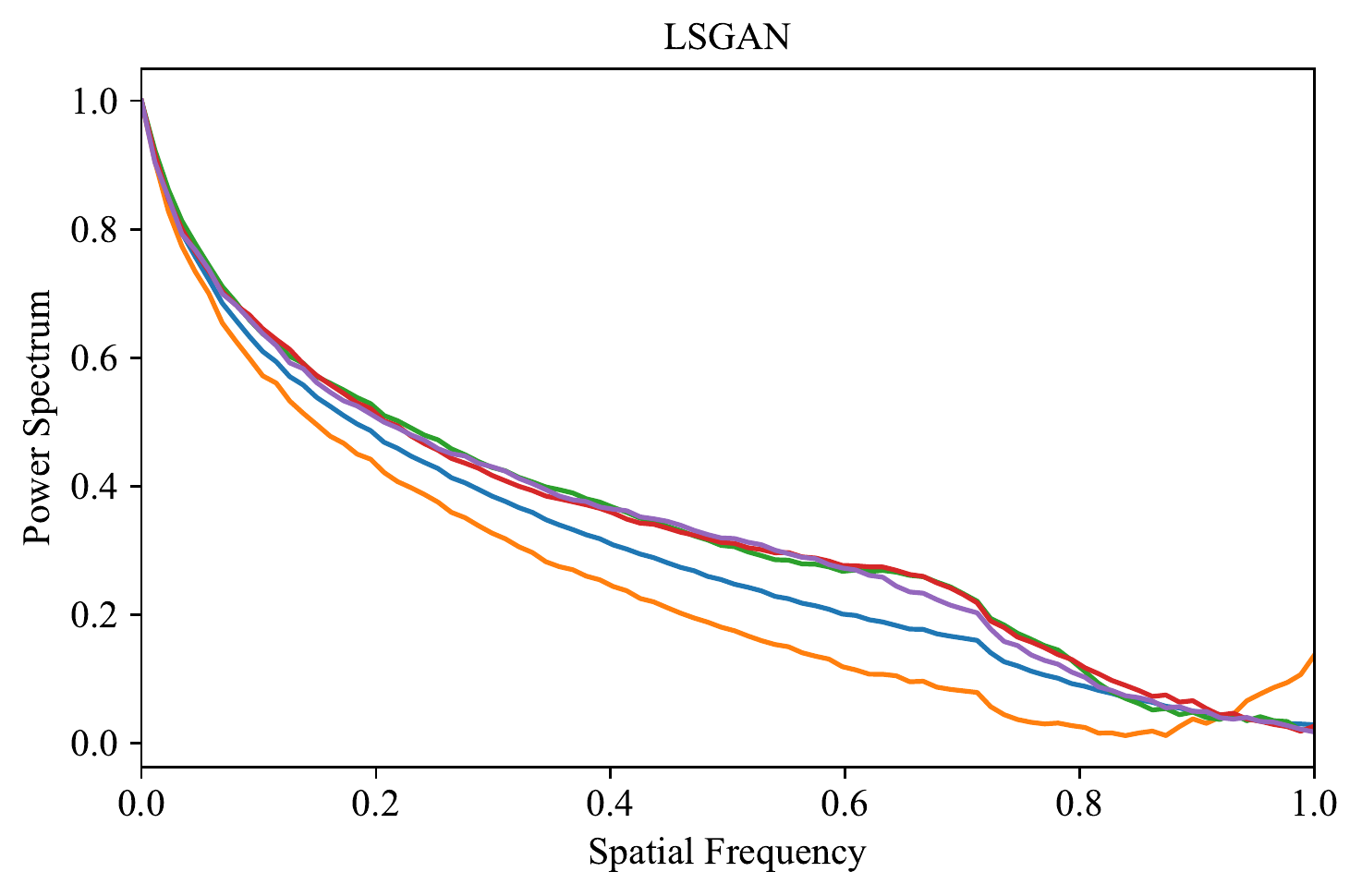} &
      \includegraphics[width=0.3\linewidth]{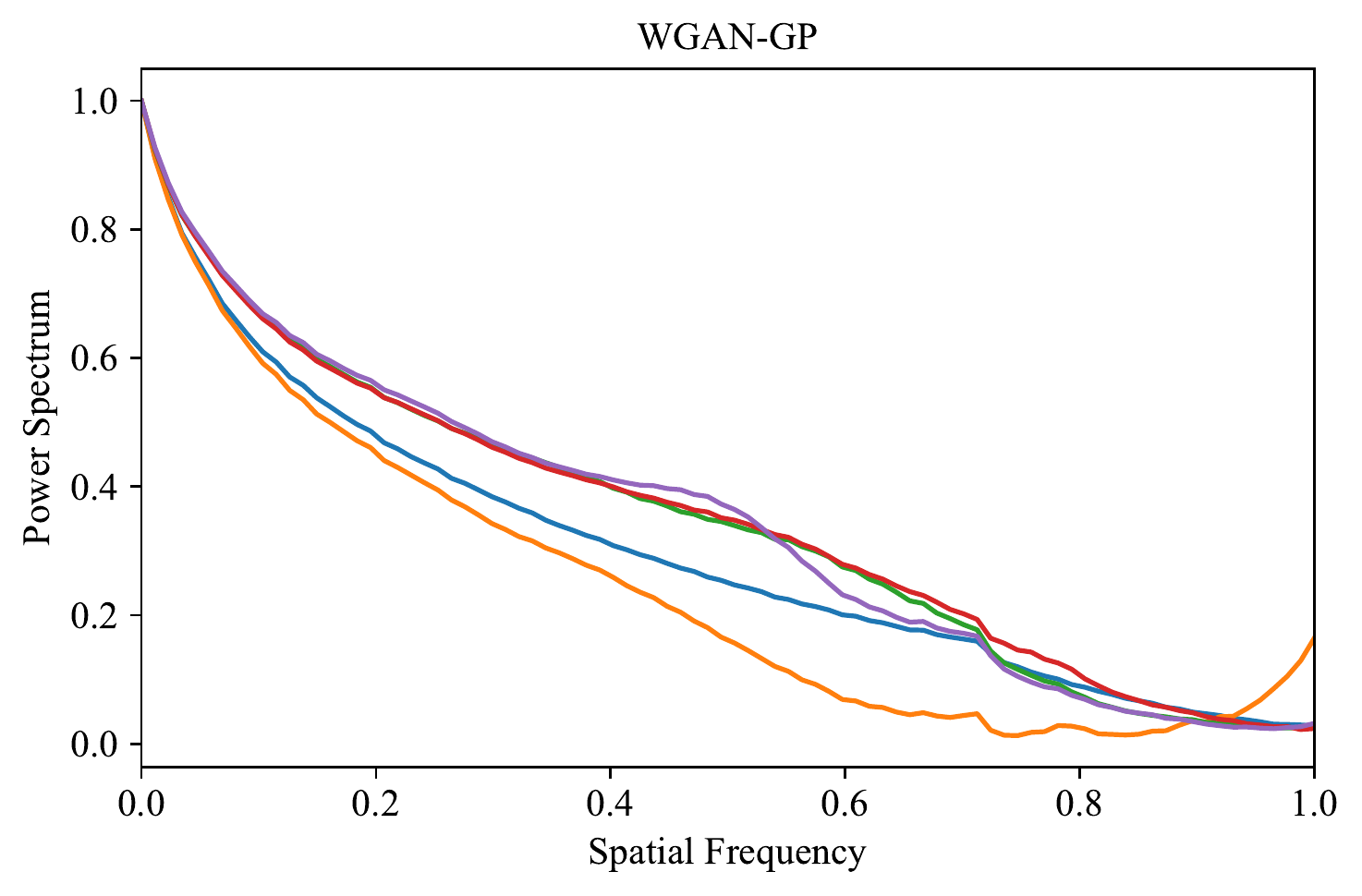}\\
\end{tabular}
\caption{We show the entire spectrum similar to \cite{Durall_2020_CVPR}. ``SR'' refers to Spectral Regularization. We observe that nearest and bilinear interpolation methods produce similar spectral distributions comparing to those models trained with SR. Refer to table 1 in main paper for experiment details.
}
\label{fig_sup:sr}
\vspace{-0.2cm}
\end{figure*}

Results of SR experiments are shown in Figure \ref{fig_sup:sr}. More specifically, SR performs generator loss scaling as there are no gradients with respect to the power spectrum difference between real and synthetic images. We were intrigued by the question on how generator loss scaling can achieve spectral consistency as claimed by \cite{Durall_2020_CVPR} and noticed that the source code uses N.3.5 setup together with SR\footnote{\url{https://github.com/cc-hpc-itwm/UpConv}}. 
Analysing SR is out of scope for this work, but we have showed that N.3.5 setup (Main paper) is sufficient to achieve spectral consistency in identical setups.

\section{Higher Resolution Experiments}
In order to further investigate our thesis statement that high frequency decay discrepancies are not inherent characteristics for CNN-generated images, we extend our analysis to larger resolutions. We use image reconstruction as a representative task to investigate these effects at higher resolutions (We use 512x512).

\begin{figure}[ht]
    \centering
    \includegraphics[width=0.95\linewidth]{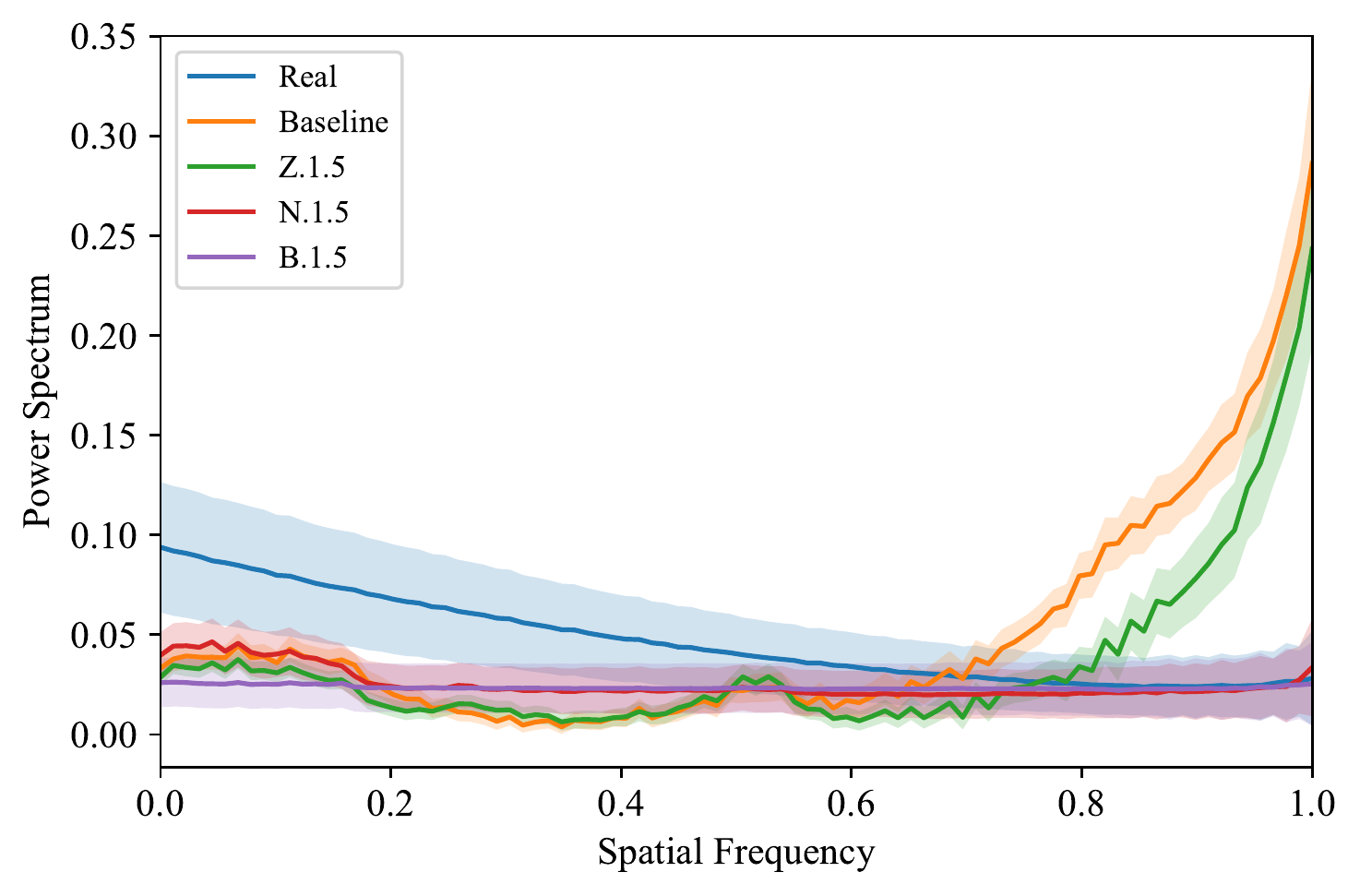}
    \caption{Spectral plots for Image reconstruction at 512x512 for CelebA-HQ. The plots are averaged over 1000 samples. We show the entire spectrum similar to \cite{Durall_2020_CVPR}. Similar to our other results, we observe that nearest and bilinear interpolation methods for the last upsampling step allows to obtain spectral consistent image reconstructions. We also show the standard deviation of experiments in the plot in addition to the mean.}
    \label{fig_sup:ae_spectrum}
    \vspace{-0.2cm}
\end{figure}

We select a subset of CelebA-HQ \cite{karras2018progressive} dataset to train a standard autoencoder for image reconstruction. Similar to experiments in the paper, we perform experiments corresponding to Baseline, Z.1.5, N.1.5 and B.1.5 setups. We observe that Baseline and Z.1.5 setups produce high frequency Fourier discrepancies for reconstructed images, and N.1.5 and B.1.5 setups produce spectral consistent reconstructed images. The spectral distributions are shown in Figure \ref{fig_sup:ae_spectrum}. This further confirms that high frequency Fourier discrepancies are not intrinsic for CNN-generated images.

\section{Samples}
\label{sec_sup:image_samples}
We show more samples and spectral distributions for important CelebA \cite{liu2015faceattributes} experiments in this section. 

\begin{figure*}
    \centering
    \begin{tabular}{b{0.05cm}|l}
        \rot{Baseline} &
        \includegraphics[width=0.90\linewidth]{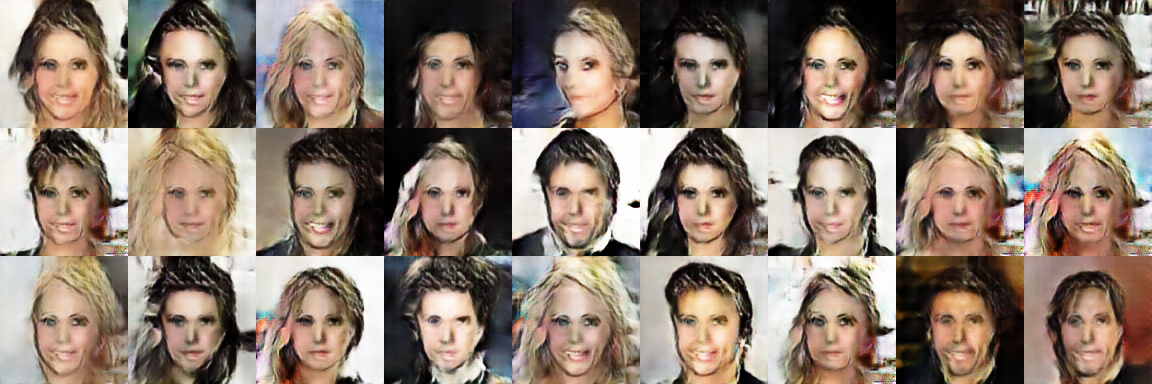} \\
        \hline
        \rot{Z.1.5} &\includegraphics[width=0.90\linewidth]{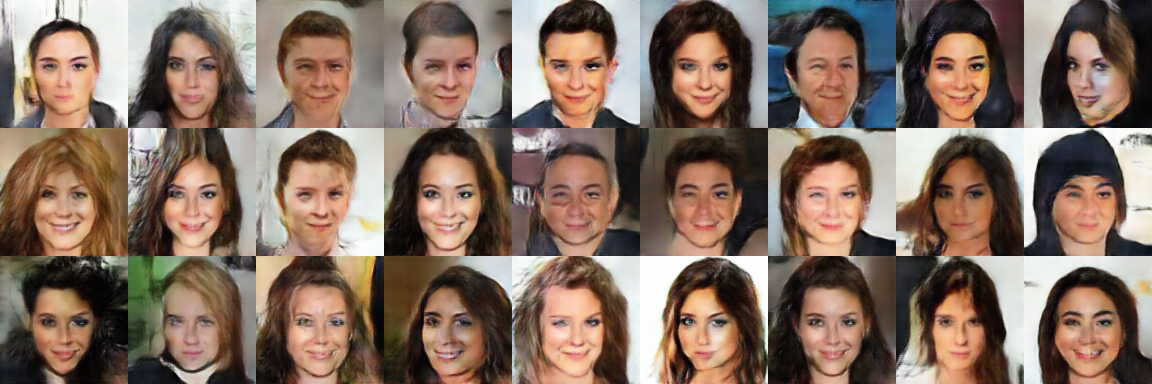} \\
        \hline
        \rot{N.1.5} &\includegraphics[width=0.90\linewidth]{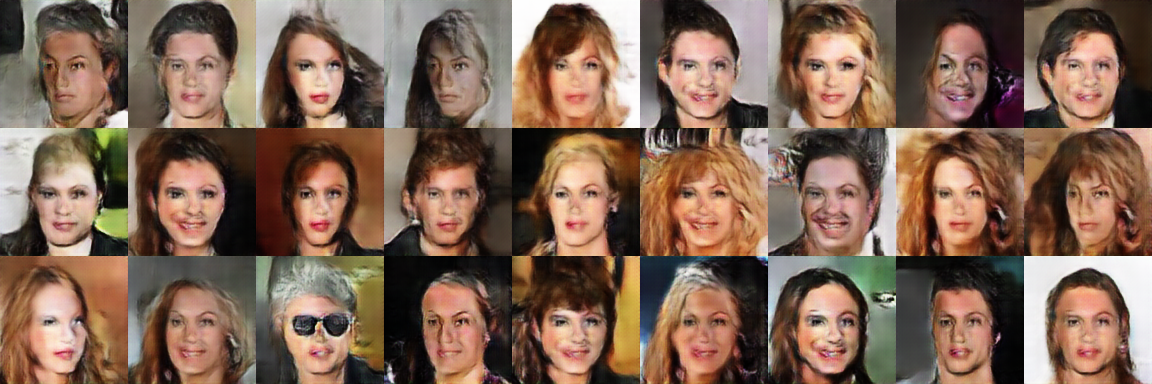} \\
        \hline
        \rot{B.1.5} &\includegraphics[width=0.90\linewidth]{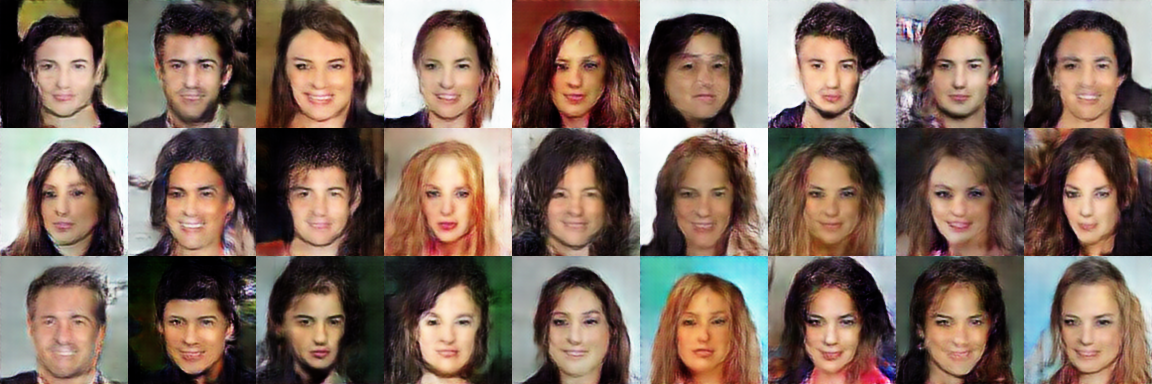}
    \end{tabular}
    
    \caption{DCGAN \cite{radford2016unsupervised} samples for CelebA \cite{liu2015faceattributes}. Refer to Table 1 in paper for experiment codes.}
    \label{fig_sup:dcgan_samples}
\end{figure*}

\begin{figure*}
    \centering
    \begin{tabular}{b{0.05cm}|l}
        \rot{Baseline} & \includegraphics[width=0.90\linewidth]{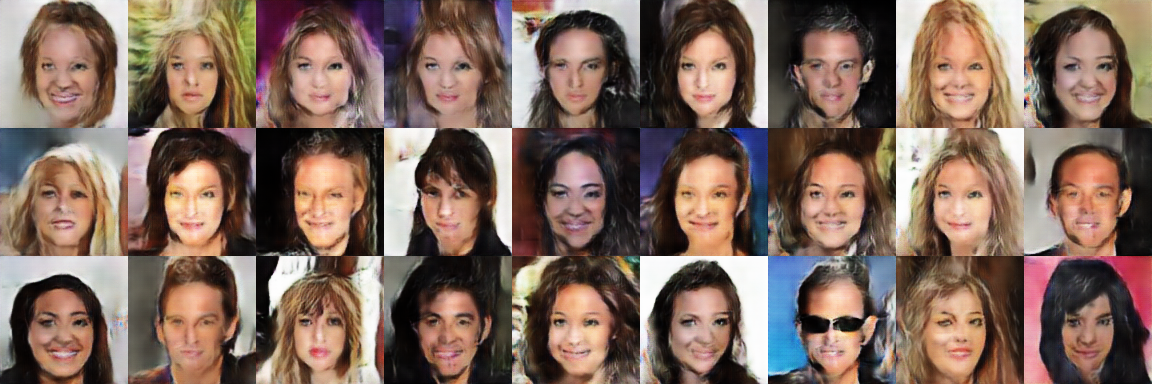} \\
        \hline
        \rot{Z.1.5} &\includegraphics[width=0.90\linewidth]{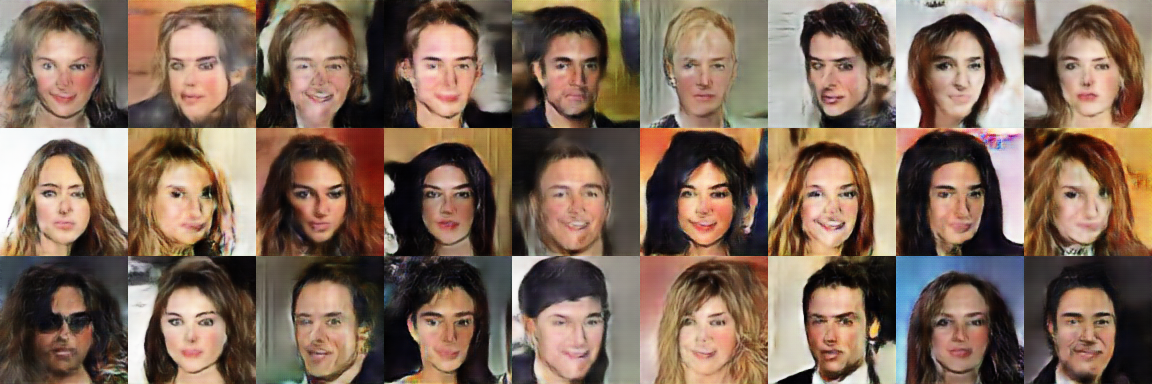} \\
        \hline
        \rot{N.1.5} &\includegraphics[width=0.90\linewidth]{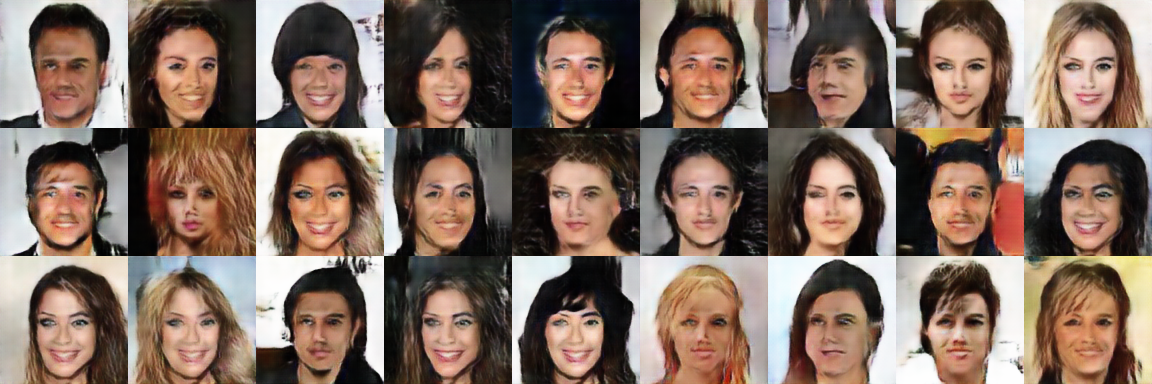} \\
        \hline
        \rot{B.1.5} &\includegraphics[width=0.90\linewidth]{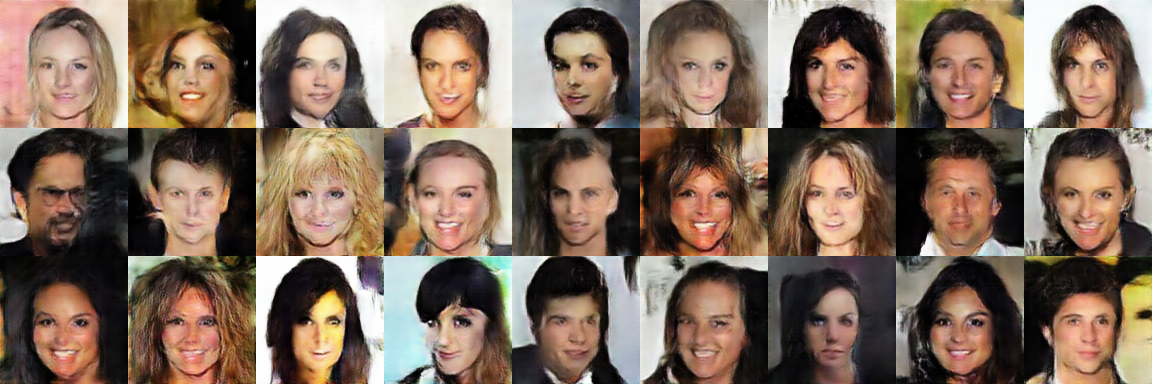}
    \end{tabular}
    
    \caption{LSGAN \cite{lsgan} samples for CelebA \cite{liu2015faceattributes}. Refer to Table 1 in paper for experiment codes.}
    \label{fig_sup:lsgan_samples}
\end{figure*}

\begin{figure*}
    \centering
    \begin{tabular}{b{0.05cm}|l}
        \rot{Baseline} & \includegraphics[width=0.90\linewidth]{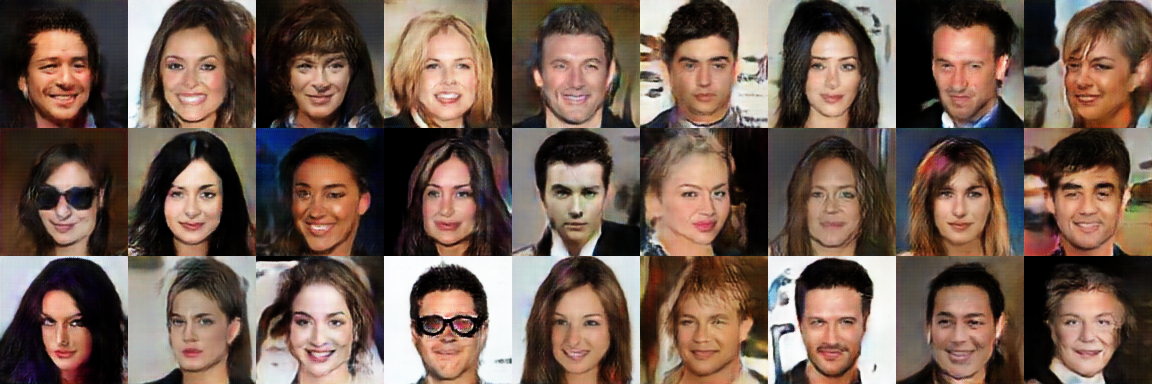} \\
        \hline
        \rot{Z.1.5} &\includegraphics[width=0.90\linewidth]{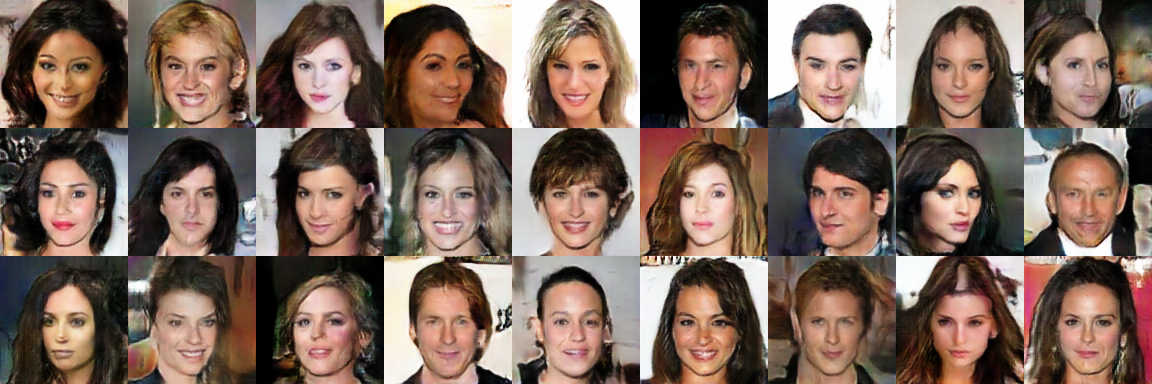} \\
        \hline
        \rot{N.1.5} &\includegraphics[width=0.90\linewidth]{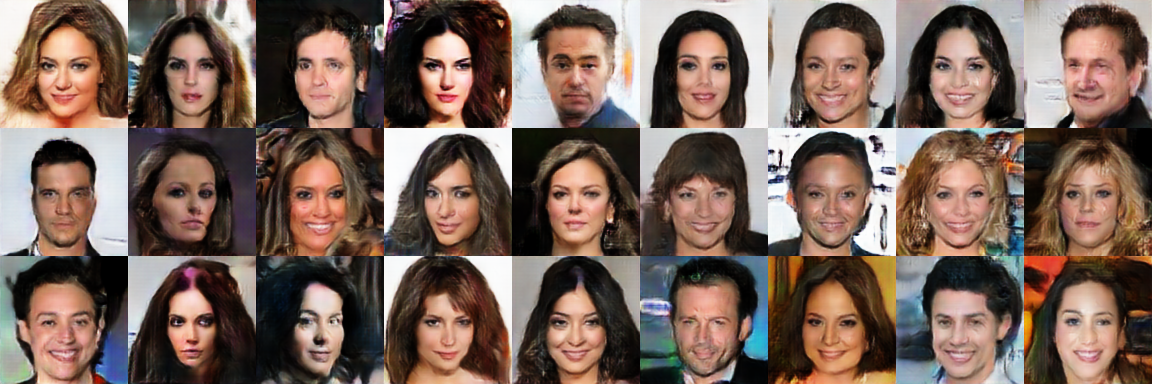} \\
        \hline
        \rot{B.1.5} &\includegraphics[width=0.90\linewidth]{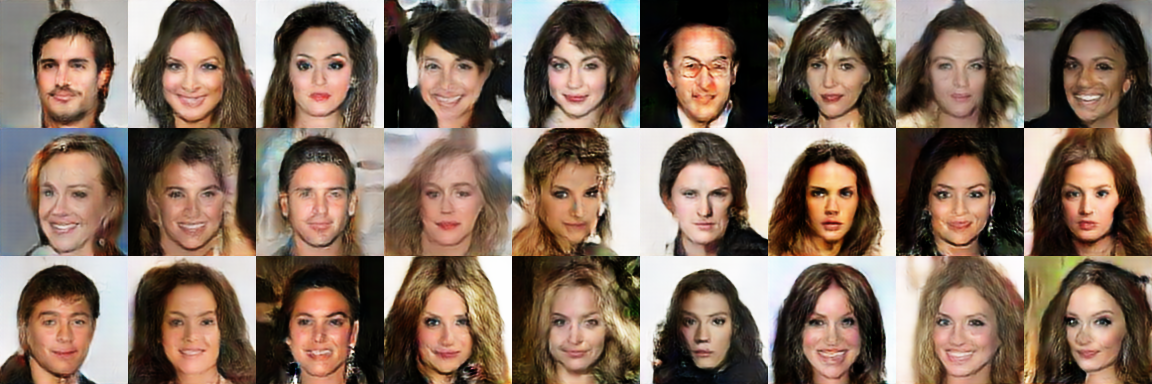}
    \end{tabular}
    
    \caption{WGAN-GP \cite{NIPS2017_892c3b1c} samples for CelebA \cite{liu2015faceattributes}. Refer to Table 1 in paper for experiment codes.}
    \label{fig_sup:wgan_samples}
\end{figure*}

\noindent\newline
\textbf{GAN Samples:} We show extensive samples corresponding to Baseline, Z.1.5, N.1.5 and B.1.5 setups for DCGAN \cite{radford2016unsupervised}, LSGAN \cite{lsgan}, WGAN-GP \cite{NIPS2017_892c3b1c} in Figures \ref{fig_sup:dcgan_samples}, \ref{fig_sup:lsgan_samples}, \ref{fig_sup:wgan_samples} respectively.

\begin{figure*}
\centering
\begin{tabular}{ccccc}
    \includegraphics[width=0.17\linewidth]{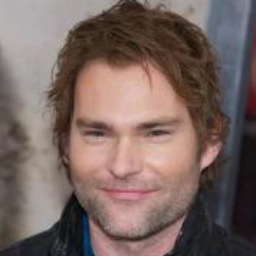} &   
    \includegraphics[width=0.17\linewidth]{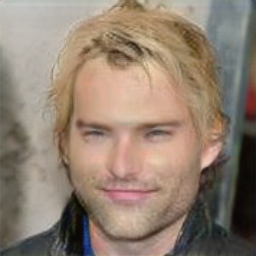} &
    \includegraphics[width=0.17\linewidth]{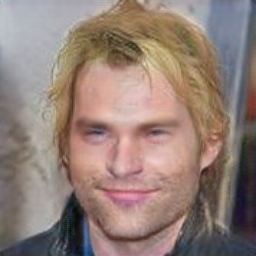} &
    \includegraphics[width=0.17\linewidth]{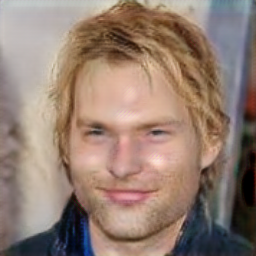} &
    \includegraphics[width=0.17\linewidth]{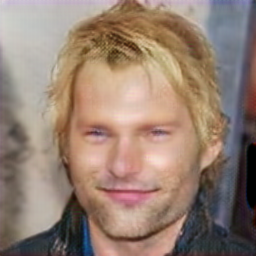}\\
\end{tabular}
\caption{Image Translation results using StarGAN. Original Image (leftmost), Baseline (column 2), 
Z.1.5 (column 3), N.1.5 (column 4), B.1.5 (rightmost) for attribute Blonde hair is shown. Corresponding high frequency spectral distributions are shown in Figure \ref{fig_sup:stargan_spectral_dist}. Refer to table 1 in paper for experiment codes.
}
\label{fig_sup:stargan_ex2}
\vspace{-0.2cm}
\end{figure*}

\begin{figure}
\centering
\begin{tabular}{cc}
    \includegraphics[width=0.90\linewidth]{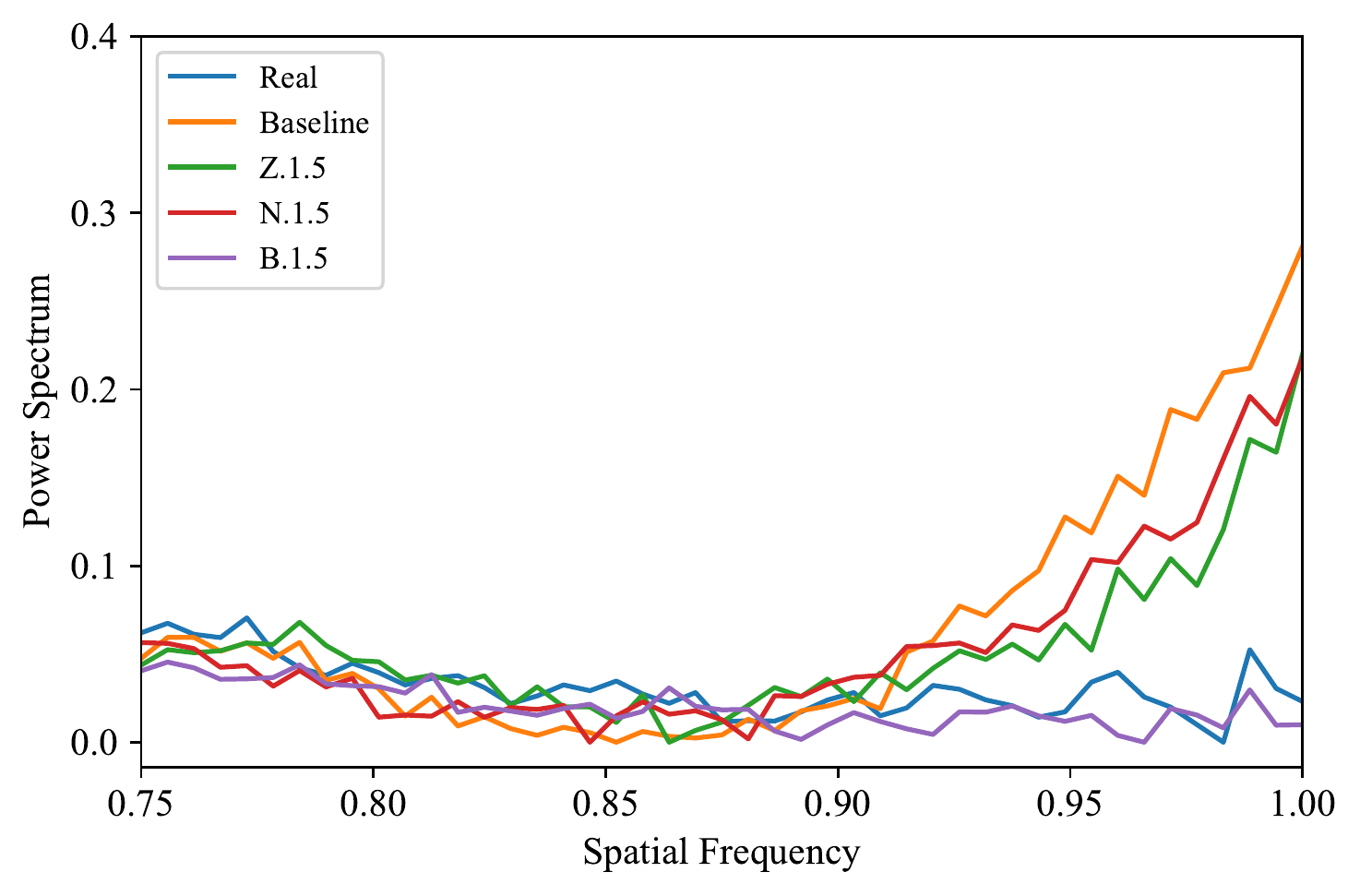}\\
\end{tabular}
\caption{Spectral distributions for examples shown in \ref{fig_sup:stargan_ex2}. Real refers to Original image. We observe that B.1.5 setup produces spectral consistent images similar to observations recorded in the paper. Refer to table 1 in paper for experiment codes.
}
\label{fig_sup:stargan_spectral_dist}
\vspace{-0.2cm}
\end{figure}

\noindent\newline
\textbf{Image-to-Image Translation: }
For StarGAN \cite{choi2018stargan} experiments, we show an example of a reference image with corresponding translated images for Baseline, Z.1.5, N.1.5 and B.1.5 setups in Figures \ref{fig_sup:stargan_ex2}. We show the corresponding spectral distributions in Figure \ref{fig_sup:stargan_spectral_dist}. 

\begin{figure*}[ht]
\centering
\begin{tabular}{ccccc}
    \includegraphics[width=0.17\linewidth]{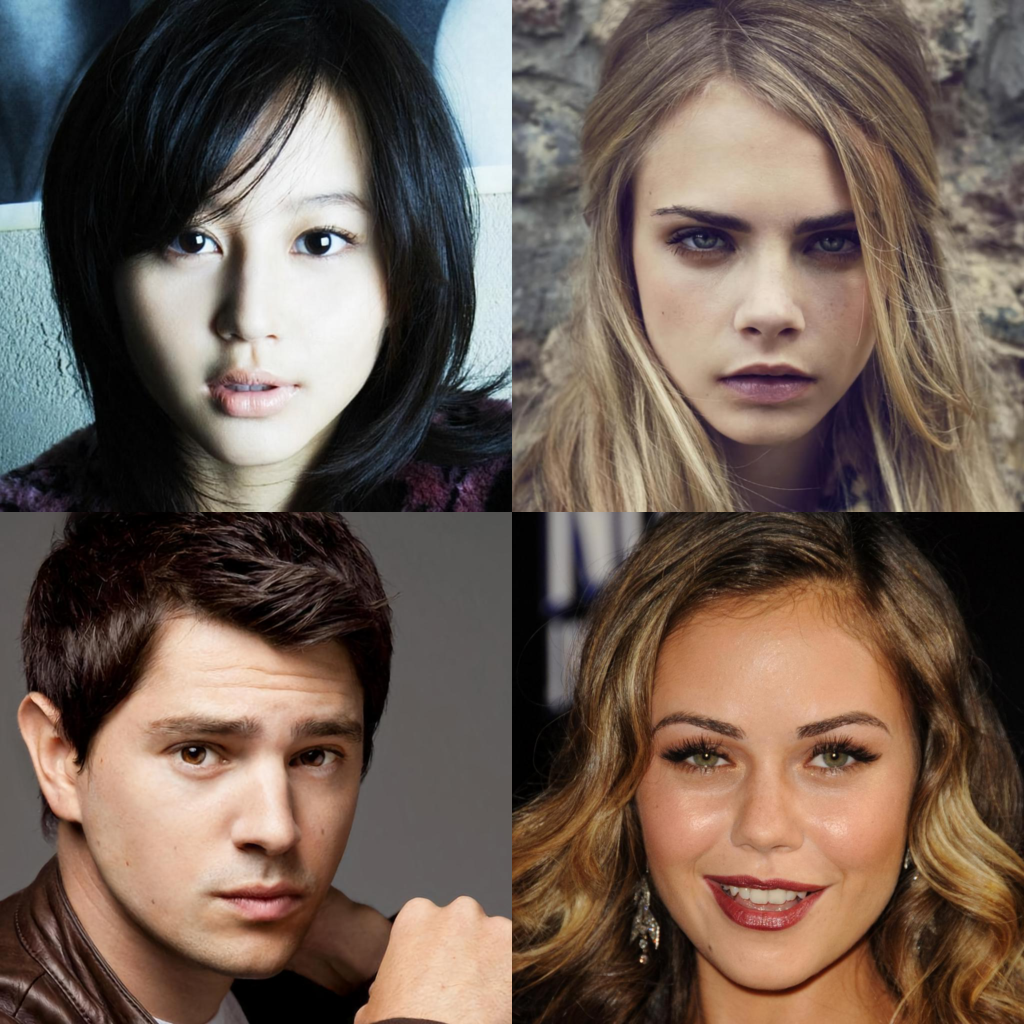} & 
    \includegraphics[width=0.17\linewidth]{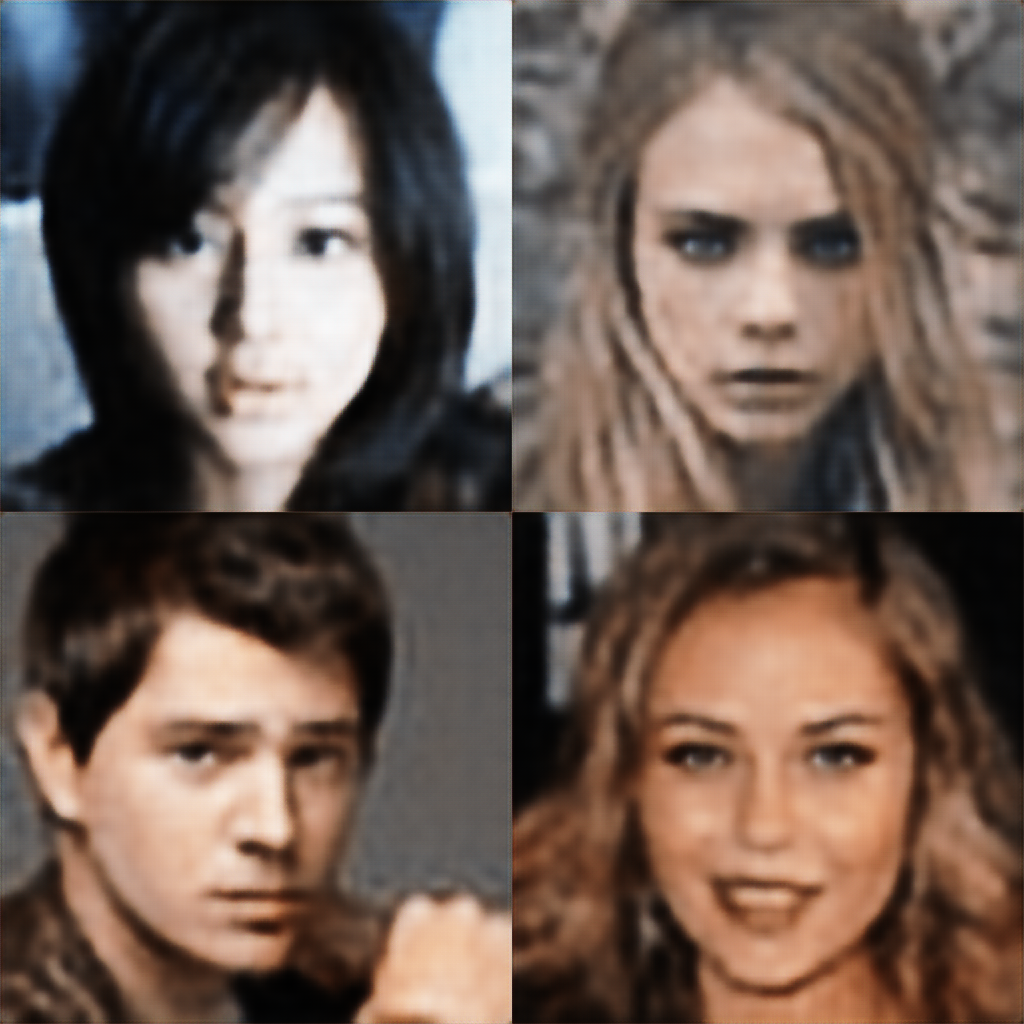} &
    \includegraphics[width=0.17\linewidth]{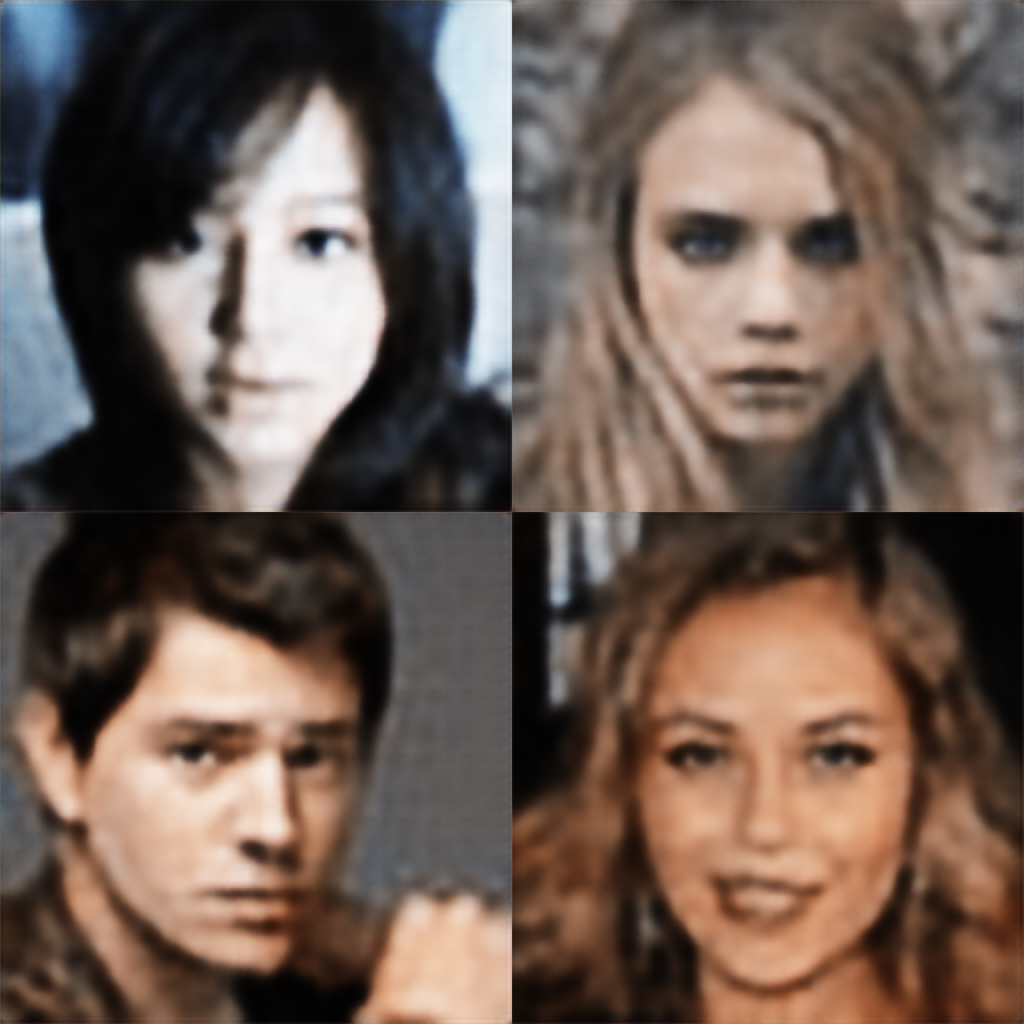} & 
    \includegraphics[width=0.17\linewidth]{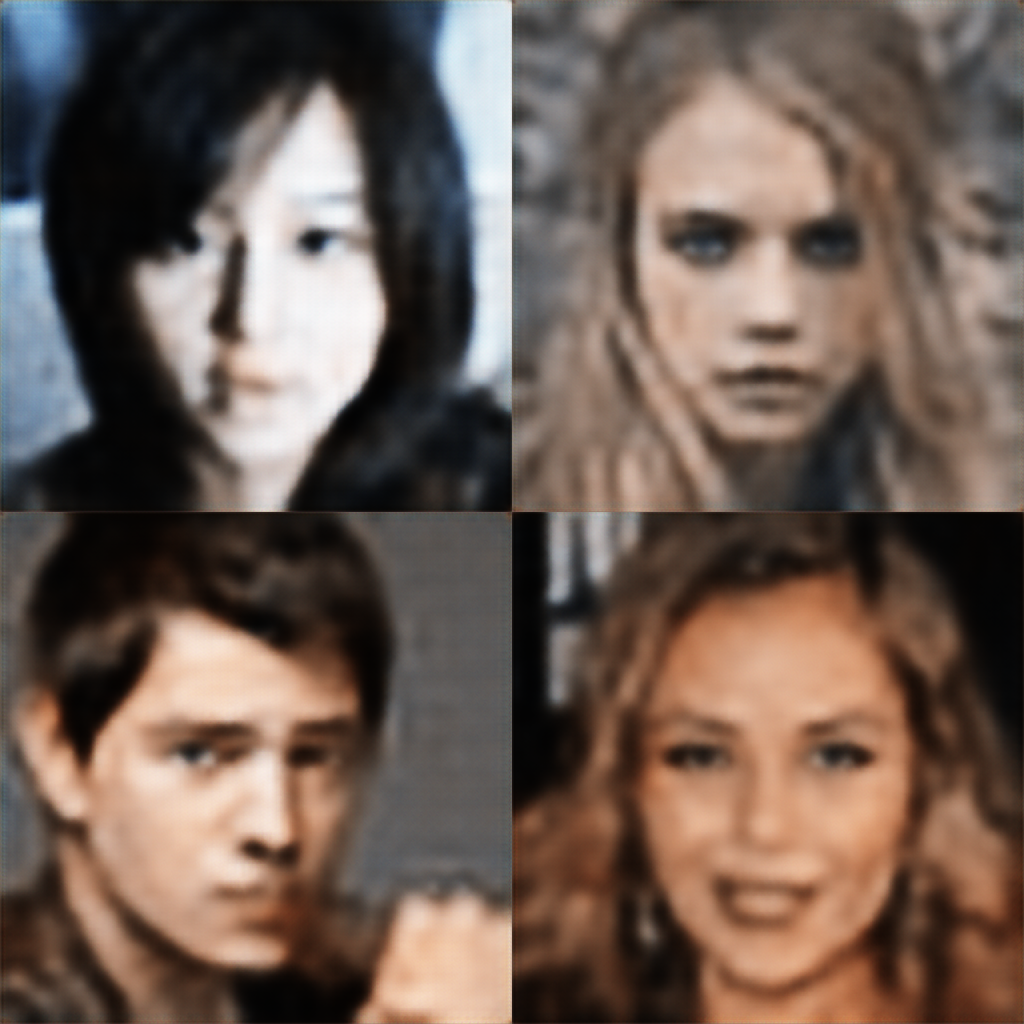} &
    \includegraphics[width=0.17\linewidth]{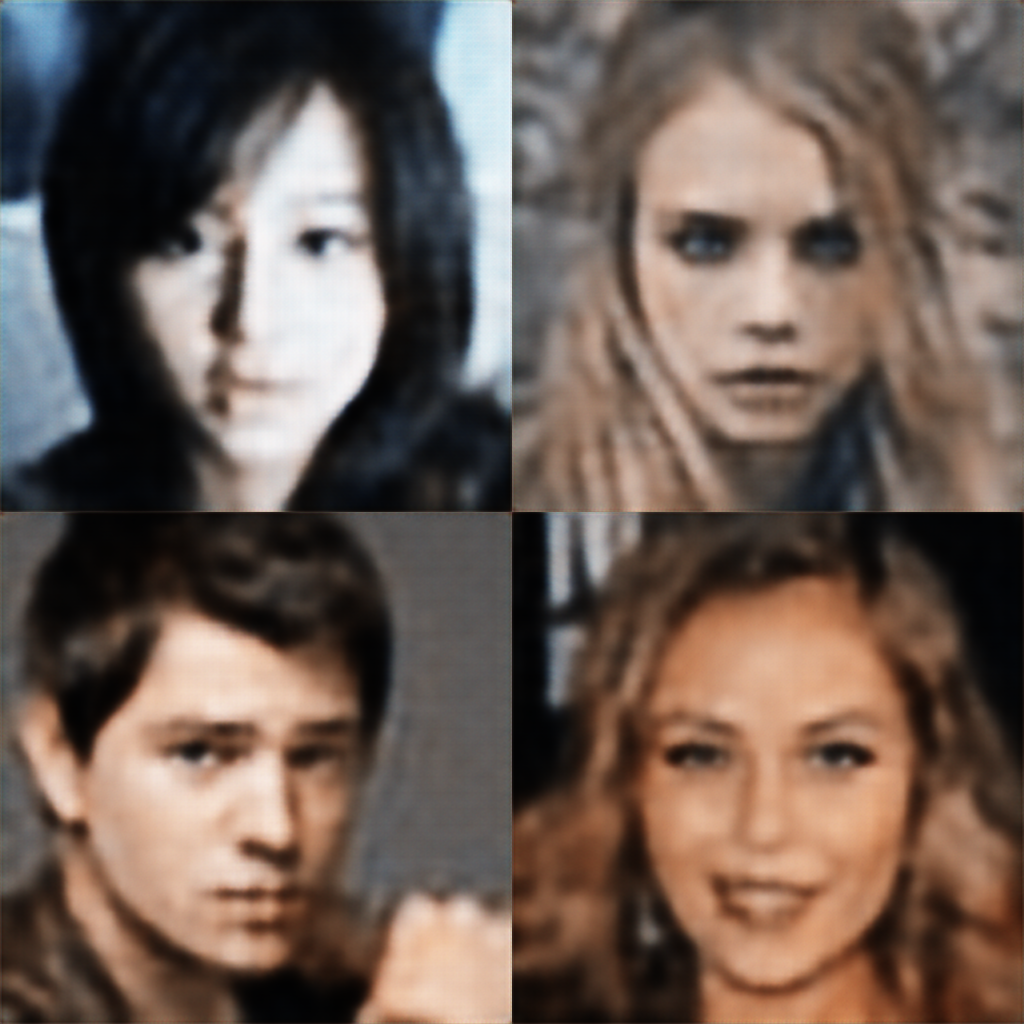}\\
\end{tabular}
\caption{Reconstruction Results. Original Image (leftmost), Baseline (column 2), 
Z.1.5 (column 3), N.1.5 (column 4), B.1.5 (rightmost) for CelebA-HQ \cite{karras2018progressive} samples are shown. Refer to table 1 in paper for experiment codes.}
\label{fig_sup:ae_samples}
\vspace{-0.2cm}
\end{figure*}

\noindent\newline
\textbf{Image Reconstruction: }
We show image reconstruction results for a few CelebA-HQ \cite{karras2018progressive} examples in Figure \ref{fig_sup:ae_samples}

\section{FID scores}
\label{sec_sup:fid}
FID scores for all CelebaA \cite{liu2015faceattributes} experiments are shown in table \ref{table_sup:fid}. We used 50k real images and 50k GAN images to calculate each FID score. We observe that the FID scores of nearest and bilinear interpolation methods are comparable or better than the Baseline FID for all GAN setups.

\begin{table}
\begin{center}
\begin{tabular}{c|c|c|c}\toprule
\textbf{Setup Code} &\textbf{DCGAN} &\textbf{LSGAN} &\textbf{WGAN-GP} \\\midrule
\textbf{Baseline} &\boldmath{$88.6$} & \boldmath{$73.26$} &\boldmath{$60.6$} \\
\hline
N.1.5 &$87.52$ &$70.69$ &$48.69$ \\
Z.1.5 &$69.14$ &$60.29$ &$47.73$ \\
B.1.5 &$84.65$ &$78.66$ &$52.18$ \\
\hline
N.1.7 &$90.8$ &$73.09$ &$60.11$ \\
Z.1.7 &$71.45$ &$59.55$ &$43.1$ \\
B.1.7 &$79.92$ &$76.33$ &$55.28$ \\
\hline
N.1.3 &$93.54$ &$74.06$ &$58.35$ \\
Z.1.3 &$65.46$ &$61.45$ &$56.91$ \\
B.1.3 &$76.04$ &$81.97$ &$58.55$ \\
\hline
N.3.5 &$73.63$ &$78.31$ &$55.47$ \\
Z.3.5 &$68.41$ &$66.27$ &$57.59$ \\
B.3.5 &$80.89$ &$72.29$ &$54.84$ \\
\hline
SR &$99.2$ &$86.16$ &$60.81$ \\
\bottomrule
\end{tabular}
\end{center}
\caption{
FID scores of GAN images trained on CelebA \cite{liu2015faceattributes} dataset. We include the FID scores of Spectral Regularized GANs (indicated as SR) for comparison.
}
\label{table_sup:fid}
\vspace{-0.2cm}
\end{table}

\section{Fourier Synthetic Image Detector}
We include more information and results for the synthetic image detector proposed by Dzanic \etal\cite{dzanic2020fourier}. All detection rates are averaged over 10 independent runs. 
\subsection{Classifier Implementation Details}
\label{sec_sup:dzanic_classifier_details}
The exact procedure used by Dzanic \etal\cite{dzanic2020fourier} to implement the classifier is shown below. For easier understanding, let us assume that we want to train a classifier to detect between real and StyleGAN images
\begin{enumerate}
  \item Collect a repository of 1000 StyleGAN images and 1000 real images.
  \item Obtain the un-normalized reduced spectrum for every real and StyleGAN image for the last 25\% spatial frequencies (0.75 - 1.0).
  \item Fit the reduced spectrum using power law function for every real and GAN image.
  \item Extract 3 features $b_{1}, b_{2}, b_{3}$ from the fitted spectrum where $b_{1}$: start value of the fitted spectrum, $b_{2}$: decay value of the fitted spectrum, $b_{3}$: end value of the fitted spectrum. Though the authors mention only $b_{1}, b_{2}$ in their paper, their official implementation contained $b_{3}$ as well. This difference is acceptable since $b_{3}$ is linearly dependent on $b_{1}, b_{2}$ under the assumption that the power law is a good fit.
  \item Train/ apply a binary KNN classifier (with k=5) using the 3 features extracted per image to predict if the GAN images are real or fake. The authors use 100 real and GAN images each to train the classifier and use the remaining 900 samples to test. 
\end{enumerate}

\subsection{Additional Results}
\label{sec_sup:additional_results}
Additional detection results when using 50\% data to train (authors used 10\% and these results are shown in the main paper) for CelebA \cite{liu2015faceattributes}, LSUN \cite{yu15lsun} and StarGAN \cite{choi2018stargan} experiments are shown in table \ref{table_sup:detection_celeba}, \ref{table_sup:detection_lsun}, \ref{table_sup:detection_stargan} respectively. We also conduct experiments using the reconstructed images and the detection results are shown in \ref{table_sup:detection_ae}.

\begin{table}
\begin{center}
\begin{adjustbox}{width=\columnwidth,center}
\begin{tabular}{l|c|c|c}\toprule
\textbf{Setup} &\textbf{DCGAN} &\textbf{LSGAN} &\textbf{WGAN-GP}\\
\hline
N.1.5 &\boldmath{$0.1\pm0$\%} &\boldmath{$0.28\pm0.04$\%} &\boldmath{$0.2\pm0$\%} \\
Z.1.5 &$82.18\pm0.26$\% &$86.05\pm0.43$\% &$99.7\pm0$\% \\
B.1.5 &\boldmath{$0\pm0$\%} &\boldmath{$0.1\pm0$\%} &\boldmath{$0.29\pm0.03$\%} \\
\hline
N.1.3 &\boldmath{$0\pm0$\%} &\boldmath{$0\pm0$\%} &\boldmath{$0.3\pm0$\%} \\
N.1.7 &\boldmath{$0\pm0$\%} &\boldmath{$0\pm0$\%} &\boldmath{$0.08\pm0.04$\%} \\
\hline
Z.1.3 &$98.43\pm0.13$\% &$71.77\pm0.48$\% &$97.79\pm0.03$\% \\
Z.1.7 &$96.55\pm0.07$\% &$94.59\pm0.09$\% &$99.9\pm0$\% \\
\hline
B.1.3 &\boldmath{$0\pm0$\%} &\boldmath{$0.12\pm0.04$\%} &\boldmath{$0.1\pm0$\%} \\
B.1.7 &\boldmath{$0\pm0$\%} &\boldmath{$0.1\pm0$\%} &\boldmath{$0.15\pm0.08$\%} \\
\hline
N.3.5 &\boldmath{$0.2\pm0$\%} &\boldmath{$0\pm0$\%} &\boldmath{$0\pm0$\%} \\
Z.3.5 &$74.07\pm0.68$\% &$62.17\pm1.05$\% &$99.97\pm0.05$\% \\
B.3.5 &\boldmath{$0\pm0$\%} &\boldmath{$0.49\pm0.03$\%} &\boldmath{$0.12\pm0.04$\%} \\
\bottomrule
\end{tabular}
\end{adjustbox}
\end{center}
\caption{
Detection results for the detectors proposed by
Dzanic \etal\cite{dzanic2020fourier}, using CelebA dataset (50\% data for training).
We follow exactly the procedure in~\cite{dzanic2020fourier} to train the detector for each
GAN model. The table shows the successful detection rates, and we highlight the cases when the detection rates  are inferior (less than $10\%$). The results are consistent with observations in the spectral plots. 
}
\label{table_sup:detection_celeba}
\vspace{-0.2cm}
\end{table}

\begin{table}
\begin{center}
\begin{adjustbox}{width=\columnwidth,center}
\begin{tabular}{l|c|c|c}\toprule
\textbf{Setup} &\textbf{DCGAN} &\textbf{LSGAN} &\textbf{WGAN-GP}\\
\hline
N.1.5 &\boldmath{$0.3\pm0\%$} &\boldmath{$0.6\pm0\%$} &\boldmath{$0\pm0\%$} \\
Z.1.5 &$98.41\pm0.15$\% &$94.84\pm0.07$\% &$99.93\pm0.05$\% \\
B.1.5 &\boldmath{$0.1\pm0\%$} &\boldmath{$0\pm0\%$} &\boldmath{$0.06\pm0.05\%$} \\
\bottomrule
\end{tabular}
\end{adjustbox}
\end{center}
\caption{
Detection results for the detectors proposed by
Dzanic \etal\cite{dzanic2020fourier}, using LSUN Bedrooms \cite{yu15lsun} dataset (50\% data for training).
The table shows the successful detection rates, and we highlight the cases when the detection rates  are inferior (less than $10\%$).
}
\label{table_sup:detection_lsun}
\vspace{-0.2cm}
\end{table}

\begin{table}
\begin{center}
\begin{adjustbox}{width=\columnwidth,center}
\begin{tabular}{l|c|c|c}\toprule
\textbf{Setup}& \textbf{N.1.5} &\textbf{Z.1.5} &\textbf{B.1.5} \\\midrule
\textbf{Accuracy}& $53.7\pm0.15$\% &$64.61\pm0.37$\% & \boldmath{$0\pm0\%$}\\
\bottomrule
\end{tabular}
\end{adjustbox}
\end{center}
\caption{
Detection results for the forensics classifiers proposed by Dzanic \cite{dzanic2020fourier}, using CelebA \cite{liu2015faceattributes} dataset (256x256) for StarGAN (50\% data for training). We observe that B.1.5 samples easily bypasses the classifier.
}
\label{table_sup:detection_stargan}
\vspace{-0.2cm}
\end{table}

\begin{table}
\begin{center}
\begin{tabular}{l|c|c}\toprule
\textbf{Setup} &\textbf{10\% train data} &\textbf{50\% train data} \\\midrule
N.1.5 &\boldmath{$0\pm0\%$} &\boldmath{$0\pm0\%$} \\
Z.1.5 &$94.8\pm1.75$\% &$95.44\pm0.26$\% \\
B.1.5 &\boldmath{$0\pm0\%$} &\boldmath{$0\pm0\%$} \\
\bottomrule
\end{tabular}
\end{center}
\caption{
Detection results for the forensics classifiers proposed by Dzanic \cite{dzanic2020fourier}, using reconstructed images. We observe that N.1.5 and B.1.5 samples can easily bypass the classifier.
}
\label{table_sup:detection_ae}
\vspace{-0.2cm}
\end{table}

\section{Stronger Classifiers}
\subsection{SVM and MLP classifiers}
We perform additional experiments using SVM and MLP classifiers (exact same setup as \cite{dzanic2020fourier}, only change in classifiers).
The results are shown in Table \ref{table_sup:svm} and Table \ref{table_sup:mlp}.
Our results are consistent:  even with SVM/MLP classifiers, we can bypass them by replacing zero insertion last layer with nearest (N) or bilinear (B). For all experiments, we use 10\% data for training, and the reported results are averaged over 10 runs. We also observe similar detection rates when using 50\% and 80\% data for training.

\begin{table}
\begin{center}
\begin{adjustbox}{width=\columnwidth,center}
\begin{tabular}{l|ccc}\toprule
\textbf{Setup} &\textbf{DCGAN} &\textbf{LSGAN} &\textbf{WGAN-GP}\\
\hline
N.1.5 &\boldmath{$0.1\pm0$\%} &\boldmath{$0.31\pm0.06$\%} &\boldmath{$0.23\pm0.16$\%} \\
Z.1.5 &$82.22\pm1.98$\% &$87.33\pm2.77$\% &$99.45\pm0.21$\% \\
B.1.5 &\boldmath{$0\pm0$\%} &\boldmath{$0.11\pm0.09$\%} &\boldmath{$0.25\pm0.17$\%} \\
\hline
N.1.3 &\boldmath{$0.01\pm0.03$\%} &\boldmath{$0.07\pm0.05$\%} &\boldmath{$0.35\pm0.22$\%} \\
N.1.7 &\boldmath{$0\pm0$\%} &\boldmath{$0\pm0$\%} &\boldmath{$0.05\pm0.05$\%} \\
\hline
Z.1.3 &$98.3\pm0.45$\% &$72.13\pm2.21$\% &$96.81\pm1.63$\% \\
Z.1.7 &$95.81\pm0.93$\% &$95.55\pm1.23$\% &$99.24\pm0.43$\% \\
\hline
B.1.3 &\boldmath{$0\pm0$\%} &\boldmath{$0.25\pm0.12$\%} &\boldmath{$0.15\pm0.15$\%} \\
B.1.7 &\boldmath{$0\pm0$\%} &\boldmath{$0.11\pm0.03$\%} &\boldmath{$0.3\pm0.27$\%} \\
\hline
N.3.5 &\boldmath{$0.1\pm0$\%} &\boldmath{$0\pm0$\%} &\boldmath{$0\pm0$\%} \\
Z.3.5 &$74.27\pm3.32$\% &$65.37\pm6.5$\% &$93.82\pm0.6$\% \\
B.3.5 &\boldmath{$0.04\pm0.07$\%} &\boldmath{$0.5\pm0.05$\%} &\boldmath{$0.21\pm0.14$\%} \\
\bottomrule
\end{tabular}
\end{adjustbox}
\end{center}
\caption{
Detection rates using SVM (RBF kernel) using same features as Dzanic \etal \cite{dzanic2020fourier}.
}
\label{table_sup:svm}
\vspace{-0.2cm}
\end{table}


\begin{table}
\begin{center}
\begin{adjustbox}{width=\columnwidth,center}
\begin{tabular}{l|ccc}\toprule
\textbf{Setup} &\textbf{DCGAN} &\textbf{LSGAN} &\textbf{WGAN-GP}\\
\hline
N.1.5 &\boldmath{$0.1\pm0$\%} &\boldmath{$0.77\pm0.15$\%} &\boldmath{$1.53\pm0.32$\%} \\
Z.1.5 &$81.14\pm2.9$\% &$83.88\pm2.59$\% &$99.77\pm0.09$\% \\
B.1.5 &\boldmath{$0.04\pm0.1$\%} &\boldmath{$0.87\pm0.46$\%} &\boldmath{$3.03\pm0.82$\%} \\
\hline
N.1.3 &\boldmath{$0.18\pm0.04$\%} &\boldmath{$0.05\pm0.13$\%} &\boldmath{$1.4\pm0.2$\%} \\
N.1.7 &\boldmath{$0\pm0$\%} &\boldmath{$0.04\pm0.05$\%} &\boldmath{$0.67\pm0.18$\%} \\
\hline
Z.1.3 &$97.54\pm0.41$\% &$72.65\pm2.64$\% &$98.11\pm0.44$\% \\
Z.1.7 &$94.53\pm0.97$\% &$93.07\pm1.6$\% &$99.97\pm0.05$\% \\
\hline
B.1.3 &\boldmath{$0.03\pm0.09$\%} &\boldmath{$1.6\pm0.54$\%} &\boldmath{$2.79\pm0.5$\%} \\
B.1.7 &\boldmath{$0.01\pm0.03$\%} &\boldmath{$0.42\pm0.29$\%} &\boldmath{$4.63\pm1.01$\%} \\
\hline
N.3.5 &\boldmath{$0.17\pm0.05$\%} &\boldmath{$0\pm0$\%} &\boldmath{$0.37\pm0.27$\%} \\
Z.3.5 &$74.88\pm2.79$\% &$71.22\pm4.46$\% &$99.8\pm0$\% \\
B.3.5 &\boldmath{$0.28\pm0.14$\%} &\boldmath{$1.89\pm0.45$\%} &\boldmath{$3.66\pm1.19$\%} \\
\bottomrule
\end{tabular}
\end{adjustbox}
\end{center}
\caption{
Detection rates using MLP (2 hidden layers of size 10 with sigmoid activation) using same features as Dzanic \etal  \cite{dzanic2020fourier}. 
}
\label{table_sup:mlp}
\end{table}

\begin{table}
\begin{center}
\begin{adjustbox}{width=\columnwidth,center}
\begin{tabular}{l|ccc}\toprule
\textbf{Setup} &\textbf{DCGAN} &\textbf{LSGAN} &\textbf{WGAN-GP}\\
\hline
N.1.5 &\boldmath{$0.12\pm0.04$\%} &\boldmath{$1.01\pm0.29$\%} &\boldmath{$0.08\pm0.06$\%} \\
Z.1.5 &$95.91\pm2.43$\% &$98.29\pm0.85$\% &$96.62\pm1.61$\% \\
B.1.5 &\boldmath{$0.09\pm0.03$\%} &\boldmath{$0.24\pm0.11$\%} &\boldmath{$0\pm0$\%} \\
\hline
N.1.3 &\boldmath{$0.26\pm0.07$\%} &\boldmath{$0.9\pm0.2$\%} &\boldmath{$0.08\pm0.04$\%} \\
N.1.7 &\boldmath{$0\pm0$\%} &\boldmath{$0.09\pm0.12$\%} &\boldmath{$0.12\pm0.04$\%} \\
\hline
Z.1.3 &$97.72\pm1.86$\% &$88.44\pm2.55$\% &$91.7\pm3.25$\% \\
Z.1.7 &$98.43\pm1.43$\% &$99.29\pm0.5$\% &$97.09\pm2.8$\% \\
\hline
B.1.3 &\boldmath{$0\pm0$\%} &\boldmath{$1.11\pm0.3$\%} &\boldmath{$0\pm0$\%} \\
B.1.7 &\boldmath{$0.01\pm0.03$\%} &\boldmath{$0.16\pm0.1$\%} &\boldmath{$0.17\pm0.05$\%} \\
\hline
N.3.5 &\boldmath{$0\pm0$\%} &$10.85\pm7.05$\% &\boldmath{$0\pm0$\%} \\
Z.3.5 &$97.32\pm1.24$\% &$97.79\pm1.4$\% &$98.69\pm1.77$\% \\
B.3.5 &\boldmath{$0.23\pm0.05$\%} &\boldmath{$1.09\pm0.17$\%} &\boldmath{$0.02\pm0.04$\%} \\
\bottomrule
\end{tabular}
\end{adjustbox}
\end{center}
\caption{
Detection rates using classifier proposed by Durall \etal \cite{Durall_2020_CVPR}. 
Following \cite{Durall_2020_CVPR},
we use entire 1D spectrum as features.
}
\label{table_sup:rich_features}
\end{table}

\subsection{Using entire spectrum as features}
We followed similar setup as Durall \etal \cite{Durall_2020_CVPR} to train a SVM classifier using entire spectrum as features (88 dimensional features for 128x128 images). The finding is consistent: we observe that the features are still non-separable when using nearest (N) and bilinear (B) for the last upsampling step.
The results are shown in Table \ref{table_sup:rich_features}.

\section{Virtual KITTI}
\begin{figure}
    \centering
    \includegraphics[width=0.95\linewidth]{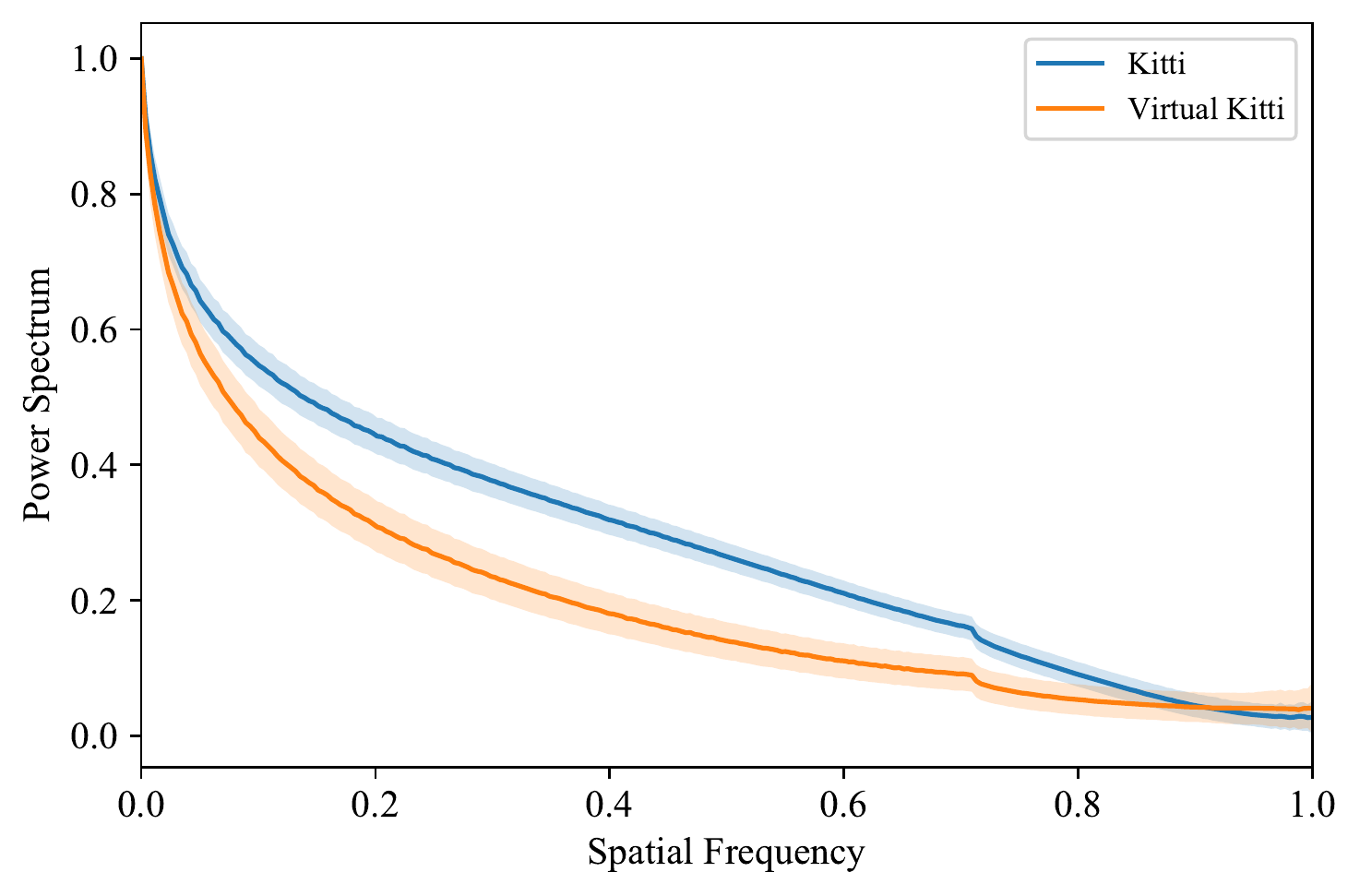}
    \caption{This figure shows the spectral plots for all the frequencies for KITTI \cite{Geiger2012CVPR} and Virtual KITTI \cite{vkitti} datasets. All the images were center cropped to 370x370. We observe that the Virtual KITTI images do not have high frequency discrepancies.}
    \label{fig_sup:game-engine-spectral-decay}
\end{figure}

\begin{figure*}
    \centering
    \includegraphics[width=0.95\linewidth]{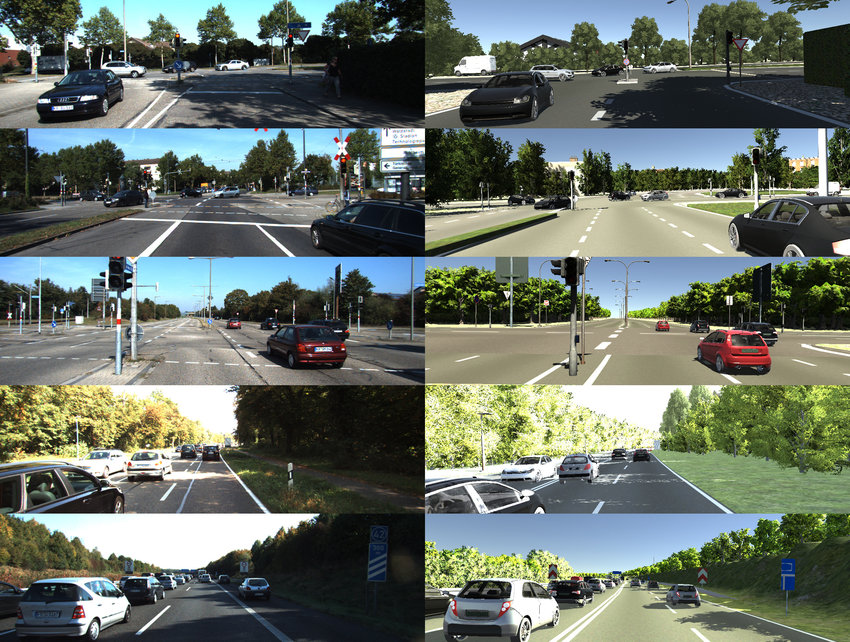}
    \caption{Samples of real images from KITTI tracking benchmark \cite{Geiger2012CVPR} dataset and the recreated images using Unity game engine obtained from Virtual KITTI \cite{vkitti} dataset (Right)}.
    \label{fig_sup:game-engine-samples}
    \vspace{-0.5cm}
\end{figure*}

GANs belong to a larger family of computational image synthesis algorithms. In this section, we investigate the high frequency decay attributes of data created entirely using Unity Game Engine. We compare the spectral behaviour between the Official KITTI tracking benchmark \cite{Geiger2012CVPR} (Real images) and the Virtual KITTI \cite{vkitti} (Synthesized images). Virtual KITTI \cite{vkitti} recreates real-world videos from the KITTI tracking benchmark \cite{Geiger2012CVPR} inside Unity\footnote{\url{https://unity.com/}} game engine. We show some samples in Figure \ref{fig_sup:game-engine-samples}. We show the entire power spectrum in Figure \ref{fig_sup:game-engine-spectral-decay} and we observe that the images synthesized from the game engine do not have high frequency spectral discrepancies.

\section{CRN/ IMLE}
\begin{figure}
    \centering
    \includegraphics[width=0.95\linewidth]{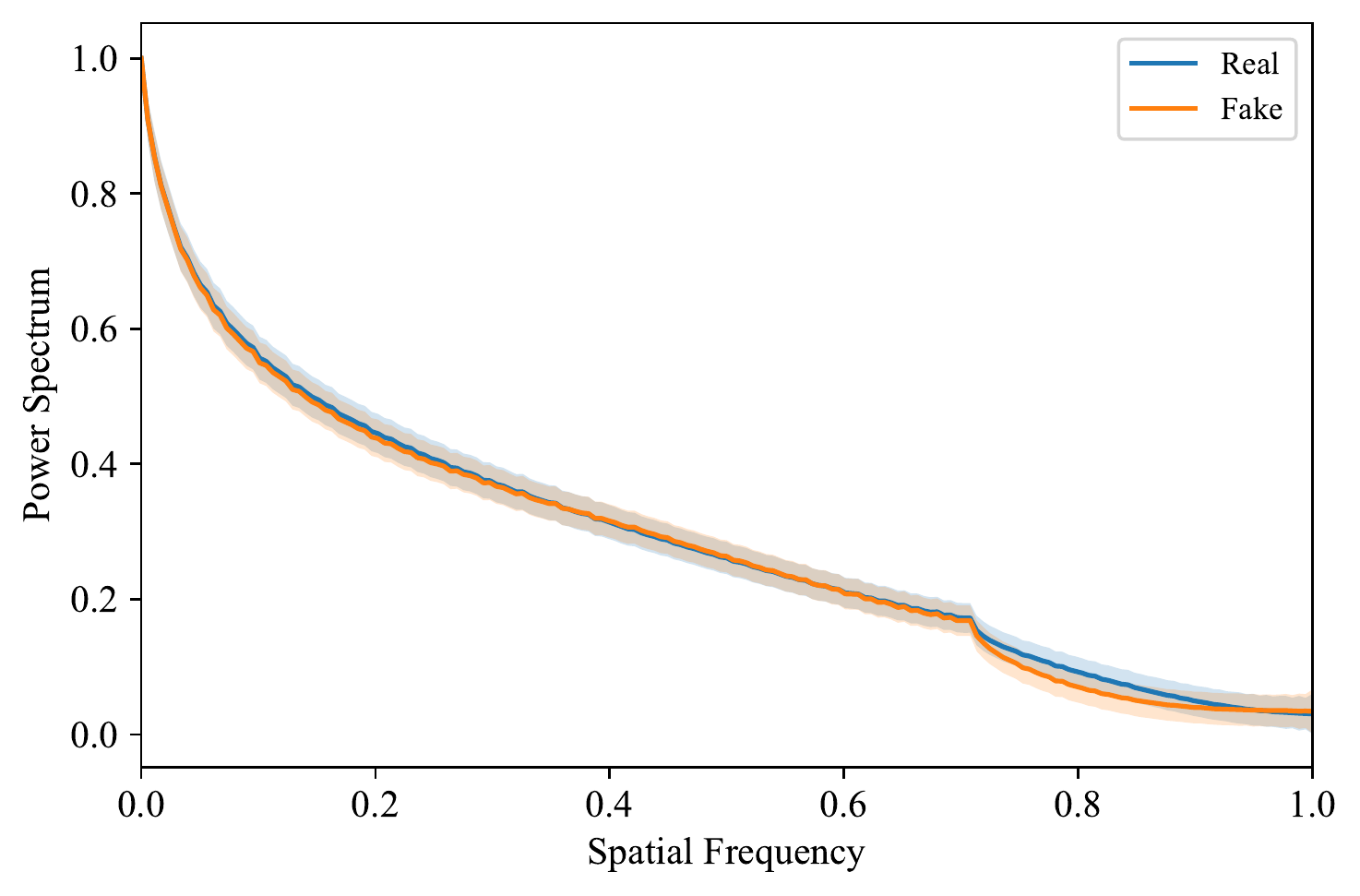}
    \caption{This figure shows the spectral plots for all the frequencies for GTA (Real) images and CRN \cite{Chen2017PhotographicIS} synthesized images. All the images were center cropped to 256x256. We observe that the CRN generated images do not have high frequency discrepancies.}
    \label{fig_sup:crn-spectral-decay}
\end{figure}

\begin{figure}
    \centering
    \includegraphics[width=0.95\linewidth]{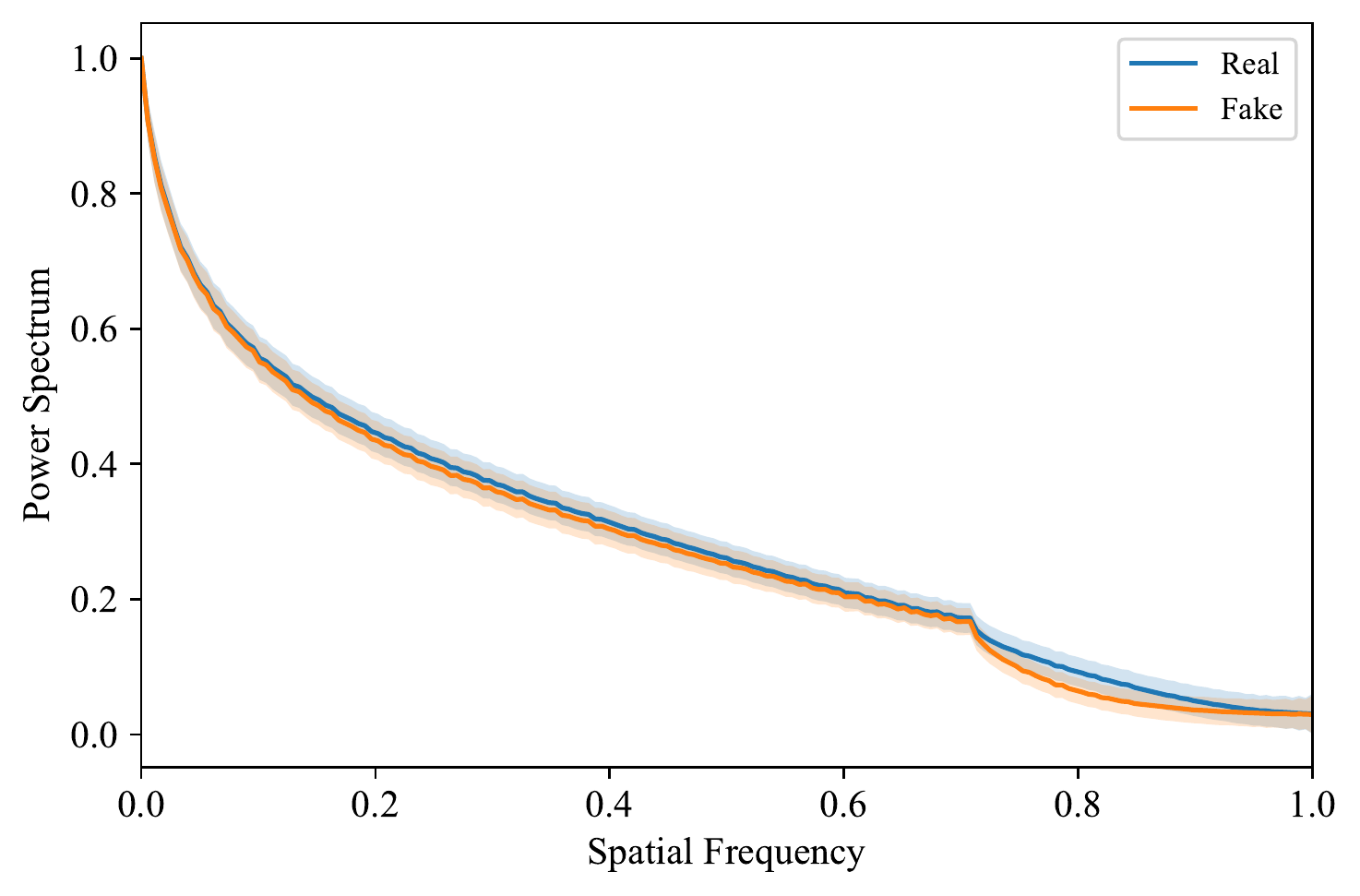}
    \caption{This figure shows the spectral plots for all the frequencies for GTA (Real) images and IMLE \cite{imle} synthesized images. All the images were center cropped to 256x256. We observe that the IMLE synthesized images do not have high frequency discrepancies.}
    \label{fig_sup:imle-spectral-decay}
    \vspace{-0.2cm}
\end{figure}

We also observe that high frequency decay discrepancies are not seen in some out-of-the-box GAN models.
Specifically, we observe that CRN \cite{Chen2017PhotographicIS} and IMLE \cite{imle} GANs do not have such discrepancies. We show the entire power spectrum for CRN \cite{Chen2017PhotographicIS} and IMLE \cite{imle} GANs in Figure \ref{fig_sup:crn-spectral-decay} and \ref{fig_sup:imle-spectral-decay} respectively. Do note that both these models are pre-trained on GTA game data (Another instance of data synthesized from game engines). This further helps to confirm that such discrepancies are not intrinsic.

\section{Dataset Details}
For CelebA \cite{liu2015faceattributes} experiments, we use the officially released train subset consisting of 162, 770 images. For LSUN \cite{yu15lsun} experiments, we select a random subset of 200, 000 images for training. For StarGAN experiments, we follow the official implementation. For autoencoder experiments, we select a random subset of 1000 images from CelebA-HQ \cite{karras2018progressive}.

\section{Implementation Details}
For GAN experiments, we use Adam optimizer with $\beta_{1}=0.5$, $\beta_{2}=0.999$ and batch size 64. For all CelebA \cite{liu2015faceattributes} experiments, we used an initial learning rate = $2\times10^{-4}$. The learning rate was reduced based on FID scores for all experiments.
For DCGAN, LSGAN and WGAN-GP experiments, we use the GitHub code used by the Spectral Regularization paper\cite{Durall_2020_CVPR} \footnote{\url{https://github.com/LynnHo/DCGAN-LSGAN-WGAN-GP-DRAGAN-Pytorch}}.
For StarGAN \cite{choi2018stargan}, we use the official implementation \footnote{\url{https://github.com/yunjey/stargan}} with default hyper-parameters.
For Spectral Regularization \cite{Durall_2020_CVPR} experiments, we use the officially released code \footnote{\url{https://github.com/cc-hpc-itwm/UpConv}}.

For all autoencoder experiments, we use Adam optimizer with $\beta_{1}=0.9$ and $\beta_{2}=0.999$, batch size 128, initial learning rate $2.5\times10^{-3}$ and learning rate decay scheme that scales the learning rate by 0.9 when reconstruction error plateaus.

For Fourier synthetic detector \cite{dzanic2020fourier} experiments, we use the officially released matlab code \footnote{\url{https://github.com/tarikdzanic/FourierSpectrumDiscrepancies}} for feature extraction and use our own script to perform KNN classification. For FID calculation, we used the open-source Pytorch FID implementation \footnote{\url{https://github.com/mseitzer/pytorch-fid}}.

More details on hyper-parameters and research reproducibility can be found in: \url{https://keshik6.github.io/Fourier-Discrepancies-CNN-Detection/}

\end{document}